\documentclass[letterpaper, 10 pt, journal, twoside]{IEEEtran}

\usepackage{algorithm}
\usepackage[noend]{algorithmic}
\usepackage{amssymb}
\usepackage{amsmath}
\usepackage{amsfonts}       %
\usepackage{arydshln}
\usepackage{booktabs}       %
\usepackage{balance}
\usepackage{booktabs}
\usepackage[utf8]{inputenc} %
\usepackage[T1]{fontenc}    %
\usepackage{color, colortbl}
\usepackage{collcell}
\usepackage[pdftex, pdfstartview={FitV}, pdfpagelayout={TwoColumnLeft},bookmarksopen=true,plainpages = false, colorlinks=true, linkcolor=black, citecolor = black, urlcolor = black,filecolor=black , pagebackref=false,hypertexnames=false, plainpages=false, pdfpagelabels,bookmarks=false ]{hyperref}
\usepackage{url}            %
\usepackage{nicefrac}       %
\usepackage{microtype}      %
\usepackage{makecell}

\usepackage[font=scriptsize]{caption}
\usepackage{subcaption}
\usepackage[]{siunitx}
\usepackage{graphicx}
\usepackage{epstopdf} %
\usepackage{mathtools}
\usepackage{marginnote}
\usepackage[dvipsnames, table]{xcolor} %
\usepackage{mdframed}           %
\usepackage[most]{tcolorbox}    %
\usepackage{float}
\usepackage[capitalize]{cleveref}
\usepackage[]{algorithmic}
\usepackage{makecell}
\usepackage{multirow}
\usepackage{paralist}
\floatstyle{boxed} \floatname{algorithm}{Algorithm}
\newfloat{algorithm}{t}{loa}[section]
\usepackage{xcolor}
\usepackage[sort,compress]{cite}
\usepackage{soul}
\usepackage{etoolbox}
\usepackage{pgf}
\usepackage{mathrsfs}

\usepackage{xstring}
\usepackage{color}
\usepackage{soul}
\usepackage{xspace}
\usepackage[capitalize]{cleveref}
\usepackage{dsfont}
\usepackage{bm}
\usepackage{bbm}
\usepackage{nicefrac}  %
\usepackage{amsfonts}
\usepackage{amssymb}
\usepackage{array}
\usepackage{mathtools}
\usepackage{subcaption}
\usepackage{fontawesome}
\usepackage{thmtools}
\usepackage{thm-restate}
\usepackage{lipsum}
\usepackage{pifont}
\usepackage{makecell}
\usepackage{multirow}
\usepackage{afterpage}
\usepackage{tabularx} 
\usepackage{textcase}
\usepackage{tikz}
\usepackage{url}
\usepackage[normalem]{ulem}
\usepackage{wrapfig}  %
\usepackage{textcomp}
\usepackage{xifthen}%

\usepackage[nolist,nohyperlinks]{acronym}
\begin{acronym}
\acro{OC}{Optimal Control}
\acro{LQR}{Linear Quadratic Regulator}
\acro{MAV}{Micro Aerial Vehicle}
\acro{GPS}{Guided Policy Search}
\acro{UAV}{Unmanned Aerial Vehicle}
\acro{MPC}{Model Predictive Control}
\acro{NMPC}{Nonlinear Model Predictive Control}
\acro{RTMPC}{Robust Tube MPC}
\acro{DNN}{deep neural network}
\acro{DA}{data augmentation}
\acro{BC}{Behavior Cloning}
\acro{DR}{Domain Randomization}
\acro{SA}{Sampling Augmentation}
\acro{IL}{Imitation Learning}
\acro{DAgger}{Dataset-Aggregation}
\acro{MDP}{Markov Decision Process}
\acro{VIO}{Visual-Inertial Odometry}
\acro{VSA}{Visuomotor Sampling Augmentation}
\acro{CoM}{Center of Mass}
\acro{SD}{Standard Deviation}
\acro{MSE}{Mean Squared Error}
\acro{RMSE}{Root Mean Square Error}
\acro{CSWaP}{cost, size, weight and power}
\acro{MAV}{Micro Aerial Vehicle}
\acro{UKF}{Unscented Kalman filter}
\acro{EKF}{extended Kalman filter}
\acro{PF}{particle filter}
\acro{LSTM}{Long Short-Term Memory}
\acro{RNN}{recurrent neural network}
\acro{CNN}{convolutional neural network}
\acro{CoM}{center of mass}
\acro{USQUE}{Unscented Quaternion estimator}
\acro{UT}{Unscented Transformation}
\acro{MSE}{Mean Squared Error}
\acro{GP}{Gaussian Process}
\acrodefplural{GP}[GPs]{Gaussian Processes}
\acro{MAP}{Maximum a Posterior}
\acro{IMU}{Inertial Measurement Unit}
\acro{RMSE}{Root-Mean-Square error}
\acro{RTE}{Relative Translation error}
\acro{AIO}{Airflow-Inertial Odometry}
\acro{VO}{visual odometry}
\acro{RBF}{radial basis function}
\acro{MRP}{Modified Rodriguez Parameters}
\acro{DDP}{differential dynamic programming}
\acro{iLQR}{iterative linear quadratic regulator}
\acro{QP}{quadratic program}
\acro{SQP}{sequential quadratic program}
\acro{NLP}{nonlinear program}
\acro{KKT}{Karush–Kuhn–Tucker}
\acro{OCP}{Optimal Control Problem}
\acro{RK}{Runge-Kutta}
\acro{RTI}{real-time iteration}
\acro{RL}{Reinforcemement Learning}
\acro{RTI}{Real-Time Iteration}
\end{acronym}

\crefname{figure}{Fig.}{Figs.}
\Crefname{figure}{Figure}{Figures}
\crefname{table}{Table}{Tables}
\crefformat{table}{Table~#1}
\crefformat{equation}{Eq. (#1)}
\crefformat{algorithm}{Algorithm~#1}
\crefformat{appendix}{Appendix~#1}
\Crefformat{appendix}{Appendix~#1}
\crefformat{appendices}{Appendices~#1}
\Crefformat{appendices}{Appendices~#1}

\newcommand\sdots{\hbox to 1em{.\hss.\hss.}} %
\DeclareMathAlphabet\mathbfcal{OMS}{cmsy}{b}{n}  %

\newcommand{\dset}{\mathcal{D}}

\newcommand{\vbf}[2][]{
    \ifthenelse{\isempty{#1}}%
    {\mathbf{#2}}%
    {\prescript{}{#1}{\mathbf{#2}}}%
}
\newcommand{\vbfd}[2][]{
    \ifthenelse{\isempty{#1}}%
    {\dot{\mathbf{#2}}}%
    {\prescript{}{\text{#1}}{\dot{\mathbf{#2}}}}%
}

\newcommand{\vbs}[2][]{
    \ifthenelse{\isempty{#1}}%
    {\boldsymbol{#2}}%
    {\prescript{}{\text{#1}}{\boldsymbol{#2}}}%
}
\newcommand{\vbsd}[2][]{
    \ifthenelse{\isempty{#1}}%
    {\dot{\boldsymbol{#2}}}%
    {\prescript{}{\text{#1}}{\dot{\boldsymbol{#2}}}}%
}

\newcommand{\xdes}{\mathbf{x}^\text{des}}    %
\newcommand{\xsafe}{\bar{\mathbf{x}}}   %
\newcommand{\usafe}{\bar{\mathbf{u}}}   %

\newcommand{\tzr}{{t_0}}

\newcommand{\expert}{\pi_{\vbs{\theta}^*}}
\newcommand{\student}[1]{\pi_{\hat{\vbs{\theta}}_{#1}}}

\newcommand{\anydomain}{{\mathcal{E}}}

\newif\ifcomments
\commentstrue

\ifcomments
	\newcommand{\aXX}[1]{\color{OliveGreen}AT: (#1)\color{black}\xspace}  %
	\newcommand{\XX}[1]{\color{red}JH: (#1)\color{black}\xspace}  %
\else
    \newcommand{\dXX}[1]{}  %
	\newcommand{\aXX}[1]{}  %
	\newcommand{\XX}[1]{}  %
\fi

\newcommand{\PreserveBackslash}[1]{\let\temp=\\#1\let\\=\temp}
\newcolumntype{C}[1]{>{\PreserveBackslash\centering}p{#1}}
\newcolumntype{R}[1]{>{\PreserveBackslash\raggedleft}p{#1}}
\newcolumntype{L}[1]{>{\PreserveBackslash\raggedright}p{#1}}

\definecolor{LightCyan}{rgb}{0.88,1,1}

\renewcommand{\b}[1]{#1}

\newif\ifblue
\bluefalse

\begin{document}
\title{Efficient Deep Learning of Robust Policies\\ 
from MPC using Imitation and\\ Tube-Guided Data Augmentation}
\author{Andrea Tagliabue and Jonathan P.\ How%
	\thanks{The authors are with the Department of Aeronautics and Astronautics, Massachusetts Institute of Technology.
	    {\texttt{\{atagliab, jhow\}@mit.edu.}}}
    \thanks{We thank Prof.\ Michael Everett,  Dr.\ Dong-Ki Kim, Tong Zhao and Prof.\ Donggun Lee, Kota Kondo,  and Xiaoyi Cai for feedback and discussions. Work funded by the AFOSR MURI FA9550-19-1-0386.}
}%

\maketitle

\begin{abstract}
\b{Imitation Learning (IL) can generate computationally efficient policies from demonstrations provided by Model Predictive Control (MPC). However, IL methods often require extensive data-collection and training-efforts, limiting changes to the policy if the task changes, and they produce policies with limited robustness to new disturbances}. In this work, we propose an IL strategy to \textit{efficiently} compress a computationally expensive MPC into a deep neural network policy that is \textit{robust} to previously unseen disturbances. 
By using a robust variant of the MPC, called Robust Tube MPC, and leveraging properties from the controller, \b{we introduce computationally-efficient data augmentation methods that enable} a significant reduction of the number of MPC demonstrations and \b{training efforts} required to generate a robust policy. %
Our approach opens the possibility of \textit{zero-shot} transfer of a policy trained from a single MPC demonstration collected in a nominal domain, such as a simulation or a robot in a lab/controlled environment, to a new domain with previously unseen bounded model errors/perturbations. 
Numerical \b{evaluations performed using linear and nonlinear MPC for agile flight on a multirotor show that our method outperforms strategies commonly employed in IL (such as Dataset-Aggregation (DAgger) and Domain Randomization (DR)) in terms of demonstration-efficiency, training time, and robustness to perturbations unseen during training. Experimental evaluations validate the efficiency and real-world robustness}. %
\end{abstract}

\vspace{-2ex}

\section*{Supplementary Material}
\b{Video: \url{https://youtu.be/-uiarBY1STU}}
\vspace{-2ex}

\acresetall
\section{Introduction} \label{sec:introduction}
\ac{MPC}~\cite{borrelli2017predictive, rawlings2017model} enables impressive performance on complex, agile robots~\cite{lopez2019dynamic, lopez2019adaptive, li2004iterative, kamel2017linear, minniti2019whole, williams2016aggressive}. However, its computational cost often limits the opportunities for onboard, real-time deployment\b{~\cite{ctx16203176620006761} on platforms with limited computation~\cite{chen2021collision,  giernacki2017crazyflie}, or diverts critical computing power} needed by other components governing the autonomous system. 
Recent works have mitigated \ac{MPC}'s computational requirements by relying on computationally efficient \ac{DNN} policies that are trained to imitate task-relevant demonstrations generated by \ac{MPC} in an offline training phase.
Such demonstrations are generally collected via \ac{IL}~\cite{kaufmann2020deep, ross2013learning, reske2021imitation}, where the \ac{MPC} acts as an expert that provides demonstrations, and the \ac{DNN} policy is treated as a student, trained via supervised learning. 

\begin{figure}[h!]
    \centering
    
    \includegraphics[width=0.99\columnwidth]{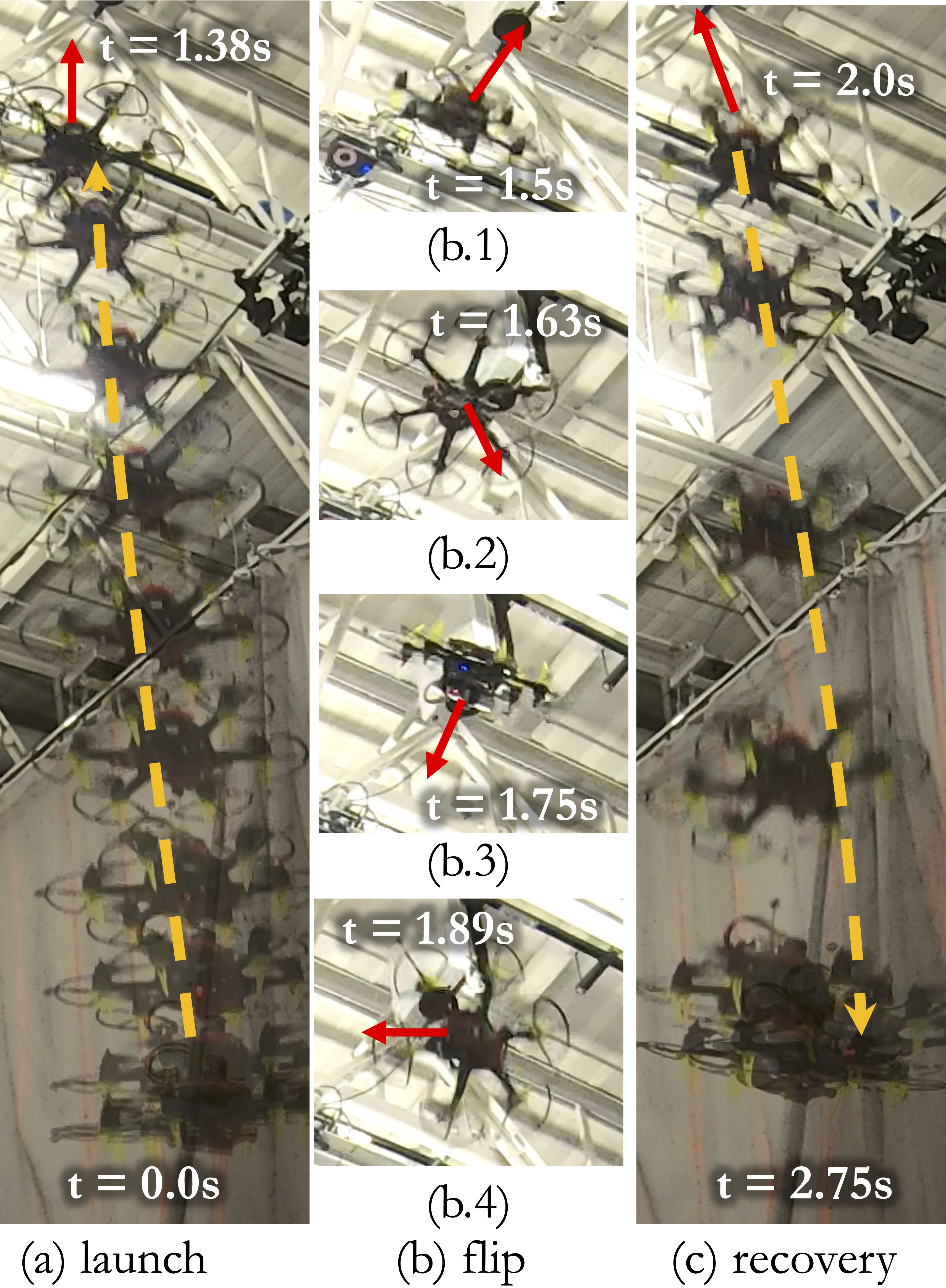}
    \caption{Time-lapse of a multirotor performing a flip using a DNN policy learned via the proposed approach. The policy is learned offboard efficiently (\textbf{in only $\boldsymbol{100}$ s of training time}), and deployed onboard (\texttt{NVIDIA Jetson TX2}, CPU), tested at \textbf{up to $\boldsymbol{500}$ Hz, with an average inference time of $\boldsymbol{15}$ $\boldsymbol{\mu}$s}). (a) Upwards acceleration phase (red arrow: thrust vector, yellow arrow: trajectory). (b) $360^\circ$ rotation around the body $x$-axis in $\sim 0.5$ s. (c) Deceleration phase.} 
    \label{fig:flip_timelapse}
    \vspace{-3ex}
\end{figure}

A common issue in existing \ac{IL} methods (e.g., \ac{BC}~\cite{pomerleau1989alvinn, osa2018algorithmic, bojarski2016end}, \ac{DAgger}~\cite{ross2011reduction}) is that they require to collect a relatively large number of \ac{MPC} demonstrations, even for a single task like tracking a specific trajectory. \b{This sample-inefficiency introduces significant challenges: (i) it necessitates a substantial number of queries to the resource-intensive \ac{MPC} expert, requiring expensive training equipment; (ii) it hinders learning from very high-dimensional MPC experts; (iii) it results in a considerable volume of queries to the training environment, limiting data collection in computationally intensive simulations  or demanding numerous hours of real-time demonstrations on a physical robot, which is impractical. Moreover, this approach complicates updating the policy when the MPC expert undergoes changes due to (iv) tuning or (v) model updates, or when (vi) performing new tasks is required, such as tracking different sets of trajectories. These aspects can be particularly critical in fields such as chemical and process control, where state dimension can reach $252$ and planning horizon of $140$ nodes \cite{VANANTWERP20001, WANG2023103049, ctx16203176620006761}, resulting in computationally intractable online optimization problems and offline policy generation procedures.}

One of the causes for such demonstration-inefficiency is the need to take into account and correct for the compounding of errors (\textit{covariate \text{or} distribution shifts}) in the learned policy~\cite{ross2011reduction}, which may otherwise create catastrophic consequences~\cite{pomerleau1989alvinn}. These distribution shifts can be caused by: 
\begin{inparaenum}[(a)]
    \item mismatches, e.g., due to modeling errors, between the simulator used to collect demonstrations and the deployment domain (i.e., sim2real gap);
    \item learning errors in the policy; or
    \item model changes or disturbances that may not be present in a controlled training environment (lab/factory when training on a real robot), but that do appear during deployment in the real-world (i.e., lab2real gap).
\end{inparaenum} 
Approaches employed to compensate for these \textit{gaps} and generate a robust policy, such as \ac{DR}~\cite{peng2018sim, loquercio2019deep}, introduce additional challenges, such as the need to apply disturbances or model changes during training.
\b{Data and computational-efficiency challenges in IL can be mitigated by DA strategies, based on augmenting the training data with extra input-output samples \textit{efficiently-generated} from the collected demonstrations~\cite{pomerleau1989alvinn, bojarski2016end, levine2013guided, krishnamoorthy2022sensitivity, krishnamoorthy2023improved}. However, existing methods for MPC \cite{levine2013guided, krishnamoorthy2022sensitivity, krishnamoorthy2023improved} do not explicitly account for uncertainties, not only in the way the demonstration are generated, but more importantly in the way the samples are generated, resulting in policies with limited robustness to uncertainties.}

\par

\begin{figure}
    \centering
    \includegraphics[width=\columnwidth, trim={0.3in, 9.25in, 4.9in, 0.4in}]{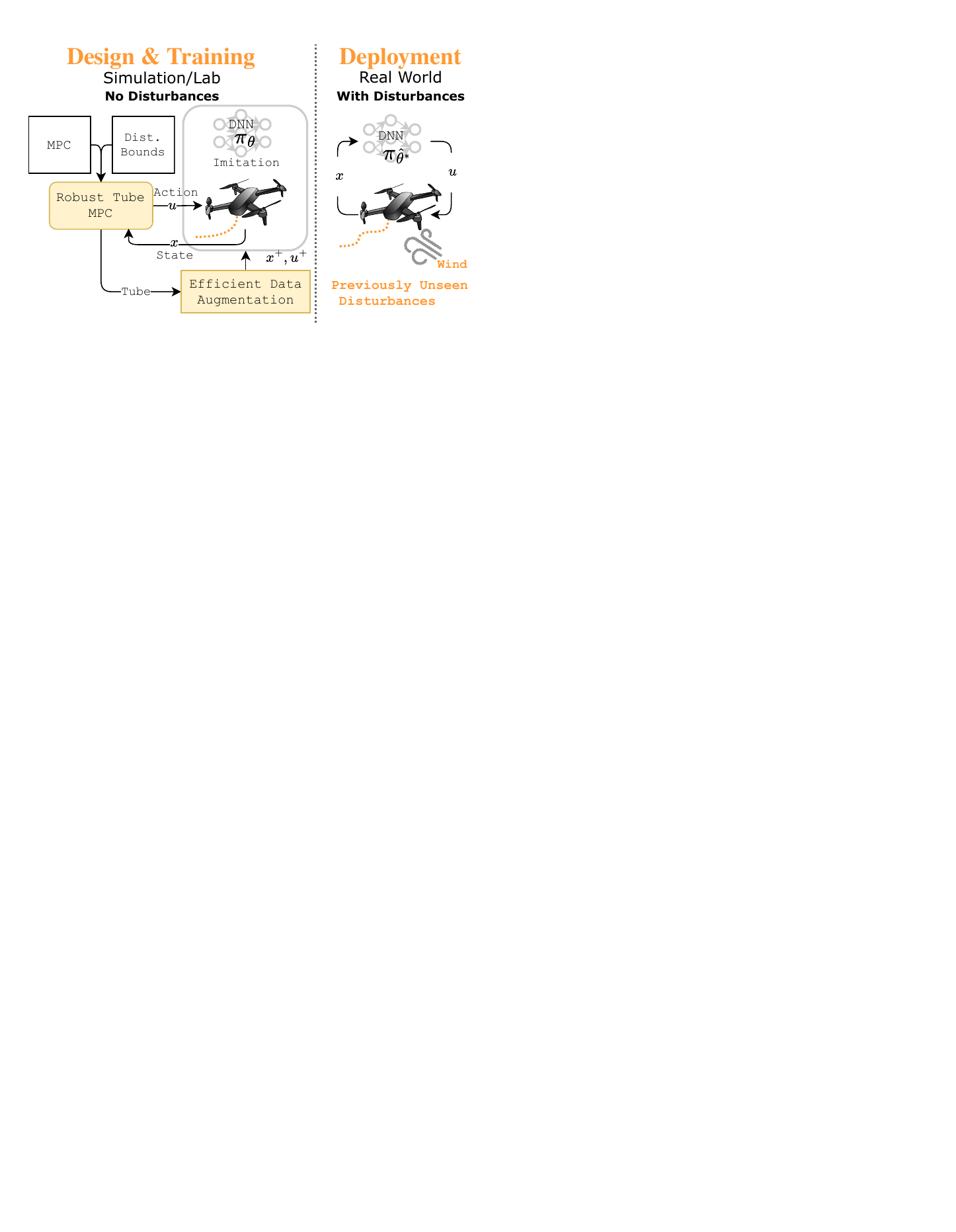}
    \caption{Overview of the approach proposed to generate a \ac{DNN}-based policy $\pi_\theta$ from a computationally expensive \ac{MPC} in a data and compute-efficient way. We do so by generating a \ac{RTMPC} using bounds of the disturbances encountered in the deployment domain. We use properties of the tube to derive a computationally efficient \ac{DA} strategy that generates extra state-action pairs $(x^+, u^+)$, obtaining $\pi_{\hat{\theta}^*}$ via \ac{IL}. Our approach enables zero-shot transfer from a single demonstration collected in simulation (\textit{sim2real}) or a controlled environment (lab, factory, \textit{lab2real}).}
    \label{fig:sim_to_real_zero_shot_cover} %
\end{figure}

\subsection{Efficient, Robust IL from MPC via Sampling Augmentation}
In this work, we address the problem of generating a robust \ac{DNN} policy from \ac{MPC} in a demonstration and computationally efficient manner by designing a computationally-efficient \ac{DA} strategy that systematically compensates for the effects of covariate shifts that might be encountered during real-world deployment. Our approach, named \acf{SA} and depicted in \cref{fig:sim_to_real_zero_shot_cover}, relies on a prior model of the perturbations/uncertainties encountered in a deployment domain, which is used to generate a robust version of the given \ac{MPC}, called \ac{RTMPC}, to collect demonstrations and to guide the \ac{DA} strategy. The key idea behind this \ac{DA} strategy consists in observing that the \ac{RTMPC} framework provides:
\begin{inparaenum}[(a)]
    \item information on the states that the robot may visit when subject to uncertainty. This is represented by a \textit{tube} that contains the collected demonstration; the tube can be used to identify/generate extra relevant states for \ac{DA}; and
    \item an \textit{ancillary controller} that maintains the robot inside the tube regardless of the realization of uncertainties; this controller can be used to generate extra actions.
\end{inparaenum}
\b{To numerically and experimentally validate our approach, we tailor \ac{SA} to the task of} efficiently learning robust policies for agile flight on a multirotor. First, we demonstrate in experiments trajectory tracking capabilities with a policy learned from a linear trajectory tracking \ac{RTMPC}. The policy is learned from a \textit{single} demonstration collected in simulation or directly on the real robot, and it is robust to previously-unseen wind disturbances. Second, we demonstrate the ability to generate a policy from a go-to-goal-state nonlinear \ac{RTMPC} capable of performing acrobatic maneuvers, such as a  $360$ degrees flip. These maneuvers are performed under real-world uncertainties, using a policy obtained from only \textit{two} demonstrations and in less than $100$s of training time. 

\subsection{Related Work}
\noindent
\textbf{Explicit MPC} \cite{borrelli2017predictive} approximates linear MPC by pre-computing a policy (look-up table or \acp{DNN} \cite{chen2022large}) offline, partitioning the state space. However, its memory and computational complexity grow exponentially with the number of constraints. Our work addresses this by training efficient \ac{DNN} policies \cite{ctx16203176620006761} through task-relevant demonstrations and \ac{IL}, focusing on the most relevant parts of the policy input space and learning from \acp{MPC} with nonlinear models.

\noindent
\textbf{IL from MPC.}
Imitation-learned policies from \ac{MPC} are widely used in robotics. 
Close to our work, \cite{kaufmann2020deep} learns to perform acrobatic maneuvers with a quadrotor from \ac{MPC} using \ac{DAgger} combined with \ac{DR}, by collecting $150$ demonstrations in simulation. Ref.~\cite{pan2020imitation} uses \ac{DAgger} combined with an \ac{MPC} based on \ac{DDP} \cite{jacobson1970differential} for agile off-road autonomous driving using about $24$ laps\footnote{Obtained using Table 2 in \cite{pan2020imitation}, considering $6000$ observation/action pairs sampled at $50$ Hz while racing on a $30$ m long racetrack with an average speed of $6$ m/s.} around their racetrack for the first \ac{DAgger} iteration.
These examples show the \textit{performance} that can be achieved using \ac{IL} from \ac{MPC}, but they also highlight that current methods require a large number of interactions with the \ac{MPC} and the training environment, resulting in longer training times or complex data collection procedures, as summarized in \cref{tab:state_of_the_art_comparison}.

\begin{table}[!t]
\newcommand{\No}{\textbf{\textcolor{red}{No}} \cellcolor{Red!10}}
\newcommand{\Yes}{\textbf{\textcolor{ForestGreen}{Yes}} \cellcolor{LimeGreen!25}} 
    \vskip2ex
    \renewcommand{\arraystretch}{1.4}
    \setlength{\tabcolsep}{1pt} %
    \scriptsize
    \caption{Strategies for policy learning from model-based planners/controllers. Our work is the only method that enables efficient learning of policies that account for uncertainties.}
    
    \vskip-1ex
    \begin{centering}
    \label{tab:state_of_the_art_comparison}
    \resizebox{1.0\columnwidth}{!}{
    \ifblue
    \begin{tikzpicture}
    \node[inner sep=0pt] (table) {
    \fi
    \begin{tabular}{
    >{\centering\arraybackslash}m{0.23\columnwidth} 
    >{\centering\arraybackslash}m{0.18\columnwidth}
    >{\centering\arraybackslash}m{0.15\columnwidth}
    >{\centering\arraybackslash}m{0.15\columnwidth}
    >{\centering\arraybackslash}m{0.18\columnwidth}
    >{\centering\arraybackslash}m{0.18\columnwidth}
    >{\centering\arraybackslash}m{0.15\columnwidth}
    >{\centering\arraybackslash}m{0.15\columnwidth} %
    }
    \toprule 
    \textbf{Method} & 
    \textbf{Explicitly Accounts for uncertainties} & 
    \textbf{Data-efficient training} &
    \textbf{Compute-efficient training} &
    \textbf{Allows both off/on-policy data collection} &
    \textbf{Demonstrations from both sim. and real robot} &
    \textbf{State $\geq10$ and underactuated} &
    \textbf{Real-world and agile deployment} %
    \tabularnewline
    \midrule
    \midrule
    \textbf{BC} \cite{pomerleau1989alvinn}      & \No   & \No  & \No  & n.a.  & \Yes   & \Yes & \No \tabularnewline
    \cline{0-0}
    \textbf{DAgger} \cite{ross2013learning, pan2020imitation}  & \No   & \No  & \No  & n.a.   & \Yes   & \Yes & \Yes  \tabularnewline
    \cline{0-0}
    \textbf{DR} \cite{peng2018sim, loquercio2019deep}           & \Yes   & \No  & \No  & \Yes  & \Yes   & \Yes & \Yes  \tabularnewline
    \cline{0-0}
    \textbf{GPS} \cite{levine2013guided, kahn2017plato}         & \No   & \Yes  & \Yes  & n.a.  & \No   & \Yes & \No  \tabularnewline
    \cline{0-0}
    \textbf{MPC-Net} \cite{carius2020mpc}                       & \No   & \Yes  & \Yes  &  \No  & \Yes  & \Yes & \No \tabularnewline
    \cline{0-0}
    \textbf{LAG-ROS} \cite{tsukamoto2021learning}               & \Yes  & n.a.  & \No   & \No   & \No   & \No  & \No  \tabularnewline
    \cline{0-0}
    \cite{krishnamoorthy2022sensitivity} (DA)                   & \No   & \No  & \Yes   & \No   & \No   & \No   & \No  \tabularnewline
    \cline{0-0}
    \cite{krishnamoorthy2023improved} (DA)                      & \No   & \Yes   & \Yes  & \No  & \No   & \No   & \No  \tabularnewline
    \cline{0-0}
    \hline
    \hline
    \textbf{SA} (proposed) & \Yes & \Yes & \Yes &  \Yes & \Yes & \Yes & \Yes \tabularnewline
    \bottomrule
    \end{tabular}%
    \ifblue
    };
    \draw[blue, thick] (table.north west) rectangle (table.south east);
    \end{tikzpicture}
    \fi
    }
    \par
    \end{centering}
    \vskip-2ex
\end{table}

\noindent
\textbf{Robustness in IL.}
Robustness in \ac{IL} is needed to compensate for the distribution shifts caused by the \textit{sim2real} or \textit{lab2real} (i.e., when collecting demonstrations on the real robot in a controlled environment and then deploying in the real world) transfers. Robustness to these types of \textit{shifts} is achieved by modifying the training domain so that its dynamics match the ones encountered in the deployment domain ~\cite{peng2018sim, chebotar2019closing}. An extremely effective method is \ac{DR}~\cite{peng2018sim}, which applies random model errors/disturbances, sampled from a predefined set of possible perturbations, during data collection in simulation. %
An alternative avenue relies on modifying the actions of the expert to ensure that the state distribution visited at training time matches the one encountered at deployment time, such as in DART~\cite{laskey2017dart}. %
Although effective, these approaches require many demonstrations/interactions with the environment in order to take into account all the possible instantiations of model errors/disturbances that might be encountered in the target domain, limiting the opportunities for \textit{lab2real} transfers, or increasing the data collection effort when training in simulation. Our work will exploit extra information available to the \ac{MPC} to reduce the number of \ac{MPC}/environment interactions.

\noindent
\textbf{Data Augmentation for Efficient/Robust IL.}
\ac{GPS} \cite{levine2013guided, zhang2016learning, kahn2017plato, carius2020mpc, reske2021imitation}, introduced first the idea to use trajectories from model-based planners, including \ac{MPC}, to generate state-action samples (guiding samples) for improved sample efficiency in policy learning. Specifically, Ref.~\cite{levine2013guided} leveraged an \b{\ac{iLQR}} \cite{li2004iterative} expert to generate guiding samples around the optimal trajectory found by the controller. Similarly, the authors in ~\cite{carius2020mpc} observe that adding extra states and actions sampled from the neighborhood of the optimal solution found by the \b{\ac{iLQR}} expert can reduce the number of demonstrations required to learn a policy when using DAgger. However, while \ac{GPS} methods are in general more sample-efficient than \ac{IL}, the nominal plans and the distribution of guiding-samples they generate do not \textit{explicitly} account for model and environment uncertainties, resulting in policies with limited robustness. Ref.~\cite{zhang2016learning} for example, demonstrates in simulation robustness to up to $3.3\%$ in weight perturbations of a multirotor, while our approach demonstrates robustness to perturbations up to $30\%$.
Our work leverages a robust variant of \ac{MPC} called \ac{RTMPC} \cite{mayne2005robust, mayne2011tube}, to provide robust demonstrations and a \ac{DA} strategy that accounts for the effects of uncertainties. Specifically, the \ac{DA} strategy is obtained by using an outer-approximation of the robust control invariant set (\textit{tube}) as a support of the sampling distribution, ensuring that the guiding samples produce robust policies. 
This idea is related to the recent LAG-ROS framework~\cite{tsukamoto2021learning}, which provides a learning-based method to compress a global planner in a \ac{DNN} by extracting relevant information from the robust tube. LAG-ROS emphasizes the importance of nonlinear contraction-based controllers (e.g., CV-STEM~\cite{tsukamoto2020neural}) to obtain robustness and stability guarantees. 
Our contribution emphasizes instead minimal requirements - namely a tube and an \textit{efficient} \ac{DA} strategy - to achieve demonstration-efficiency and robustness to real-world conditions. By decoupling these aspects from the need for complex control strategies, our work greatly simplifies the controller design. Additionally, different from LAG-ROS, the \ac{DA} procedures presented in our work do not require solving a large optimization problem for every extra state-action sample generated (achieving computational efficiency during training) 
and can additionally leverage interactive experts (e.g., DAgger) to trade off the number of interactions with the environment with the number of extra samples from \ac{DA} (further improving training efficiency).  

Recent work \cite{krishnamoorthy2022sensitivity, krishnamoorthy2023improved} exploited local approximations of the solutions found when solving the \ac{NLP} associated with \ac{MPC} to efficiently generate extra state-actions samples for \ac{DA} in policy learning. Similar to our work, \cite{krishnamoorthy2022sensitivity} uses a parametric sensitivity-based approximation of the solution to efficiently generate extra states and actions. Different from our work, however, their method 
proposes sampling of the entire feasible state space to learn a policy, while our work focuses instead on task-relevant demonstrations, a more computationally and data-efficient solution. The recently presented extension \cite{krishnamoorthy2023improved} solves this issue by leveraging interactive experts (e.g., DAgger). However, both \cite{krishnamoorthy2022sensitivity, krishnamoorthy2023improved} do not \textit{explicitly} account for the effects of uncertainties, neither in the design of the expert, nor in the way that extra states are generated, resulting in policies with limited robustness. Thanks to our robust expert, our approach not only accounts for uncertainties during demonstration collection and in the distribution of samples for \ac{DA}, but it can additionally account for the errors introduced in the \ac{DA} procedure by further constraint tightening and updating the tube size. Additionally, thanks to the strong prior on the state distribution under uncertainty produced by the tube in \ac{RTMPC}, our \ac{DA} strategy can quickly cover the task-relevant parts of the state space, obtaining demonstration-efficiency.
Last, unlike prior work, we experimentally validate our approach, demonstrating it on a system whose models has a large state size (state size $8$ and $10$), whereas previous work focuses on lower-dimensional systems (state size $2$) and only in simulation.

\b{
\noindent
\textbf{Robustness and Computational Challenges in MPC for Agile Flight.} 
\ac{MPC} has been widely employed in the aerial robotics community \cite{nguyen2021model}, enabling impressive performance in trajectory tracking and minimum-time planning for agile flights, and particularly in drone racing \cite{sun2022comparative, foehn2021time, romero2022model}. However, the authors of \cite{sun2022comparative} highlight that one of the biggest drawbacks of \ac{MPC} is in its required computational resources, limiting its deployment on platforms with a small computational budget. In addition, they highlight that their \ac{MPC} tends to fail when subject to a large external force disturbance or model errors. Our work is motivated by these findings and employs robust variants of MPC that explicitly account for uncertainties, such as disturbances and model errors, while reducing the computational complexity of MPC. %
}
\b{Impressive agile flight has also been achieved by MPC with models learned offline~\cite{manzoor2022model, kaiser2018sparse} or online~\cite{saviolo2022physics, spielberg2021neural}, or with MPC combined with non-parametric adaptation laws \cite{9632352}. While our work does not directly tackle the numerous challenges associated with adaptation and model learning in MPC, we highlight that our approach can benefit these fields, as RTMPC can explicitly account for uncertainties in learned models and can account for the dynamics introduced by adaptation laws \cite{lopez2019adaptive}, reducing the  constraint violations observed in \cite{9632352}}.

\subsection{Contributions}
This article extends our prior conference paper \cite{tagliabue2022efficient}, where the focus was on efficiently generating robust trajectory tracking policies from a \textit{linear} \ac{MPC}. In this new work, we additionally provide a strategy to generate a \ac{DNN} policy to reach a desired state using a \textit{nonlinear} \ac{MPC} expert, presenting a new methodology that can be used to perform \ac{DA} in a computationally-efficient way. 
This extension is non-trivial, as the \textit{ancillary controller} in the nonlinear \ac{RTMPC} framework \cite{mayne2011tube} requires, unlike the linear case, expensive computations to generate extra actions for \ac{DA}, resulting in long training times when performing \ac{DA}. This new work solves the computational-efficiency issues in the ancillary controller of nonlinear \ac{RTMPC} by generating a sensitivity-based approximation of the ancillary controller that is used to more efficiently compute the actions corresponding to extra state samples for \ac{DA}. \b{While sensitivity-based approximations of traditional MPC were first explored in recent work \cite{krishnamoorthy2022sensitivity}, this work extends \cite{krishnamoorthy2022sensitivity} not only by (1) learning from demonstrations from a controller \cite{mayne2011tube} whose nominal plans account for uncertainties, but additionally 
(1.1) proposes a sampling strategy based on the tube in \cite{mayne2011tube}, used as support of the sampling distribution, rather than considering arbitrary neighborhoods of the state space \cite{krishnamoorthy2022sensitivity}, achieving robustness and data-efficiency;
(1.2) leverages both on-policy (DAgger) and off-policy (BC) data collection methods, enabling trade-offs in terms of performance of the learned policy versus ease of data collection; 
(1.3) presents a policy fine-tuning procedure to minimize the impact on performance introduced by the sensitivity-based approximation;
(1.4) leverages further constraint tightening (e.g., makes constraints more conservative) in \cite{mayne2011tube} to account for errors introduced by the approximate DA strategy, ensuring robustness. }
\noindent
Additionally, this work presents
(2) a formulation of nonlinear \ac{RTMPC} for acrobatic flights on multirotors; 
(3) numerical comparison to \ac{IL} baselines;
(4) numerical comparison of different tube-sampling strategies;
(5) new real-world experiments with policies that leverage nonlinear models.

In summary, our work presents the following \textbf{contributions:} 
\begin{itemize}

\item A procedure to \textit{efficiently} learn \textit{robust} policies from \ac{MPC}. Our procedure is:
    \begin{inparaenum}
    \item \textit{demonstration-efficient}, as it requires a small number of queries to the training environment, resulting in a method that enables learning from a single \ac{MPC} demonstration collected in simulation or on the real robot; 
    \item \textit{training-efficient}, as it reduces the number of computationally expensive queries to the computationally expensive \ac{MPC} expert using a computationally efficient \ac{DA} strategy; 
    \item \textit{generalizable}, as it produces policies robust to disturbances not experienced during training.
    \end{inparaenum}
\item We generalize the demonstration-efficient policy learning strategy proposed in our previous conference paper \cite{tagliabue2022efficient} with the ability to efficiently learn robust and generalizable policies from variants of \ac{MPC} that use nonlinear models. 

\item Extensive simulations and comparisons with state-of-the-art \ac{IL} methods and robustification strategies. 
\item Experimental evaluation on the challenging task of trajectory tracking and acrobatic maneuvers on a multirotor, presenting the first instance of sim2real transfer of a policy trained after a single demonstration, and robust to previously unseen real-world uncertainties. 
\end{itemize}

\section{Problem Statement} \label{sec:imitation}
This part describes the problem of learning a robust policy in a demonstration and computationally efficient way by imitating an \ac{MPC} expert demonstrator. Robustness and efficiency are determined by the ability to design an \ac{IL} procedure that can compensate for the covariate shifts induced by uncertainties encountered during real-world deployment while collecting demonstrations in a domain (the training domain) that presents only a subset of those uncertainty realizations. %
Our problem statement follows the one of robust \ac{IL} (e.g., DART \cite{laskey2017dart}), modified to use deterministic policies/experts and to account for the differences in uncertainties encountered in deployment and training domains. 
Additionally, we present a common approach employed to address the covariate shift issues caused by uncertainties, \ac{DR}, highlighting its limitations. 

\subsection{Assumptions and Notation}
\label{sec:transfer}
\noindent
\textbf{System Dynamics.} 
We assume the dynamics of the real system are Markovian and stochastic~\cite{sutton2018reinforcement}, and can be described by a twice continuously differentiable function $f(\cdot)$:
\begin{equation}
\label{eq:system_dynamics}
\vbf{x}_{t+1} = f(\vbf{x}_t, \vbf{u}_t) + \vbf{w}_t,
\end{equation}
where $\vbf{x}_t \in \mathbb{X} \subseteq \mathbb{R}^{n_x}$ represents the state, $\vbf{u}_t \in \mathbb{U} \subseteq \mathbb{R}^{n_u}$ the control input in the compact subsets $\mathbb{X}$, $\mathbb{U}$. $\vbf{w}_t \in \mathbb{W}_\anydomain \subset \mathbb{R}^{n_x}$ is an unknown state perturbation, belonging to a compact convex set $\mathbb{W}_\anydomain$ containing the origin. Stochasticity in \cref{eq:system_dynamics} is introduced by $\vbf{w}_t$, sampled from \b{a probability distribution having support} $\mathbb{W}_\anydomain$, under a (possibly unknown) probability \b{density function}, capturing the effects of noise, approximation errors in the learned policy, model changes, and other disturbances acting on the system during training or under real-world conditions at deployment.

\noindent
\textbf{Sim2Real and Lab2Real Transfer Setup.}
\b{
Three different environments/domains $\anydomain$ are considered: a training domain based on a simulation $\mathcal{S}_\text{sim}$ (where $\mathcal{S}$ denotes \textit{source}), a training domain based on the real robot in a controlled lab environment $\mathcal{S}_\text{lab}$, and a deployment domain $\mathcal{T}$ (\textit{target}). 
Mathematically, the three domains differ in their transition probabilities. 
In all the domains, we do not assume knowledge on density function from which $\vbf{w}_t$ is sampled, but we assume available prior knowledge of $\mathbb{W}_\mathcal{T}$, the support of the distribution (e.g, worst-case uncertainty realization) at deployment. This is a common assumption in robust control \cite{mayne2005robust, mayne2011tube}, where such knowledge can come from historical data, regulatory requirements, or can be assumed to match the physical limits of the robot. Additionally, we assume $\mathbb{W}_{\mathcal{S}_\text{lab}} \subset \mathbb{W}_\mathcal{T}$ and $\mathbb{W}_{\mathcal{S}_\text{sim}} \subset \mathbb{W}_\mathcal{T}$, representing the fact that training is usually performed in simulation or in a controlled/lab environment under some nominal model errors/disturbances, while at deployment a larger set of perturbations can be encountered. Note that for convenience we use $\mathcal{S}$ to denote both $\mathcal{S}_\text{sim}$ and $\mathcal{S}_\text{lab}$. 

}

\noindent
\textbf{\b{Tracking} MPC Expert.} We \b{consider a tracking} \ac{MPC} expert demonstrator that plans along an $N+1$-steps horizon. 
The expert is given the current state $\vbf{x}_t$, and \b{$\vbf{X}_t^\text{des} \in \mathbb{X}_{N_\text{des}}^\text{des} \coloneqq \{\vbf{x}^\text{des}_{0|t}, \dots, \vbf{x}_{N_\text{des}|t}^\text{des} | \vbf{x}_i^\text{des} \in \mathbb{R}^{n_x}\}$}, representing a desired state to be reached, or a state trajectory to be followed. Then, the \ac{MPC} expert generates control actions by solving an \ac{OC} problem of the form: 
\begin{equation}
\begin{split}
\bar{\mathbf{X}}_t^*, \bar{\mathbf{U}}_t^*
    \in \underset{\bar{\mathbf{X}}_t, \bar{\mathbf{U}}_t}{\text{argmin}} & 
       \: \b{J_{N}}(\bar{\mathbf{X}}_t, \bar{\mathbf{U}}_t, \mathbf{X}^\text{des}_t) \\
    \text{subject to} \:\: & \bar{\vbf{x}}_{0|t} = \vbf{x}_t, \\
    & \bar{\mathbf{x}}_{i+1|t} = f(\bar{\mathbf{x}}_{i|t}, \bar{\mathbf{u}}_{i|t}),  \\
    & \bar{\mathbf{x}}_{i|t} \in \mathbb{X}, \:\: \bar{\mathbf{u}}_{i|t} \in \mathbb{U}, \\
    & i = 0, ..., N-1.
\end{split}
\end{equation}
where \b{$J_N$ represents the cost to be minimized (where $N$ denotes the dependency on the planning horizon)}, and $\bar{\mathbf{X}}_t = \{\bar{\vbf{x}}_{0|t},\dots,\bar{\vbf{x}}_{N|t}\}$ and $\bar{\vbf{U}}_t = \{\bar{\vbf{u}}_{0|t},\dots,\bar{\vbf{u}}_{N-1|t}\}$ are sequences of states and actions along the planning horizon, where the notation $\bar{\vbf{x}}_{i|t}$ indicates the planned state at the future time $t+i$, as planned at the current time $t$. 
At every timestep $t$, given $\vbf{x}_t$, the control input applied to the real system is the first element of $\bar{\vbf{U}}^*_t$, resulting in an implicit deterministic control law (policy) that we denote as $\pi_{\vbs{\theta}^*}: \mathbb{X} \times \mathbb{X}^\text{des} \rightarrow \mathbb{U}$.

\noindent 
\textbf{\ac{DNN} Student Policy.}
As for the \ac{MPC} expert, we model the \ac{DNN} student policy as a deterministic policy $\pi_{\vbs{\theta}}$, with parameters $\vbs{\theta}$\b{, that does not necessarily belong to the same policy class as the expert.}
When considering trajectory tracking tasks, the policy takes as input the current state and the desired reference trajectory segment, $\pi_{\vbs{\theta}} : \mathbb{X} \times \mathbb{X}_{N+1}^\text{des} \rightarrow \mathbb{U}$. When considering the task of reaching a goal state, the policy takes as input the current state, the desired goal state and the current timestep $t \in \mathbb{I}_{\geq 0}$, $\pi_{\vbs{\theta}}: \mathbb{X} \times \mathbb{X}_0^\text{des} \times \mathbb{I}_{\geq 0} \rightarrow \mathbb{U}$.

\noindent
\textbf{Transition Probabilities}. 
We denote the state transition probability under $\pi_{\vbs{\theta}}$ in a domain $\anydomain$ for a given goal-reaching or trajectory tracking task
as $p_{\pi_{\vbs{\theta}}, \anydomain}(\vbf{x}_{t+1}|\vbf{x}_t)$. 
\b{The probability of collecting a $T$-(state, action) pairs trajectory $\boldsymbol{\xi}=\{(\vbf{x}_t, \vbf{u}_t)_{t=0}^{T-1} \}$}, given a policy $\pi_{\vbs{\theta}}$, depends on the deployment environment $\anydomain$:  
\begin{equation}
    p(\boldsymbol{\xi}|\pi_{\vbs{\theta}}, \anydomain) = p(\vbf{x}_0) \prod_{t=0}^{T-1} p_{\pi_{\vbs{\theta}}, \anydomain}(\vbf{x}_{t+1}|\vbf{x}_t),
\end{equation}
where $p(\vbf{x}_0)$ represents the initial state distribution. 

\noindent
\subsection{Robust Imitation Learning Objective}
The objective of robust \ac{IL}, following~\cite{laskey2017dart}, is to find parameters $\vbs{\theta}$ of $\pi_{\vbs{\theta}}$ that minimize a distance metric $\mathcal{L}(\vbs{\theta}, \vbs{\theta}^*|\boldsymbol{\xi})$ from the \ac{MPC} expert $\pi_{\vbs{\theta}^*}$: 
\begin{equation}
    \hat{\vbs{\theta}}^* = \text{arg}\min_{\vbs{\theta}} \mathbb{E}_{p(\boldsymbol{\xi}|\pi_{\vbs{\theta}}, \mathcal{T})}\mathcal{L}(\vbs{\theta}, \vbs{\theta}^*|\boldsymbol{\xi}).
    \label{eq:il_obj_target}
\end{equation}
This metric captures the differences between the actions generated by the expert $\pi_{\vbs{\theta}^*}$ and the action produced by the student $\pi_{\vbs{\theta}}$ \b{across the distribution of trajectories induced by the student policy $\pi_{\vbs{\theta}}$ in the perturbed domain $\mathcal{T}$, as denoted by $p(\boldsymbol{\xi}|\pi_{\vbs{\theta}}, \mathcal{T})$}.
The distance metric considered in this work is the \ac{MSE} loss:
\begin{equation}
\mathcal{L}(\vbs{\theta}, \vbs{\theta}^*|\boldsymbol{\xi}) = \frac{1}{T}\sum_{t=0}^{T-1}\| \pi_{\vbs{\theta}}(\vbf{x}^\text{in}_t) - \pi_{\vbs{\theta}^*}(\vbf{x}_t, \vbf{X}_t^\text{des})\|_2^2. \label{eq:il_loss}
\end{equation}
where $\vbf{x}^\text{in}_t = \{\vbf{x}_t, \vbf{X}_t^\text{des}\}$ for trajectory tracking tasks, and $\vbf{x}^\text{in}_t = \{\vbf{x}_t, \vbf{X}_t^\text{des}, t\}$ for go-to-goal-state tasks.

\noindent
\textbf{Covariate Shift due to Sim2real and Lab2real Transfer.}
Because in practice we do not have access to the target environment, the goal of Robust IL is to try to solve~\cref{eq:il_obj_target} by finding an approximation of the optimal policy parameters $\hat{\vbs{\theta}}^*$ using data from the source environment: 
\begin{equation}
    \hat{\vbs{\theta}}^* = \text{arg}\min_{\vbs{\theta}} \mathbb{E}_{p(\boldsymbol{\xi}|\pi_{\vbs{\theta}}, \mathcal{S})}\mathcal{L}(\vbs{\theta}, \vbs{\theta}^*|\boldsymbol{\xi}).
    \label{eq:il_obj_source}
\end{equation}
The way this minimization is solved depends on the chosen \ac{IL} algorithm. The performance of the learned policy in the target and source domains can be related via: 
\begin{equation}
\begin{multlined}
\hspace*{-0.45in}    \mathbb{E}_{p(\boldsymbol{\xi}|\pi_{\vbs{\theta}}, \mathcal{T})}\mathcal{L}(\vbs{\theta}, \vbs{\theta}^*|\boldsymbol{\xi}) = \\
    \underbrace{
    \mathbb{E}_{p(\boldsymbol{\xi}|\pi_{\vbs{\theta}}, \mathcal{T})}\mathcal{L}(\vbs{\theta}, \vbs{\theta}^*|\boldsymbol{\xi}) - 
    \mathbb{E}_{p(\boldsymbol{\xi}|\pi_{\vbs{\theta}}, \mathcal{S})}\mathcal{L}(\vbs{\theta}, \vbs{\theta}^*|\boldsymbol{\xi})}_{\text{covariate shift due to transfer}} \\ + 
    \underbrace{\mathbb{E}_{p(\boldsymbol{\xi}|\pi_{\vbs{\theta}}, \mathcal{S})}\mathcal{L}(\vbs{\theta}, \vbs{\theta}^*|\boldsymbol{\xi})}_{\text{\ac{IL} objective}},
\end{multlined}
\label{eq:}
\end{equation}
which clearly shows the presence of a covariate shift induced by the transfer. The last term corresponds to the objective minimized by performing \ac{IL} in $\mathcal{S}$. Attempting to solve \cref{eq:il_obj_target} by directly optimizing \cref{eq:il_obj_source} (e.g., via \ac{BC}~\cite{pomerleau1989alvinn}) offers no assurances of finding a policy with good performance in $\mathcal{T}$.

\subsection{Shift Compensation via Domain Randomization.}
\label{subsec:domain_randomization}
A well-known strategy to compensate for the effects of covariate shifts between source and target domain is \ac{DR}~\cite{peng2018sim}, which modifies the transition probabilities of the source $\mathcal{S}$ by trying to ensure that the trajectory distribution in the modified training domain $\mathcal{S}_\text{DR}$ matches the one encountered in the target domain: $p(\boldsymbol{\xi}|\pi_{\vbs{\theta}}, \mathcal{S}_\text{DR}) \approx p(\boldsymbol{\xi}|\pi_{\vbs{\theta}}, \mathcal{T})$.
This is done by applying perturbations to the robot during demonstration collection, sampling perturbations $\vbf{w} \in \mathbb{W}_\text{DR}$ according to some knowledge/hypotheses on their distribution $p_\mathcal{T}(\vbf{w})$ in the target domain~\cite{peng2018sim}, obtaining the perturbed trajectory distribution $p(\boldsymbol{\xi}|\pi_{\vbs{\theta}}, \mathcal{S}, \vbf{w})$. The minimization of \cref{eq:il_obj_target} can then be approximately performed by minimizing instead:
\begin{equation}
\label{eq:il_dr_modified_source}
    \mathbb{E}_{p_\mathcal{T}(\vbf{w})}[\mathbb{E}_{p(\boldsymbol{\xi}|\pi_{\vbs{\theta}}, \mathcal{S}, \vbf{w})}\mathcal{L}(\vbs{\theta}, \vbs{\theta}^*|\boldsymbol{\xi})].
\end{equation}
This approach, however, requires the ability to apply disturbances/model changes to the system, which may be unpractical e.g., in the \textit{lab2real} setting, and may require a large number of demonstrations due to the need to sample enough state perturbations $\vbf{w}$.

\section{Efficient Learning from Linear RTMPC} \label{sec:rtmpc_linear}

\label{sec:robust_tube_mpc}
In this Section, we present the strategy to efficiently learn robust policies from \ac{MPC} when the system dynamics in \cref{eq:system_dynamics} can be well approximated by a linear model of the form:
\begin{equation}
\label{eq:linearized_dynamics}
\vbf{x}_{t+1} = \vbf{A} \vbf{x}_t + \vbf{B} \vbf{u}_t + \vbf{w}_t.
\end{equation}
First, we present the Robust Tube variant of linear MPC, \ac{RTMPC}, that we employ to collect demonstrations (\cref{subsec:rtmpc_expert}). Then, we present a strategy that leverages information available from the \ac{RTMPC} expert to compensate for the covariate shifts caused by uncertainties and mismatches between the training and deployment domains (\cref{sec:sa}). Our strategy is based on a \ac{DA} procedure that can be combined with different \ac{IL} methods (on-policy, such as \ac{DAgger} \cite{ross2011reduction}, and off-policy, such as \ac{BC}, \cite{pomerleau1989alvinn}) for improved efficiency/robustness in the policy learning procedure. The \ac{RTMPC} expert is based on \cite{mayne2005robust} but with the objective function modified to track desired trajectories, as trajectory-tracking tasks will be the focus of the experimental evaluation of policies learned from this controller (\cref{sec:evaluation_linear}). 

\subsection{Trajectory Tracking \ac{RTMPC} Expert Formulation} \label{subsec:rtmpc_expert}
\ac{RTMPC} is a type of robust \ac{MPC} that regulates the system in \cref{eq:linearized_dynamics} while ensuring satisfaction of the state and actuation constraints $\mathbb X, \mathbb U$ regardless of the disturbances $\mathbf w \in \mathbb{W}_\mathcal{T}$.

\noindent
\textbf{Mathematical Preliminaries.} Let $\mathbb{A} \subset \mathbb{R}^{n}$ and $\mathbb{B} \subset \mathbb{R}^{n}$ be convex polytopes, and let $\mathbf{C} \in \mathbb{R}^{m \times n}$. Then we define:
\begin{enumerate}[a)]
\item Linear mapping: $\mathbf C \mathbb{A} \coloneqq \{\mathbf{C} \mathbf a \in \mathbb{R}^m \:|\: \mathbf a \in \mathbb{A}\}$
\item Minkowski sum: $\mathbb{A} \oplus \mathbb{B} \coloneqq \{\mathbf a + \mathbf b \in \mathbb{R}^n \:|\: \mathbf a \in \mathbb{A}, \: \mathbf b \in \mathbb{B} \}$
\item Pontryagin diff.: $\mathbb{A} \ominus \mathbb{B} \coloneqq \{\mathbf c \in \mathbb{R}^n \:|\: \mathbf{c + b} \in \mathbb{A}, \forall \mathbf  b \in \mathbb{B} \}$. 
\end{enumerate}

\textbf{Optimization Problem.} 
At each time step $t$, trajectory tracking \ac{RTMPC} receives the current robot state $\mathbf x_t$ and a desired trajectory $\mathbf{X}^\text{des}_t = \{\xdes_{0|t},\dots,\xdes_{N|t}\}$ spanning $N+1$ steps as input. It then computes a sequence of reference (``safe'') states $\bar{\mathbf{X}}_t = \{\xsafe_{0|t},\dots,\xsafe_{N|t}\}$ and actions $\bar{\mathbf{U}}_t = \{\usafe_{0|t},\dots,\usafe_{N-1|t}\}$ that ensure constraint compliance regardless of the realization of $\mathbf{w}_t \in \mathbb{W}_\mathcal{T}$. This  is achieved by solving the following \b{\ac{QP} (e.g., via the solver \cite{osqp})}:
\begin{align}
\label{eq:rtmpc_optimization_problem}
    \mathbf{\bar{U}}_t^*, \mathbf{\bar{X}}_t^* &= \underset{\mathbf{\bar{U}}_t, \mathbf{\bar{X}}_t}{\text{argmin}}
        \| \mathbf e_{N|t} \|^2_{\mathbf{P}_x} + 
        \sum_{i=0}^{N-1} 
            \| \mathbf e_{i|t} \|^2_{\mathbf{Q}_x} + 
            \| \mathbf u_{i|t} \|^2_{\mathbf{R}_u} \notag \\
    &\text{subject to} \:\:  \xsafe_{i+1|t} = \mathbf A \xsafe_{i|t} + \mathbf B \usafe_{i|t}, \\
    &\xsafe_{i|t} \in \mathbb{X} \ominus \mathbb{Z}, \:\: \usafe_{i|t} \in \mathbb{U} \ominus \mathbf{K} \mathbb{Z}, \notag\\
    &\vbf{x}_t \in \mathbb{Z} \oplus  \xsafe_{0|t}, \: i = 0, \dots, {N-1}  \notag
\end{align}
where $\mathbf e_{i|t} = \xsafe_{i|t} - \xdes_{i|t}$ is the tracking error. \b{The matrix $\mathbf{R}_u$ (positive definite) and $\mathbf{Q}_x$ (positive semi-definite)} define the trade-off between deviations from the desired trajectory and actuation usage, while $\| \mathbf e_{N|t} \|^2_{\mathbf{P}_x}$ is the terminal cost. $\mathbf{P}_x$ 
and $\mathbf K$ are obtained by formulating an infinite horizon optimal control LQR problem using $\mathbf A$, $\mathbf B$, $\mathbf{Q}_x$ and $\mathbf{R}_u$ and by solving the associated algebraic Riccati equation \cite{aastrom2021feedback}.
To achieve recursive feasibility, we ensure a sufficiently long prediction horizon is selected, as commonly practiced \cite{kamel2017model}, while omitting the inclusion of terminal set constraints.

\begin{figure}
    \centering
    \includegraphics[width=0.9\columnwidth, trim={0.3in, 1.2in, 0.3in, 7.2in}, clip]{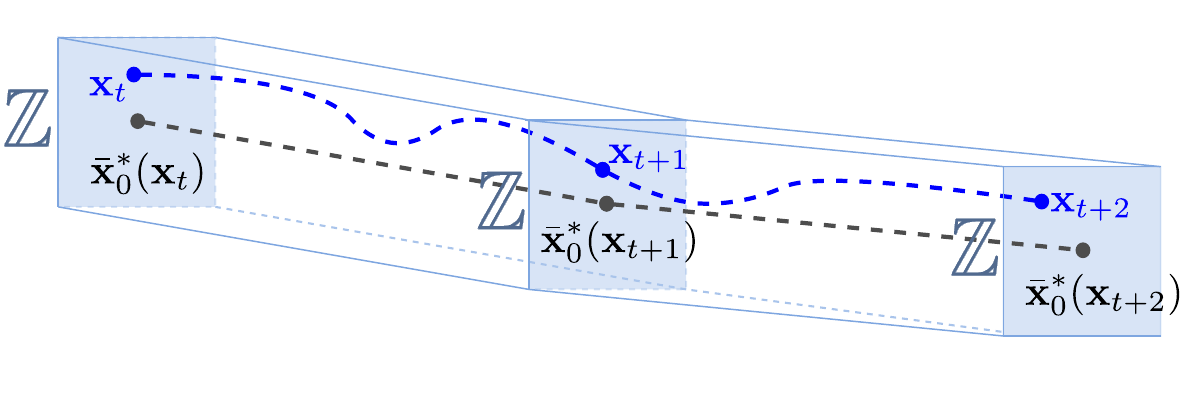}
    \caption{
    Illustration of the sequence of robust control invariant sets ${\mathbb{Z} \oplus \bar{\mathbf{x}}_0^*(\mathbf{x}_t)}$ computed by RTMPC for a system with state $\mathbf{x}_t$ and dimension $n_x = 2$.}
    \label{fig:tube_illustration}
\end{figure}

\noindent
\textbf{Tube and Ancillary Controller.}  A control input for the real system is generated by \ac{RTMPC} via an \textit{ancillary controller}:
\begin{equation}
\label{eq:ancillary_controller}
    \mathbf u_t = \usafe^*_{t} + \mathbf K (\mathbf{x}_t - \xsafe^*_{t}),
\end{equation}
where $\usafe^*_t = \usafe^*_{0|t}$ and $\xsafe^*_t = \xsafe^*_{0|t}$. 
As shown in \cref{fig:tube_illustration}, 
This controller ensures that the system remains inside a \textit{tube} (with ``cross-section'' $\mathbb{Z}$) centered around $\xsafe_t^*$ regardless of the realization of the disturbances in $\mathbb{W}_\mathcal{T}$, provided that the tube contains the initial state of the system (constraint $\vbf{x}_t \in \mathbb{Z} \oplus \xsafe_{0|t}$).
The set $\mathbb{Z}$ is a disturbance invariant set for the closed-loop system $\mathbf{A}_K := \mathbf{A + B K}$, satisfying the property that $\forall \mathbf{x}_j \in \mathbb{Z}$, $\forall \mathbf{w}_j \in \mathbb{W}_\mathcal{T}$, $\forall j \in \mathbb{N}^+$, $\mathbf{x}_{j+1} = \mathbf{A}_K \mathbf{x}_j + \mathbf{w}_j \in \mathbb{Z}$ \cite{mayne2005robust}. 
$\mathbb{Z}$ can be computed offline using $\mathbf{A}_K$ and the model of the disturbance $\mathbb{W}$ via ad-hoc analytic algorithms~\cite{borrelli2017predictive, mayne2005robust}, or can be learned from data~\cite{fan2020deep}. 
\b{Note that tracking aggressive trajectories may introduce large deviations from the operating points, resulting in linearization errors; these errors are treated as an additional source of process uncertainty when computing the tube. In addition, aggressive changes of the reference may result in infeasibility (e.g., when the terminal region is unreachable within the horizon, see \cite{limon2010robust}), which can be addressed, as typical in MPC, via an adequate choice of the planning horizon ($N=20$ or $N=30$ in our work).}

\subsection{Shift Compensation via Sampling Augmentation}
\label{sec:sa}
\b{
Training a policy by collecting demonstrations in a controlled source domain $\mathcal{S}$, with the objective of deploying it in a perturbed target domain $\mathcal{T}$ introduces a sample selection bias \cite{kouw2018introduction}, i.e., data is not collected around the distribution encountered in $\mathcal{T}$. Such bias is a known cause of distribution shifts \cite{kouw2018introduction}, and can be mitigated by re-weighting collected samples based on their likelihood of appearing in the target domain $\mathcal{T}$ via importance-sampling \cite{levine2013guided}. Importance-sampling, however, does not apply in our case, since we do not have access to samples/demonstrations collected in $\mathcal{T}$. 
 
In this work, distribution shifts are addressed by additionally utilizing the tube in \ac{RTMPC} to obtain knowledge of the states that the system may visit when subjected to perturbations in $\mathcal{T}$. Given this information, we propose a tube-guided \ac{DA} strategy, called \acf{SA}, that samples states from the tube and \textit{efficiently} computes corresponding actions via the ancillary controller in \ac{RTMPC}.} %

\noindent 
\textbf{Tube as a Model of State Distribution Under Uncertainties.} 
The key intuition of the proposed approach is the following. We observe that, although the density \b{function of} $p(\boldsymbol{\xi}|\pi_{\vbs{\theta}},\mathcal{T})$ is unknown, an approximation of its support $\mathfrak{R}$, given a demonstration $\boldsymbol{\xi}$ collected in the source domain $\mathcal{S}$, is known and corresponds to the tube in \ac{RTMPC} when collecting $\boldsymbol{\xi}$:
\begin{equation}
    \mathfrak{R}_{\boldsymbol{\xi}^+|\pi_{\vbs{\theta}^*},\boldsymbol{\xi}} = \b{\{\bar{\vbf{x}}_{t}^* \oplus \mathbb{Z}\}_{t=0}^{T-1}}.
    \label{eq:tube}
\end{equation}
where $\boldsymbol{\xi}^+$ is a trajectory in the tube of $\boldsymbol{\xi}$.
This is true thanks to the ancillary controller in \cref{eq:ancillary_controller}, which ensures that the system remains inside \cref{eq:tube} for every possible realization of $\vbf{w} \in \mathbb{W}_\mathcal{T}$.
The ancillary controller additionally provides a \textit{computationally efficient} way to obtain the actions to apply for every state inside the tube. Let $\vbf{x}_{t,j}^+ \in \bar{\vbf{x}}_t^* \oplus \mathbb{Z}$, i.e., $\vbf{x}_{t,j}^+$ is a state inside the tube computed when the system is at $\vbf{x}_t$, then the corresponding robust control action $\vbf{u}_{t,j}^+$ is:
\begin{equation}
\vbf{u}_{t,j}^+ = \bar{\vbf{u}}_t^* +  \vbf{K}(\vbf{x}_{t,j}^+ - \bar{\vbf{x}}_t^*).
\label{eq:tubempc_feedback_policy}
\end{equation}
For every timestep $t$ in $\boldsymbol{\xi}$, extra state-action samples $(\vbf{x}_{t,j}^+, \vbf{u}_{t,j}^+)$, with $j = 1, \dots, N_s$ collected from within the tube can be used to augment the dataset employed to train the policy, obtaining a way to approximate the expected risk in the domain $\mathcal{T}$ by only having access to demonstrations collected in $\mathcal{S}$: 
\begin{equation}
\begin{multlined}
    \mathbb{E}_{p(\boldsymbol{\xi}|\pi_{\vbs{\theta}}, \mathcal{T})}\mathcal{L}(\vbs{\theta}, \vbs{\theta}^*|\boldsymbol{\xi}) \approx \\
    \mathbb{E}_{p(\boldsymbol{\xi}|\pi_{\vbs{\theta}}, \mathcal{S})}[\mathcal{L}(\vbs{\theta}, \vbs{\theta}^*|\boldsymbol{\xi}) +  \mathbb{E}_{p(\boldsymbol{\xi}^+|\pi_{\vbs{\theta}^*},\boldsymbol{\xi})}\mathcal{L}(\vbs{\theta}, \vbs{\theta}^*|\boldsymbol{\xi}^+)].
    \label{eq:sampling_augmentation}
\end{multlined}
\end{equation}
\b{
\begin{algorithm} %
\small
\caption{\b{Sampling Augmentation (SA) for efficient learning from trajectory-tracking linear \ac{RTMPC}.}}
\label{alg:sa_lrtmpc}
\b{
\begin{algorithmic}[1]
\renewcommand{\COMMENT}[1]{\textcolor{black}{// #1}}
\renewcommand{\algorithmicrequire}{\textbf{Input:}}
\renewcommand{\algorithmicensure}{\textbf{Output:}}
\REQUIRE  $\mathbf{A}, \mathbf{B}, \mathbb{X}, \mathbb{U},\mathbf{Q}_x, \mathbf{R}_u,\mathbb{W}_\mathcal{T}, \beta, \mathcal{S}, \mathbf{X}^\text{des}$
\ENSURE Trained policy $\student{M}$
\STATE $\expert, \mathbf{K}, \hat{\mathbb{Z}} \leftarrow \text{DesignRtmpc}(\mathbf{A}, \mathbf{B}, \mathbb{X}, \mathbb{U}, \mathbf{Q}_x, \mathbf{R}_u, \mathbb{W}_\mathcal{T})$ \label{alg:sa_lrtmpc:init_expert}
\STATE  $\dset, \student{0} \leftarrow \emptyset, \text{InitializePolicy()}$ \label{alg:sa_lrtmpc:init_policy}
\FOR{$i = 1$ \TO $M$}
    \STATE $\dset \leftarrow \emptyset$ \COMMENT{optional}
    \FOR{$t = 0$ \TO $T-1$} \label{alg:sa_lrtmpc:collect_demo}
        \STATE $\vbf{u}^\text{RTMPC}_t, \bar{\vbf{x}}_t^*, \bar{\vbf{u}}_t^* \leftarrow \expert(\vbf{x}_t, \mathbf{X}_t^\text{des})$ \COMMENT{\cref{eq:rtmpc_optimization_problem} and \cref{eq:ancillary_controller}}

        \STATE $\dset \leftarrow \dset \cup \{ ( \vbf{x}_{t}, \vbf{X}_t^\text{des}, \vbf{u}^\text{RTMPC}_t ) \}$ \label{alg:sa_lrtmpc:store_demo}
        \FOR{$j=1$ \TO $N_\text{s}$}
            \STATE $\vbf{u}_{t,j}^+ = \bar{\vbf{u}}_t^* +  \vbf{K}(\vbf{x}_{t,j}^+ - \bar{\vbf{x}}_t^*)$, $\vbf{x}_{t,j}^+ \in \bar{\vbf{x}}_t^* \oplus \hat{\mathbb{Z}}$ \label{alg:sa_lrtmpc:ancillary_da} %
            \STATE $\dset \leftarrow \dset \cup \{(\mathbf{x}_{t,j}^+, \mathbf{X}_t^\text{des}, \mathbf{u}_{t,j}^+)\}$ 
            
        \ENDFOR
        \STATE $\mathbf{u}_t\!\! \leftarrow \beta_i \vbf{u}^\text{RTMPC}_t + (1-\beta_i) \; \student{i-1}\!\!(\vbf{x}_t, \mathbf{X}_t^\text{des})$ \COMMENT{DAgger/BC}
        
        \STATE $\vbf{x}_{t+1}\!\!\leftarrow \!\!\text{StepSystem}(\vbf{u}_t, \vbf{x}_t, \mathcal{S})$ \COMMENT{Sim./Physical Robot} 
    \ENDFOR

    \STATE $\student{i} \leftarrow \text{UpdatePolicy}(\dset, \hat{\vbs{\theta}}_{i-1})$ \label{alg:sa_lrtmpc:update_policy}
\ENDFOR
\end{algorithmic}
}  %
\end{algorithm}

}

\noindent
\textbf{Tube Approximation and Sampling Strategies.}
\begin{figure}
    \centering
    \includegraphics[width=0.7\columnwidth]{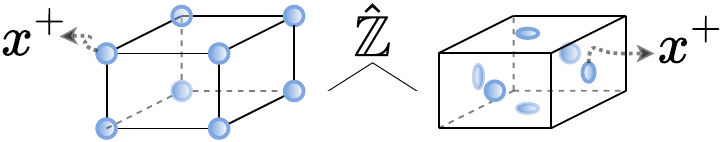}
    \caption{The possible strategies to sample extra state-action pairs from an axis-aligned bounding box, approximation of robust control invariant set of the \ac{RTMPC} expert: dense (left) and sparse (right). The diagram is for a system with state dimension $n_x = 3$.}
    \label{fig:tube_sampling_strategies} %
    \vspace*{-0.1in}
\end{figure}
In practice, the density $p(\boldsymbol{\xi}^+|\b{\pi_{\vbs{\theta}^*},\boldsymbol{\xi}})$ may not be available, making it difficult to establish which states to sample for \ac{DA}. We consider an adversarial approach to the problem by sampling states that may be visited under worst-case perturbations. To efficiently compute those samples, we (outer) approximate the tube $\mathbb{Z}$ with an axis-aligned bounding box $\hat{\mathbb{Z}}$. Note that an axis-aligned bounding box approximation is also used in the design of RTMPC for demonstration collection (\cref{eq:rtmpc_optimization_problem}). We investigate two strategies, shown in~\cref{fig:tube_sampling_strategies}, to obtain state samples $\vbf{x}_{t,j}^+$ at every state $\vbf{x}_t$ in $\boldsymbol{\xi}$:
\begin{inparaenum}[i)]
\item dense sampling: sample extra states from the vertices of $\bar{\vbf{x}}_t^*\oplus\hat{\mathbb{Z}}$. The approach produces $N_s = 2^{n_x}$ extra state-action samples. It is more conservative, as it produces more samples, but more computationally expensive.
\item sparse sampling: sample one extra state from the center of each \textit{facet} of $\bar{\vbf{x}}_t^* \oplus\hat{\mathbb{Z}}$, producing  $N_s = 2n_x$ additional state-action pairs. It is less conservative and more computationally efficient.
\end{inparaenum}

\b{\noindent 
\textbf{Algorithm Summary.} The procedure is summarized in \cref{alg:sa_lrtmpc}. 
First, SA designs the RTMPC expert according to the uncertainties in the target $\mathbb{W}_\mathcal{T}$ (line \ref{alg:sa_lrtmpc:init_expert})  and randomly initializes the student policy (line \ref{alg:sa_lrtmpc:init_policy}). Then, \ac{SA} collects in the source domain $\mathcal{S}$ a demonstration, using \ac{DAgger} or \ac{BC}, where $\beta_i$ is an hyperparameter of \ac{DAgger} controlling the probability of using actions from the expert and $\beta = 1$ corresponds to \ac{BC}, storing state and actions in the dataset $\mathcal{D}$ (line \ref{alg:sa_lrtmpc:store_demo}). The safe plan from the expert is then used to generate extra data via \cref{eq:tubempc_feedback_policy} (line \ref{alg:sa_lrtmpc:ancillary_da}), and the policy is updated (line \ref{alg:sa_lrtmpc:update_policy}, \cref{eq:il_obj_target} and \cref{eq:il_loss} using the data in $\dset$ and starting from the previous policy weights $\hat{\vbs{\theta}}_{i-1}$). The data collection and training procedure can be repeated across $M$ demonstrations.}

\section{Efficient Learning from Nonlinear RTMPC} \label{sec:rtmpc_nonlinear}
In this Section, we design an \ac{IL} and \ac{DA} strategy, which is an extension of the one presented in \cref{sec:rtmpc_linear}, that enables robust and efficient policy learning from an \ac{MPC} that employs nonlinear models of the form in \cref{eq:system_dynamics}. Different from \cref{sec:rtmpc_linear}, the focus here is on obtaining policies capable of reaching a desired goal state, as this will enable acrobatic maneuvers -- the scenario considered in the evaluation of policies learned from this controller (\cref{sec:evaluation_nonlinear}). %
To accomplish this, first, we use a nonlinear version of \ac{RTMPC}, based on \cite{mayne2011tube}, to collect demonstrations that account for the effects of uncertainties. This expert is summarized in \cref{subsec:rtnmpc_expert}. Second, we develop a computationally efficient tube-guided \ac{DA} strategy leveraging the ancillary controller of the nonlinear \ac{RTMPC} expert. Unfortunately, unlike in the linear \ac{RTMPC} case, nonlinear \ac{RTMPC} \cite{mayne2011tube} uses \ac{NMPC} as an ancillary controller. This limits the computational efficiency in \ac{DA}, as the generation of extra state-action samples requires solving a large \ac{NLP} associated with the ancillary \ac{NMPC} (discussed in \cref{subsec:ancillary_nonlinear_mpc}). We overcome this issue by presenting, in \cref{subsec:sensitivity}, a time-varying linear feedback law, approximation of the ancillary \ac{NMPC}, that enables efficient generation of the extra data leveraging the sensitivity of the control input to perturbations in the states visited during an initial demonstration collection procedure. 
Finally, in \cref{subsec:rob_perf_under_approx_samples}, we address the approximation errors introduced by the sensitivity-based \ac{DA} by presenting strategies to mitigate the gap, in performance and robustness, between the learned policy and the \ac{RTMPC} expert.

\subsection{Nonlinear RTMPC Expert Formulation} \label{subsec:rtnmpc_expert}
Nonlinear RTMPC \cite{mayne2011tube} ensures state and actuation constraint satisfaction while controlling a nonlinear, uncertain system of the form in \cref{eq:system_dynamics}. This controller operates by solving two \acp{OCP}, one to compute a nominal safe plan, and one to track the safe plan (ancillary \ac{NMPC}).

\noindent
\textbf{Nominal Safe Planner.} \label{subsec:nmpc_nominal_plan}
The first \ac{OCP}, given an $N+1$-steps planning horizon, generates nominal safe state and action open-loop plans 
$\vbf{Z}_{\tzr} = \{\vbf{z}_{0|\tzr}, \dots, \vbf{z}_{N|\tzr}\}, \vbf{V}_{\tzr} = \{\vbf{v}_{0|\tzr},\dots, \vbf{v}_{N-1|\tzr}\}$. 
The plans are open-loop because they are generated only at time $\tzr$, when the desired state and action equilibrium pair $\vbf{X}^\text{des}_{\tzr} = \{\vbf{x}_{\tzr}^e, \vbf{u}_{\tzr}^e \}$ for the nominal system changes. 
The nominal safe plan is obtained from: 
\begin{equation} \label{eq:nmpc_nominal}
\begin{split}
\vbf{V}_{\tzr}^*, \vbf{Z}_{\tzr}^*
    = \underset{\vbf{V}_{\tzr}, \mathbf{Z}_{\tzr}}{\text{argmin}} & 
       \: J_\text{RTNMPC}(\vbf{Z}_{\tzr}, \vbf{V}_{\tzr}, \vbf{X}_{\tzr}^\text{des}) \\
    \text{subject to} \:\: & \mathbf{z}_{i+1|\tzr} = f(\mathbf{z}_{i|\tzr}, \mathbf{v}_{i|\tzr}),  \\
    & \mathbf{z}_{i|\tzr} \in \bar{\mathbb{Z}}, \:\: \mathbf{v}_{i|\tzr} \in \bar{\mathbb{V}}, \\
    & \mathbf{z}_{0|\tzr} = \vbf{x}_{\tzr}, \:\: \mathbf{z}_{N|\tzr} = \vbf{x}_{\tzr}^e.
\end{split}
\end{equation}
$J_\text{RTNMPC} = \sum_{i=0}^{N-1} \| \vbf{z}_{i|\tzr} - \vbf{x}^e_{\tzr} \|^2_{\mathbf{Q}_z} + \| \vbf{v}_{i|{\tzr}} - \vbf{u}^e_{\tzr} \|^2_{\mathbf{R}_v}$, where $\mathbf{Q}_z$, $\mathbf{R}_v$ are positive definite. A key idea in this approach involves imposing modified state and actuation constraints $\bar{\mathbb{Z}} \subset \mathbb{X}$ and $\bar{\mathbb{V}} \subset \mathbb{U}$ so that the generated nominal safe plan is at a specific distance from state and actuation constraints. 
To be more precise, similar to the linear \ac{RTMPC} case (\cref{eq:rtmpc_optimization_problem}), the given state constraints $\mathbb{X}$ and actuation constraints $\mathbb{U}$ are tightened (made more conservative) by an amount that accounts for the spread of trajectories induced by the ancillary controller when the system is subject to uncertainties, obtaining $\bar{\mathbb{Z}} \subset \mathbb{X}$ and $\bar{\mathbb{V}} \subset \mathbb{U}$. \b{Such spread of trajectories corresponds to state and action tubes $\mathbb{T}^\text{state} \subset \mathbb{R}^{n_x}, \mathbb{T}^\text{action} \subset \mathbb{R}^{n_u}$ that contain the current nominal safe state and action trajectories $\mathbf{z}_{t|\tzr}^*$, $\mathbf{v}_{t|\tzr}^*$. Different from the linear case, however, analytically computing the tightened constraints and the tubes is challenging}. Fortunately, as highlighted in \b{\cite[Section 7]{mayne2011tube}}, accurately computing these sets is not needed, and an outer approximation is sufficient. This approximation can be obtained via Monte-Carlo simulations \cite{mayne2011tube} of the system under disturbances, or learned \cite{fan2020deep}; \b{the procedure employed in our work is tailored to our application domain, and is described in details in \cref{subsec:nonlinar_mpc}}  %
\b{Last we note that, as in \cite{mayne2011tube}, \cref{eq:nmpc_nominal} is assumed to be feasible.}

\noindent
\textbf{Ancillary \ac{NMPC}.}
The second \ac{OCP} corresponds to a trajectory tracking \ac{NMPC}, that acts as an ancillary controller, to maintain the state of the uncertain system close to the reference generated by \cref{eq:nmpc_nominal}. The \ac{OCP} is:
\begin{align} 
\label{eq:ancillary_nmpc_eq}
    \mathbf{\bar{U}}_t^*, \mathbf{\bar{X}}_t^*
    = \underset{\mathbf{\bar{U}}_t, \mathbf{\bar{X}}_t}{\text{argmin}} & 
        \| \mathbf e_{N|t} \|^2_{\mathbf{P}_x} \!\! + \!\!
        \sum_{i=0}^{N-1} 
            \| \mathbf e_{i|t} \|^2_{\mathbf{Q}_x} \!\!+\! 
            \| \bar{\vbf{u}}_{i|t}\!-\!\vbf{v}_{i + t|\tzr}^* \|^2_{\mathbf{R}_u} \notag \\
    \text{subject to} \:\: &  \xsafe_{i+1|t} = f(\xsafe_{i|t}, \usafe_{i|t})  \\ 
    & \xsafe_{0|t} = \vbf{x}_t, \usafe_{i|t} \in \b{\mathbb{U}} \notag 
\end{align}
where $\mathbf e_{i|t} = \xsafe_{i|t} - \vbf{z}^*_{i + t - t_0|\tzr}$ is the state tracking error. The positive definite matrices $\mathbf{Q}_x$ and $\mathbf{R}_u$ are tuning parameters, while $\mathbf{P}_x$ defines a terminal cost. Note that the terminal cost can be set as in \cite{mayne2011tube}, or using the solution for the infinite horizon Riccati equation for the linearized system associated with the state at the end of the planning horizon. However, owing to the fact that the expert can use a sufficiently long planning horizon without affecting onboard computation of the learned policy, in our experiments we set the terminal cost to $\mathbf{Q}_x$, additionally demonstrating that our approach introduces opportunities to simplify control design.
\cref{eq:ancillary_nmpc_eq} is solved at each timestep using the current state $\vbf{x}_t$, while the action applied to the robot is $\vbf{u}_t = \bar{\vbf{u}}^*_{0|t}$. %
We note that the ancillary \ac{NMPC} can have different tuning parameters than \cref{eq:nmpc_nominal} providing additional degrees of freedom to shape the response of the system under uncertainties.    %

A key result  \b{(presented in \cite[Section 5]{mayne2011tube})} of the employed nonlinear \ac{RTMPC} \cite{mayne2011tube} is that the ancillary \ac{NMPC} in \cref{eq:ancillary_nmpc_eq} maintains the trajectories of the uncertain system in \cref{eq:system_dynamics} inside state and action tubes $\mathbb{T}^\text{state}, \mathbb{T}^\text{action}$ that contain the current nominal safe state and action trajectories $\mathbf{z}_{t|\tzr}^*$, $\mathbf{v}_{t|\tzr}^*$ from the \ac{OCP} in \cref{eq:nmpc_nominal}. The state and action tubes are used to obtain the tightened state and actuation constraints $\bar{\mathbb{Z}}$, $\bar{\mathbb{V}}$, ensuring constraint satisfaction.  %

\noindent
\textbf{Solving the Ancillary \ac{NMPC}}
\label{subsec:ancillary_nonlinear_mpc}
A large portion of the computational cost of deploying or collecting demonstrations from nonlinear \ac{RTMPC} comes from the need to solve the \ac{OCP} of the ancillary NMPC (\cref{eq:ancillary_nmpc_eq}) at each timestep. In contrast, the \ac{OCP} of the nominal safe plan (\cref{eq:nmpc_nominal}) can be solved once per task (e.g., whenever the desired goal state $\vbf{X}^\text{des}_{\tzr}$ changes). A state-of-the-art method to solve the optimization in \cref{eq:ancillary_nmpc_eq} is \ac{SQP}, i.e., by repeatedly:
\begin{inparaenum}[i)]
\item linearizing the \ac{NLP} around a given linearization point; 
\item generating and solving a corresponding \ac{QP}, obtaining a refined linearization point for the next \ac{SQP} iteration.
\end{inparaenum}
While capable of producing high-quality solutions, \ac{SQP} methods incur large computational requirements due to computationally-expensive system linearizations, and solving the associated \ac{QP} one or more times per timestep.

\subsection{Computationally-Efficient Data Augmentation using the Parametric Sensitivities} \label{subsec:sensitivity} %
The tube $\mathbb{T}^\text{state}$ induced by the ancillary controller in \cref{eq:ancillary_nmpc_eq} identifies relevant regions of the state space for \ac{DA}, as it approximates the support of the state distribution under uncertainties, as discussed in \cref{sec:sa}. However, generating the corresponding extra action samples using \cref{eq:ancillary_nmpc_eq} can be very computationally inefficient, as it requires solving the associated \ac{SQP} for every extra state sample,
making \ac{DA} computationally impractical, and defeating our initial objective of designing \textit{computationally efficient} \ac{DA} strategies.  

In this work, \acf{SA}, is extended to efficiently learn policies from nonlinear \ac{RTMPC} by employing a time-varying, linear approximation of the ancillary \ac{NMPC} -- enabling efficient generation of extra state-action samples. Specifically, we observe that \cref{eq:ancillary_nmpc_eq} solves the implicit feedback law:
\begin{equation}
\label{eq:ancillary_nmpc_implicit}
\vbf{u}_t \! = \! \bar{\vbf{u}}_{0|t}^*(\vbs{\chi}_t) \! \coloneqq \! \kappa(\vbs{\chi}_t), \;  \vbs{\chi}_t \! \coloneqq \! \{\!\vbf{x}_t, \!t; \!\vbf{V}_{\tzr}^*, \vbf{Z}_{\tzr}^*\}
\end{equation}
where the current inputs are denoted $\vbs{\chi}_t$. Then, for each timestep of the trajectory collected during a demonstration in the source environment $\mathcal{S}$, with current ancillary \ac{NMPC} input $\tilde{\vbs{\chi}}_t = \{ \tilde{\vbf{x}}_t, \tilde{t}; \!\vbf{V}_{\tzr}^*, \vbf{Z}_{\tzr}^* \}$, we generate a local linear approximation of \cref{eq:ancillary_nmpc_implicit} by computing the first-order sensitivity of $\vbf{u}_t$ to the initial state $\vbf{x}_t$: %
\begin{equation} \label{eq:sensitivity_matrix}
\vbf{K}_{\tilde{\vbs{\chi}}_t} \!\!
\coloneqq 
\!
\left.
\frac{\partial \vbf{\bar{u}}_{0|t}^*}{\partial \vbf{x}_t} 
\right|_{{\vbs{\chi}}_t = \tilde{\vbs{\chi}}_t}
\!\!\!\!\! = \!\!
\begin{bmatrix}
\left.\dfrac{\partial{\bar{\vbf{u}}_{0|t}^*}}{\partial{\left[ \vbf{x}_t \right]_{1}}}
\right|_{\tilde{\vbs{\chi}}_t},\!\!& 
\!\!
\dots, 
\!\!
& \!\!
\left.
\dfrac{\partial{\bar{\vbf{u}}_{0|t}^*}}{\partial{[\vbf{x}_t]_{n_x}}}
\right|_{\tilde{\vbs{\chi}}_t}
\\
\end{bmatrix}.
\end{equation}
The sensitivity matrix $\vbf{K}_{\tilde{\vbs{\chi}}_t} \in \mathbb{R}^{n_u \times n_x}$, 
enables us to compute extra actions $\vbf{u}_{t,j}^+$ from states inside the tube $\vbf{x}_{t,j}^+ \in \mathbb{T}^\text{state}$, with $j = 1, \dots, N_s$, sampled from the tube: 
\begin{equation}
\label{eq:approximate_ancillary_controller}
    \vbf{u}_{t,j}^+ = \vbf{\bar{u}}_{0|t}^* + \vbf{K}_{\tilde{\vbs{\chi}}_t} (\vbf{x}_{t,j}^+ - \vbf{\bar{x}}_{0|t}^*) \coloneqq \hat{\kappa}(\vbf{x}_{t,j}^+, \tilde{\vbs{\chi}}_t).
\end{equation}
The \ac{DA} procedure enabled by this approximation is computationally-efficient, as we do not need to solve an \ac{SQP} for each extra state-action sample $(\vbf{x}^+_{t,j}, \vbf{u}^+_{t,j})$ generated for \ac{DA}, and we only need to compute, once per timestep, the sensitivity matrix $\vbf{K}_{\tilde{\vbs{\chi}}_t}$.
Note that the linearization points of \cref{eq:sensitivity_matrix} are based on the trajectory $\vbs{\xi}$ executed during demonstration collection. in the source environment $\mathcal{S}$.
We remark, additionally, that the actions computed when collecting demonstrations are obtained by solving the entire \ac{SQP}, and the sensitivity-based approximation is used only for \ac{DA}. 

\noindent
\textbf{Sensitivity Matrix Computation.} As described in \cite[\S 8.6]{rawlings2017model}, an expression to compute the sensitivity matrix in \cref{eq:sensitivity_matrix} (also called \textit{tangential predictor}) can be obtained by re-writing the \ac{NLP} in \cref{eq:ancillary_nmpc_eq} in a parametric form $\mathfrak{p}(\left[ \vbf{x}_t \right]_i)$, highlighting the dependency on scalar parameter representing the $i$-th component of the initial state $\vbf{x}_t$ (part of $\vbs{\chi}_t$). The parametric \ac{NLP} $\mathfrak{p}_{\mathcal{X}_t}(\left[ \vbf{x}_t \right]_i)$ is: %
\begin{equation}
\begin{split}
\label{eq:nmpc_opt_problem} 
\vspace{-.5in}
\underset{\vbf{y}}{\text{min}} & \: F_{\vbs{\chi}_t}(\vbf{y}) \\
    \text{subject to} \:\: & G_{\vbs{\chi}_t}(\left[ \vbf{x}_t \right]_i, \vbf{y}) = \vbs{0} \\
    & H(\vbf{y}) \leq \vbs{0},
\end{split}
\end{equation}
where $\vbf{y} \in \mathbb{R}^{n_{\vbf{y}}}$ corresponds to the optimization variables in \cref{eq:ancillary_nmpc_eq}, and $F_{{\vbs{\chi}}_t}(\cdot), G_{{\vbs{\chi}}_t}(\cdot), H(\cdot)$ are, respectively, the objective function, equality, and inequality constraints in \cref{eq:ancillary_nmpc_eq}, given the current state and reference trajectory in $\vbs{\chi}_t$. Additionally, we denote the solution of \cref{eq:nmpc_opt_problem} at $\tilde{\vbs{\chi}}_t$ (computed during the collected demonstration) as $(\tilde{\vbf{y}}^*, \tilde{\vbs{\lambda}}^*, \tilde{\vbs{\mu}}^*)$, where $\tilde{\vbs{\lambda}}^*, \tilde{\vbs{\mu}}^*$ are, respectively, the Lagrange multipliers for the equality and inequality constraints at the solution found. Then, each $i$-th column of the sensitivity matrix (\cref{eq:sensitivity_matrix}) can be computed by solving the \ac{QP}~(~\cite[Th. 8.16]{rawlings2017model}), \b{denoted $\mathfrak{p}_{\mathcal{X}_t,L}(\left[ \vbf{x}_t \right]_i)$}: 
\begin{align}
\underset{\mathbf{y}}{\text{min}} \quad& \hspace{-.1in} F_{\vbs{\chi}_t,L}(\mathbf{y}; \tilde{\mathbf{y}}^*) 
+ \frac{1}{2}(\mathbf{y} - \tilde{\mathbf{y}}^*)^\top \nabla^2_{\mathbf{y}} \mathscr{L}(\tilde{\mathbf{y}}^*, \tilde{\boldsymbol{\lambda}}^*, \tilde{\boldsymbol{\mu}}^*)(\mathbf{y} - \tilde{\mathbf{y}}^*) \notag \\
\text{s.t.} \quad & G_{\vbs{\chi}_t,L}([\mathbf{x}_t]_i, \mathbf{y}; \tilde{\mathbf{y}}^*) = \mathbf{0} \label{eq:tangential_predictor_qp} \\
& H_L(\mathbf{y}; \tilde{\mathbf{y}}^*) \leq \mathbf{0} \notag
\end{align}
where $F_{\vbs{\chi}_t,L}(\cdot; \tilde{\vbf{y}}^*)$, $G_{\vbs{\chi}_t,L}(\cdot; \tilde{\vbf{y}}^*)$, $H_L(\cdot; \tilde{\vbf{y}}^*)$ denote the respective functions in \cref{eq:nmpc_opt_problem} linearized at the solution found. $\nabla^2_{\vbf{y}}\mathscr{L}$ denotes the Hessian of the Lagrangian associated with \cref{eq:nmpc_opt_problem}, while the parameter is perturbed (e.g., $\left[ \vbf{x}_t \right]_i \leftarrow \left[ \vbf{x}_t \right]_i+1$). The $i$-th column of the sensitivity matrix can be extracted from the entries of $\vbf{y}^*$, solution of \cref{eq:tangential_predictor_qp}, at the position corresponding to $\bar{\vbf{u}}_{0|t}$.
We highlight that \cref{eq:tangential_predictor_qp} can be computed efficiently, as it leverages the latest internal linearization of the \ac{KKT} conditions performed in the \ac{SQP} employed to solve \cref{eq:ancillary_nmpc_eq}, and therefore it does not require to re-execute the computationally expensive system linearization routines that are carried out at each \ac{SQP} iteration.
We note that this local approximation exists when the assumptions in \cite[Th. 8.15]{rawlings2017model} 
are satisfied, i.e., that the solution $(\tilde{\vbf{y}}^*, \tilde{\vbs{\lambda}}^*, \tilde{\vbs{\mu}}^*)$ found during demonstration collection is a strongly regular \ac{KKT} point, and satisfies strict complementary conditions. 
Last, extra samples are generated using \cref{eq:approximate_ancillary_controller} under the assumption that the set of active inequality constraints (i.e., the index set $p \in \{1, \dots, n_H\}$ such that $[H(\tilde{\vbs{y}}^*)]_p = 0$) does not change.

\noindent
\textbf{Generalized Tangential Predictor.} A strategy that applies to the cases where strict complementary conditions do not hold, or where the extra state samples cause a change in the active set of constraints, is based on the \textit{generalized tangential predictor} \cite[\S 8.9.1]{rawlings2017model}. 
This predictor can be obtained by solving the \ac{QP} in \cref{eq:tangential_predictor_qp} with the set of equality constraints modified to be $G_{\vbs{\chi}_t,L}(\vbf{x}_{t,j}^+, \vbf{y}; \tilde{\vbf{y}}) = \vbf{0}$ \cite[Eq. 8.60]{rawlings2017model}. %
Although this approach requires solving a \ac{QP} to compute the action $\vbf{u}_{t,j}^+$ corresponding to each state $\vbf{x}_{t,j}^+$ sampled from the tube, it does not require re-generating the computationally expensive linearization performed at each \ac{SQP} iteration (and other performance optimization routines, such as condensing \cite{rawlings2017model}) nor solving the entire \ac{SQP} for multiple iterations -- resulting in a much more computationally-efficient procedure than solving the entire $\ac{SQP}$ \textit{ex-novo}. We remark that the linearization point in \cref{eq:tangential_predictor_qp} is updated at every timestep when a full \ac{SQP} is solved for demonstration-collection.

\subsection{Robustness and Performance Under Approximate Samples} \label{subsec:rob_perf_under_approx_samples}
While the described sensitivity-based \ac{DA} strategy enables the efficient generation of extra state-action samples, it introduces approximation errors that may affect the performance and robustness of the learned policy. Here, we discuss strategies to account for these errors, reducing the gaps between the nonlinear \ac{RTMPC} expert and the learned policy in terms of robustness and performance.

\noindent
\textbf{Robustness.}
A key property of \ac{RTMPC} is the ability to explicitly account for uncertainties, including the ones introduced by the proposed sensitivity-based \ac{DA} framework, by further tightening state and actuation constraints for the nominal safe plan (\cref{eq:nmpc_nominal}). The general nonlinear formulation of the dynamics in \cref{eq:system_dynamics}, however, makes it challenging to compute an \textit{exact} additional tightening bound for state and actuation constraints. A possible avenue to establish a tightening procedure for the actuation constraints is to observe that the linear approximation of \cref{eq:ancillary_nmpc_implicit} introduces an error upper bounded by~(\cite[Th. 8.16]{rawlings2017model}):
\begin{equation}
\| \kappa(\vbs{\chi}_t) - \hat{\kappa}(\!\vbf{x}^+_{t,j}, \vbs{\chi}_t)\|  \leq D \|\vbf{x}^+_{t,j} - \vbf{x}_t\|^2
 \end{equation}
where $D$ may be obtained by considering the Lipschitz constant of the controller (e.g., \cite{krishnamoorthy2022sensitivity}). However, estimating this constant may be difficult or computationally expensive for large-dimensional systems, as is the case herein. 
An alternative is to update the tubes as was done in \cref{subsec:nmpc_nominal_plan}, e.g., by employing Monte-Carlo simulations of the closed-loop system, starting from an initial (possibly conservative) tightening guess and by iteratively adjusting the cross-section (size) of the tube, or by directly learning the tubes from simulations or previous (conservative) real-world deployments \cite{fan2020deep}. These iterative procedures are particularly appealing in our context, as our efficient policy learning methodologies enable rapid training/updates of the learned policy, and the computational efficiency of the policy enables rapid numerical validations.

\noindent
\textbf{Performance Improvements via Fine-Tuning.}
In the context of learning policies from nonlinear \ac{RTMPC}, we include in \ac{SA} an (optional) fine tuning-step. This fine-tuning step consists in training the policy with additional demonstrations, without \ac{DA}, therefore avoiding introducing further approximate samples, and having discarded the extra data used to train the policy after an initial demonstration. Therefore, tube-guided \ac{DA} is treated as a methodology to efficiently generate an initial guess of the policy parameters.

\begin{algorithm}
\small
\caption{\b{Sampling Augmentation for efficient learning from Nonlinear \ac{RTMPC}}}
\label{alg:sa_nrtmpc}
\begin{algorithmic}[1]
\renewcommand{\COMMENT}[1]{\textcolor{black}{// #1}}
\renewcommand{\algorithmicrequire}{\textbf{Input:}}
\renewcommand{\algorithmicensure}{\textbf{Output:}}
\b{
\REQUIRE $f(\cdot), \mathbb{X}, \mathbb{U}, \mathbf{Q}_x, \mathbf{R}_u, \mathbb{W}_\mathcal{T}, \beta, \mathcal{S}$, $\vbf{X}^\text{des}_{\tzr}$
\ENSURE Trained policy $\student{M+L}$
\STATE $\expert, \mathbb{T}^\text{states} \leftarrow \text{DesingNRtmpc}(f(\cdot), \mathbb{X}, \mathbb{U}, \mathbf{Q}_x, \mathbf{R}_u, \mathbb{W}_\mathcal{T}, \vbf{X}^\text{des}_{\tzr})$
\STATE $\vbf{Z}_{\tzr}^*, \vbf{V}_{\tzr}^* \leftarrow \text{GetNominalSafePlan}(\expert)$ \COMMENT{\cref{eq:nmpc_nominal}} \label{eq:sa_nrtmpc:precompute_safe_plan}
\STATE $\kappa \leftarrow \text{GetAncillaryNmpc}(\expert)$ \COMMENT{\cref{eq:ancillary_nmpc_eq}, \cref{eq:ancillary_nmpc_implicit}} \label{eq:sa_nrtmpc:get_ancillary} 
\FOR{$i = 1$ \TO $M$}
    \FOR{$t = 0$ \TO $T$}
        \STATE $\mathcal{X}_t \leftarrow (\!\vbf{x}_t, \!t; \!\vbf{V}_{\tzr}^*, \vbf{Z}_{\tzr}^*)$ \COMMENT{Current operating point} \label{alg:sa_nrtmpc:collect_demo_init}
        \STATE $\vbf{u}^\text{N-RTMPC}_t, \mathfrak{p}_{\mathcal{X}_t,L}\!\!\leftarrow \!\kappa(\mathcal{X}_t)$ \COMMENT{\cref{eq:ancillary_nmpc_eq}, save QP~\cref{eq:tangential_predictor_qp}} 
        \STATE $\dset \leftarrow \dset \cup \{ (\vbf{x}_{t}, \vbf{X}^\text{des}, t, \vbf{u}^\text{N-RTMPC}_t ) \}$
        \STATE $\mathbf{K}_{\mathcal{X}_t} \leftarrow \text{Sensitivity}(\mathfrak{p}_{\mathcal{X}_t,L})$ \COMMENT{\cref{eq:sensitivity_matrix}}
        \FOR{$j=1$ \TO $N_\text{s}$} \label{alg:sa_nrtmpc:da_init}
            \STATE $\vbf{u}_{t,j}^+ = \bar{\vbf{u}}_t^* +  \mathbf{K}_{\mathcal{X}}(\vbf{x}_{t,j}^+ - \bar{\vbf{x}}_t^*)$, $\vbf{x}_{t,j}^+ \in \bar{\vbf{x}}_t^* \oplus \mathbb{T}^\text{states}$
            \STATE $\dset \leftarrow \dset \cup \{(\mathbf{x}_{t,j}^+, \mathbf{X}_t^\text{des}, \mathbf{u}_{t,j}^+)\}$
        \ENDFOR \label{alg:sa_nrtmpc:da_end}
        \STATE $\mathbf{u}_t\!\! \leftarrow \beta_i \vbf{u}^\text{N-RTMPC}_t + (1-\beta_i) \; \student{i-1}\!\!(\vbf{x}_t, \mathbf{X}_t^\text{des})$ \COMMENT{DAgger/BC}
        \STATE $\vbf{x}_{t+1}\!\!\leftarrow \!\!\text{StepSystem}(\vbf{u}_t, \vbf{x}_t, \mathcal{S})$ \label{alg:sa_nrtmpc:collect_demo_end}
    \ENDFOR
    \STATE $\student{i} \leftarrow \text{UpdatePolicy}(\dset, \hat{\mathcal{\theta}}_{i-1})$ \label{alg:sa_nrtmpc:train_policy}
\ENDFOR
\IF{FineTuning} \label{alg:sa_nrtmpc:fine_tuning}
    \STATE $\dset \leftarrow \emptyset$ \label{alg:sa_nrtmpc:discard_data}
    \FOR{$l=1$ \TO $L$}
        \STATE $\vbs{\xi} = \{ (\vbf{x}_t, \vbf{X}^\text{des}_{\tzr}, \vbf{u}_t^\text{N-RTMPC})) \}_{t=0}^{T-1} 
        \newline \hspace{100pt}\!\leftarrow\!\text{\text{CollectDemo}}(\kappa,\!\vbf{Z}_{\tzr}^*,\!\vbf{V}_{\tzr}^*,\!\pi_{\theta_{M+l-1}},\!\beta_i,\!\mathcal{S})$\COMMENT{DAgger/BC}\label{alg:sa_nrtmpc:collect_demo_finetune}
        \STATE $\dset \leftarrow \dset \cup \{ \vbs{\xi} \}$
        \STATE $\student{M+l} \! \leftarrow \! \text{UpdatePolicy}(\dset, \hat{\mathcal{\theta}}_{M+l-1})$ \label{alg:sa_nrtmpc:sa_nrtmpc:update_policy_fine_tuning}
    \ENDFOR
\ENDIF 
} %
\end{algorithmic}
\end{algorithm}

\noindent
\b{\textbf{Algorithm Summary.} The \ac{SA} procedure for nonlinear \ac{RTMPC} with the fine-tuning step \b{is summarized in \cref{alg:sa_nrtmpc}}. It consists of the following: 
\begin{enumerate}[1)]
\item Pre-compute the safe plan from the expert (line \ref{eq:sa_nrtmpc:precompute_safe_plan}).
\item Collect a single ($M=1$) task demonstration $\vbs{\xi}$ that tracks the safe plan using the ancillary \ac{NMPC} (line \ref{alg:sa_nrtmpc:collect_demo_init}-\ref{alg:sa_nrtmpc:collect_demo_end}), while additionally storing the variables of the \ac{QP} in \cref{eq:tangential_predictor_qp}.
\item Perform \ac{DA} using the parametric sensitivity (\cref{subsec:sensitivity}, line \ref{alg:sa_nrtmpc:da_init}-\ref{alg:sa_nrtmpc:da_end}, shown for the case where no active change of constraints occurs and strict complementary conditions hold, else use \cref{eq:tangential_predictor_qp}) and train the policy, obtaining the parameters $\hat{\vbs{\theta}}_1$ (line \ref{alg:sa_nrtmpc:train_policy}).
\item Optional \textit{fine-tuning} step \b{(line \ref{alg:sa_nrtmpc:fine_tuning})}:
\begin{enumerate}[i)]
\item Discard the collected data so far, including the data generated by the \ac{DA} (line \ref{alg:sa_nrtmpc:discard_data}).
\item Collect new demonstrations using DAgger\cite{ross2011reduction} and the pre-trained policy, or \ac{BC}, line \ref{alg:sa_nrtmpc:sa_nrtmpc:update_policy_fine_tuning}, and re-train the pre-trained policy (with parameters $\hat{\vbs{\theta}}_1$) after every newly collected demonstration.
\end{enumerate}
\end{enumerate}}

\section{Application to Agile Flight} \label{sec:agile_flight}
In this Section, we tailor the proposed efficient policy learning strategies to agile flight tasks, as this will be the focus of our numerical and experimental evaluation. First, in \cref{subsec:mav_model}, we present the nonlinear model of the multirotor used to collect demonstrations in simulation. Then, in \cref{subsec:linear_mpc}, we present a \ac{RTMPC} expert for \textit{trajectory tracking} based on a \textit{linear} multirotor model and that will be used with the \ac{IL} procedure described in \cref{sec:rtmpc_linear}. Because the considered trajectories require the robot to operate around a fixed, pre-defined condition (near hover), a hover-linearized model is suitable for the design of this controller. Last, in \cref{subsec:nonlinar_mpc}, we design a nonlinear \ac{RTMPC} expert capable of performing a $360^\circ$ flip in \textit{near-minimum time} - a maneuver that demands exploitation of the full \textit{nonlinear} dynamics of the multirotor, and that requires large and careful actuation usage; this controller is used with the learning in \cref{sec:rtmpc_nonlinear}. 

\subsection{Nonlinear Multirotor Model} \label{subsec:mav_model}
We consider an inertial reference frame $\text{W}$ attached to the ground, and a non-inertial frame $\text{B}$ attached to the \ac{CoM} of the robot. The translational and rotational dynamics of the multirotor are:  
\begin{subequations} \label{eq:mav_model_full}
\begin{align}
    \vbsd[W]{p} & = \vbs[W]{v} \label{eq:mav_model_full:tr_kin} \\
    \vbsd[W]{v} & = m^{-1}(\vbs{R}_\text{WB} \vbs[B]{t}_\text{cmd} + \vbs[W]{f}_\text{drag} + \vbs[W]{f}_\text{ext}) - \vbs[W]{g} \label{eq:mav_model_full:tr_dyn} \\
    \vbsd{q}_\text{WB} & = \frac{1}{2} \vbs{\Omega} (\vbs[B]{\omega}) \vbs{q}_\text{WB}  \label{eq:mav_model_full:rot_kin} \\
    \vbsd[B]{\omega} &  = \vbs{I}_\text{mav}^{-1}(- \vbs[B]{\omega} \times \vbs{I}_\text{mav} \vbs[B]{\omega} + \vbs[B]{\tau}_\text{cmd} + \vbs[B]{\tau}_\text{drag}) \label{eq:mav_model_full:rot_dyn}
\end{align}
\end{subequations}
where $\boldsymbol{p}$, $\boldsymbol{v}$, $\boldsymbol{q}$, $\boldsymbol{\omega}$ are, respectively, position, velocity, attitude quaternion and angular velocity of the robot, with the prescript denoting the corresponding reference frame. The attitude quaternion $\boldsymbol{q} = [q_w, \boldsymbol{q}_v^\top]^\top$ consists of a scalar part $q_w$ and a vector part $\boldsymbol{q}_v = [q_x, q_y, q_z]^\top$ and it is unit-normalized; the associated $3 \times 3$ rotation matrix is $\boldsymbol{R} = \vbs{R}(\vbs{q})$, while
\begin{equation}
\vbs{\Omega} (\boldsymbol{\omega}) = 
\begin{bmatrix}
0 & -\boldsymbol{\omega}^\top \\
\boldsymbol{\omega} & \lfloor \boldsymbol{\omega} \rfloor_\times \\
\end{bmatrix},
\end{equation} with $\lfloor \boldsymbol{\omega} \rfloor_\times$ denoting the $3 \times 3$ skew symmetric matrix of $\boldsymbol{\omega}$. $m$ denotes the mass, $\vbs{I}_\text{mav}$ the $3 \times 3$ diagonal inertial matrix, and $\boldsymbol{g} = [0, 0, g]^\top$ the gravity vector. Aerodynamic effects are taken into account via $\boldsymbol{f}_\text{drag} = - c_{D,1} \boldsymbol{v} - c_{D,2} \|\boldsymbol{v}\| \boldsymbol{v}$ and isotropic drag torque $\boldsymbol{\tau} = - c_{D,3} \boldsymbol{\omega}$, capturing the parasitic drag produced by the motion of the robot. The robot is additionally subject to external force disturbances $\boldsymbol{f}_\text{ext}$, such as the one caused by wind or by an unknown payload. 
Last, $\boldsymbol{t}_\text{cmd} = [0, 0, t_\text{cmd}]^\top$ is the commanded thrust force, and $\boldsymbol{\tau}_\text{cmd}$ the commanded torque. These commands can be mapped to the desired thrust $f_{\text{prop},i}$ for the $i$-th propeller ($i = 1, \dots, n_p$) via a linear mapping (\textit{allocation} matrix) $\boldsymbol{\mathcal{A}}$: 
\begin{equation}
\label{eq:allocation_matrix}
    \begin{bmatrix}
    t_\text{cmd} \\
    \vbs{\tau}_\text{cmd} \\ 
    \end{bmatrix}
    = \boldsymbol{\mathcal{A}}
    \begin{bmatrix}
    f_{\text{prop},1} \\
    \vdots \\
    f_{\text{prop}, n_p}
    \end{bmatrix} = \boldsymbol{\mathcal{A}} \vbs{f}_\text{prop}.
\end{equation}
The attitude of the quadrotor is controlled via the geometric attitude controller in \cite{lee2011geometric}. This controller generates desired torque commands $\vbs[B]{\tau}_\text{cmd}$ given a desired attitude $\vbs{R}_\text{WB}^\text{des}$, angular velocity $\vbs[B]{\omega}^\text{des}$ and acceleration $\vbsd[B]{\omega}^\text{des}$ \b{via\cite{lee2011geometric}: 
\begin{equation}
\small
\label{eq:mav_reduced_model_for_nmpc}
\begin{split}
    \vbs[B]{\tau}_\text{cmd} = & -\mathbf{K}_R \mathbf{e}_R - \mathbf{K}_\omega \mathbf{e}_\omega + \vbs[B]{\omega} \times \mathbf{J} \vbs[B]{\omega} \\ 
    & - \mathbf{J}(\vbs[B]{\omega}^\wedge \vbs{R}_\text{WB}^\top \vbs{R}_\text{WB}^\text{des} \vbs[B]{\omega}^\text{des} - \vbs{R}_\text{WB}^\top \vbs{R}_\text{WB}^\text{des} \vbsd[B]{\omega}^\text{des}), \\
    & \mathbf{e}_R =  \frac{1}{2}({\vbs{R}_\text{WB}^\text{des}}^\top \vbs{R}_\text{WB} -  \vbs{R}_\text{WB}^\top \vbs{R}_\text{WB}^\text{des})^{\vee}, \\
    & \mathbf{e}_\omega = \vbs[B]{\omega} - \vbs{R}_\text{WB}^\top \vbs{R}_\text{WB}^\text{des} \vbs[B]{\omega}^\text{des}.
\end{split}
\small
\end{equation}

The diagonal matrices $\mathbf{K}_R, \mathbf{K}_\omega$ of size $3 \times 3$ are tuning parameters of the controller, while $\mathbf{e}_R$ denotes the attitude error, and $\mathbf{e}_\omega$ is its time derivative. The symbol $(\mathbf{r}^\wedge)^\vee = \mathbf{r}$ denotes the operation transforming a $3 \times 3$ skew-symmetric matrix $\mathbf{r}^\wedge$ in a vector $\mathbf{r} \in \mathbb{R}^3$.
}
The position controllers designed in the next sections output setpoints for the attitude controller, and desired thrust $t_\text{cmd}$.

\subsection{Linear \ac{RTMPC} for Trajectory Tracking} \label{subsec:linear_mpc}
The model employed by the linear \ac{RTMPC} for trajectory tracking (\cref{eq:rtmpc_optimization_problem}) is based on a simplified, hover-linearized model derived from \cref{eq:mav_reduced_model_for_nmpc}, using the approach in \cite{kamel2017linear}, but modified to account for uncertainties. First, similar to \cite{kamel2017linear}, we express the model in a yaw-fixed, gravity-aligned frame $\text{I}$ via the rotation matrix $\vbs{R}_\text{BI}$
\begin{equation}
    \begin{bmatrix}
    \phi \\
    \theta \\
    \end{bmatrix} =
    \vbs{R}_\text{BI}
    \begin{bmatrix}
    \prescript{}{I}{\phi} \\
    \prescript{}{I}{\theta} \\
    \end{bmatrix}, \hspace{1pt}
    \vbs{R}_\text{BI} =
    \begin{bmatrix}
    \cos(\psi) & \sin(\psi) \\
    -\sin(\psi) & \cos(\psi) \\
    \end{bmatrix},
\end{equation}
where the attitude has been represented, for interpretability, via the Euler angles yaw $\psi$, pitch $\theta$, roll $\phi$ (\textit{intrinsic} rotations around the $z$-$y$-$x$ such that $\vbs{R} = \vbs{R}_{z}(\psi)\vbs{R}_{y}(\theta) \vbs{R}_{x}(\phi)$, with $\vbs{R}_{l}(\alpha)$ being a rotation of $\alpha$ around the ${l}$-th axis). 
Second, as in \cite{kamel2017linear}, we assume that the closed-loop attitude dynamics can be described by a first-order dynamical system that can be identified from experiments, replacing \cref{eq:mav_model_full:rot_kin}, \cref{eq:mav_model_full:rot_dyn}. 
Last, different from \cite{kamel2017linear}, we assume $\vbs[W]{f}_\text{ext}$ in \cref{eq:mav_model_full:tr_dyn} to be an unknown disturbance/model errors that capture the uncertain parts of the model, such that $\vbs[W]{f}_\text{ext} \in \mathbb{W}$.

The controller generates tilt (roll, pitch) and thrust commands ($n_u = 3$) given the state of the robot ($n_x=8$, consisting of position, velocity, and tilt), and given the reference trajectory. 
The desired yaw is fixed, and it is tracked by the cascaded attitude controller; similarly, $\vbs[B]{\omega}^\text{des}$ and $\vbsd[B]{\omega}^\text{des}$ are set to zero.  We employ the nonlinear attitude compensation scheme in \cite{kamel2017linear}.

The controller takes into account position constraints (e.g., available 3D flight space), actuation limits, and velocity/tilt limits via $\mathbb{X}$ and $\mathbb{U}$. The cross-section of the tube $\mathbb{Z}$ is a constant outer approximation based on an axis-aligned bounding box. It is estimated via Monte-Carlo sampling, by measuring the state deviations of the closed loop linear system $\vbf{A}_K$ under the disturbances in $\mathbb{W}$.

\subsection{Nonlinear \ac{RTMPC} for Acrobatic Maneuvers} \label{subsec:nonlinar_mpc}

\noindent 
\textbf{Ancillary \ac{NMPC}.}
We start by designing the ancillary \ac{NMPC} (\cref{eq:ancillary_nmpc_eq}). The selected nominal model is the same used in the high-performance trajectory tracking \ac{NMPC} for multirotors~\cite{loquercio2019deep}: 
\begin{equation}
\label{eq:full_geometric_attitude_controller}
\begin{split}
    \vbsd[W]{p} & = \b{\vbs[W]{v}} \\
    \vbsd[W]{v} & = \b{m^{-1}(\vbs{R}_\text{WB} \vbs[B]{t}_\text{cmd} + \vbs[W]{f}_\text{drag}) - \vbs[W]{g}} \\
    \vbsd{q}_\text{WB} & = \frac{1}{2} \vbs{\Omega} (\vbs[B]{\omega}_\text{cmd})\vbs{q}_\text{WB},
\end{split}
\end{equation}

where the rotational dynamics (\cref{eq:mav_model_full:rot_dyn}) have been neglected, assuming that the cascaded attitude controller enables fast tracking of the desired angular velocity setpoint $\vbs[B]{\omega}_\text{cmd}$.  
The controller uses the state and control input:
\begin{equation}
\label{eq:ancillary_nmpc_state_and_control_input}
    \bar{\vbf{x}} = [\vbs[W]{p}^\top, \vbs[W]{v}^\top, \vbs{q}_\text{WB}^\top ]^\top, \;\; \bar{\vbf{u}} = [t_\text{cmd}, \vbs[B]{\omega}_\text{cmd}^\top]^\top.
\end{equation}
\b{The feed-forward angular acceleration for the attitude controller $\vbsd[B]{\omega}_\text{cmd}$ is obtained via numerical differentiation. We do not explicitly generate an attitude setpoint (we set $\vbs{R}^\text{des}_\text{WB} = \vbs{R}_\text{WB}$), so that \cref{eq:full_geometric_attitude_controller} acts as a proportional body-rates controller with feed-forward accelerations}.

\noindent 
\textbf{Near-Minimum Time Safe Plan Generation.}
To compute safe nominal plans for acrobatic maneuvers (by solving the \ac{OCP} in \cref{eq:nmpc_nominal}), we employ an extended version of the full nonlinear dynamic model in \cref{subsec:mav_model}. More specifically, we solve the  \ac{OCP} in \cref{eq:nmpc_nominal} by using the following state $\tilde{\vbf{z}} \in \tilde{\bar{\mathbb{Z}}}$ and control inputs $\tilde{\vbf{v}} \in \tilde{\bar{\mathbb{V}}}$: %
\begin{equation}
\label{eq:agile_fligh:nominal_dynamics}
\tilde{\vbf{z}} =
[
    \vbs[W]{p}^\top, \vbs[W]{v}^\top, \vbs{q}_\text{WB}^\top, \vbs[B]{\omega}^\top, \vbs[B]{f}_\text{prop}^\top 
]^\top
\;\; \tilde{\vbf{v}} = \vbsd[B]{f}_\text{prop},
\end{equation}
where the state has been extended to include the thrust produced by each propeller $\vbs{f}^\text{prop}$ to ensure continuity in the reference thrust, accounting for the unmodeled actuators' dynamics. As for the linear case, uncertainties are modeled by $\vbs[W]{f}_\text{ext} \in \mathbb{W}$.
The cost function captures the near-minimum time objective: 
\begin{equation}
    \tilde{J}_\text{RTNMPC} = T_f + \alpha_1\!\vbs{v}^\top\!\!\vbs{v} + \alpha_2 {\vbs{f}_\text{prop}}^\top\!\!\vbs{f}_\text{prop} + \alpha_3\!\tilde{\vbf{v}}^\top\!\!\tilde{\vbf{v}}
\end{equation}
where $T_f$ is the total time of the maneuver, while the remaining terms act as a regularizer for the optimizer, with $\alpha_i \ll T_f$ (i.e., $\alpha_i \approx 10^{-2}, ~\forall ~i$). 

We note that $\tilde{J}_\text{RTNMPC}$ contains a non-quadratic term, therefore differing from the quadratic cost employed in the safe nominal planner in \cite{mayne2011tube} (our \cref{eq:nmpc_nominal}); such cost function was chosen to automate the selection of the prediction horizon $N$ for the safe nominal plan. Our evaluation will demonstrate that the ancillary \ac{NMPC} maintains the system within a tube from the generated reference, further highlighting the flexibility of the framework. 

Additionally, we note that state and control input (\cref{eq:agile_fligh:nominal_dynamics}) have been extended compared to the ones (\cref{eq:ancillary_nmpc_state_and_control_input}) selected for the ancillary \ac{NMPC}, as emphasized by our notation $\tilde{\cdot}$. For this reason, the optimal safe nominal plan $\tilde{\vbf{Z}}_t^*$, $\tilde{\vbf{V}}_t^*$ found using the extended state needs to be mapped to the reference trajectory for the ancillary \ac{NMPC}, $\vbf{Z}_t^*$ $\vbf{V}_t^*$. This is done by simply selecting position, velocity and attitude from $\tilde{\vbf{Z}}_t^*$ to obtain $\vbf{Z}_t^*$. The thrust setpoint $t_\text{cmd}$ in $\vbf{V}_t^*$ is computed via $\vbs{\mathcal{A}}$ in \cref{eq:allocation_matrix} from $\vbs{f}_\text{prop}$ in $\tilde{\vbf{Z}}_t^*$, while the angular velocity setpoint $\vbs{\omega}_\text{cmd}$ is obtained by assuming it equal to the angular velocity $\vbs{\omega}$ in $\tilde{\vbf{Z}}_t^*$.

\noindent 
\textbf{Constraints.} The state constraint $\bar{\vbf{x}}_t \in \mathbb{X}$ encodes the maximum safe linear velocity $\vbs{v}$ and position boundaries $\vbs{p}$ of the environment, while actuation constraints $\bar{\vbf{u}}_t \in \mathbb{U}$ account for physical limits of the robot, restricting the nominal angular velocities $\vbs{\omega}_\text{cmd}$ (to prevent saturation of the onboard gyroscope), and the maximum/minimum thrust force $t_\text{cmd}$ produced by the propellers.
We impose tightened constraints on the thrust force by constraining $\vbs{f}_\text{prop}$  in $\tilde{\vbf{z}} \in \tilde{\bar{\mathbb{Z}}}$. These constraints are obtained via a conservative approach, i.e. we require a minimal thrust to generate a trajectory feasible within our position and velocity constraints. \b{Such feasible trajectory is found via an iterative tightening procedure for the thrust constraints, using the previously-obtained feasible trajectory as an initial guess for the subsequent optimization under tightened thrust constraints}.
\b{This procedure ensures} that sufficient control authority is left to the ancillary \ac{NMPC} to account for the presence of large unknown aerodynamic effects and mismatches in the mapping from commanded thrust/actual thrust. This cautious approach enabled successful real-world execution of the maneuver without further real-world tuning. We additionally leverage the further degrees of freedom introduced by the extended state $\tilde{\vbf{z}}$ by shaping the safe plan through upper-bounding the thrust rates $\dot{\vbs{f}}_\text{prop}$ via $\tilde{\bar{\mathbb{V}}}$, although this constraint will not be enforced by the ancillary \ac{NMPC}. Last, using Monte-Carlo closed-loop simulations with disturbances sampled from $\mathbb{W}$, we verify that $\mathbb{X}$ and $\mathbb{U}$ are satisfied, and we generate a constant estimate (outer approximation, axis-aligned bounding box) of the cross-section of the tubes $\mathbb{T}^\text{state}$ and $\mathbb{T}^\text{action}$.

\noindent 
\textbf{Tube and Data Augmentation with Attitude Quaternions.}
The normalized attitude quaternion, part of the states $\bar{\vbf{x}}$, $\tilde{\vbf{z}}$ of nonlinear \ac{RTMPC}, and part of the reference $\vbf{Z}_t^*$ for the ancillary \ac{NMPC}, does not belong to a vector space, and therefore it is not trivial to describe its tube nor to generate extra samples for \ac{DA}. In this work, we employ an attitude error representation $\vbs{\epsilon} \in \mathbb{R}^3$ based on the \ac{MRP} \cite{shuster1993survey} to generate a representation that can be treated as an element of a vector space. Specifically, we use
\begin{equation}
\vbs{\epsilon}_t = \text{MRP}(\vbs{q}_t \odot {\vbs{q}^*_t}^{-1}),
\end{equation}
where $\vbs{q}_t$ is the current attitude, ${\vbs{q}_t^*}$ is the desired attitude (from the safe plan $\vbf{z}_t^*$), $\text{MRP}(\cdot)$ maps a quaternion to the corresponding three-dimensional attitude representation, while $\odot$ denotes the quaternion product.

\section{Evaluation - Learning From Linear RTMPC} \label{sec:evaluation_linear}
We start by evaluating our policy learning approach for the task of trajectory tracking using the linear \ac{RTMPC} expert. 

\subsection{Evaluation Approach and Details}
\noindent
\textbf{Simulation Environment.} Demonstration collection and policy evaluations are performed in a simulation environment implementing the nonlinear multirotor dynamics in \cref{subsec:mav_model}, discretized at \b{$400$ Hz, while the attitude controller runs at $200$ Hz}. The robot follows desired trajectories, starting from randomly generated initial states centered around the origin. Given the specified external disturbance magnitude bound $\mathbb{W}_\anydomain = \{ f_\text{ext} \in \mathbb{R}| \underline{f}_\text{ext} \leq f_\text{ext} \leq \overline{f}_\text{ext}\}$, disturbances are applied in the domain $\anydomain$ by sampling $\vbs[W]{f}_\text{ext}$ via the spherical coordinates: 
\vspace{-0.3cm}
\begin{equation} \label{eq:disturbance_sampling}
\vbs[W]{f}_\text{ext} = 
f_\text{ext} 
\begin{bmatrix}
\cos(\phi) \sin(\theta) \\
\sin(\phi) \sin(\theta) \\
\cos(\theta)
\end{bmatrix}, ~~~
\begin{aligned}
&f_\text{ext} \sim \mathcal{U}(\underline{f}_\text{ext}, \overline{f}_\text{ext}), \\
&\theta \sim \mathcal{U}(0, \pi), \\
&\phi \sim \mathcal{U}(0, 2\pi).
\end{aligned}
\end{equation}
\vspace{-0.3cm}

\noindent
\textbf{Linear \ac{RTMPC}.}
The linear \ac{RTMPC} \b{expert demonstrator runs in simulation at $10$ Hz, and its tube is designed assuming $\mathbb{W} = \{f_\text{ext} \in \mathbb{R} | 0 \leq f_\text{ext} \leq 0.35 mg\}$, corresponding to the safe physical limit of the actuators of the robot}. The reference fed to the expert is a sequence of desired positions and velocities for the next $3$s, discretized with a sampling time of $0.1$ s; the expert uses a corresponding planning horizon of $N=30$, (resulting in a reference being a $180$-dim. vector). 

\noindent
\textbf{Policy.} The student policy is a $2$-hidden layer, fully connected \ac{DNN}, with $(32,32)$ or $(64,32)$ neurons/layer, and \texttt{ReLU} activation function. The input/output size match the ones of the optimization problem solved by the expert in \cref{subsec:linear_mpc}: the input has size $188$ (state and reference trajectory), while the output has size $3$ (thrust, and tilt in an inertial frame).

\noindent
\textbf{Baselines and Training Details.} \b{We apply the proposed \ac{SA} strategies to every demonstration collected via \ac{DAgger} or \ac{BC}, and we consider \ac{DAgger} or \ac{BC} without any augmentation/robustification approach (denoted \textbf{n.a.}), or combined with:
\begin{enumerate}[a)]
\item \textbf{DA (linear interpolation)}: a \ac{DA} that groups the collected demonstrations  based on the input reference trajectory (or reference position/time for the go-to-goal-position case), and than randomly samples pairs of input-outputs in each cluster, linearly interpolating the state/action to obtain a new state-action pair. 
\item \textbf{DA (expert neighborhood)}: a \ac{DA} strategy that uniformly samples states from a region corresponding to $5\%$ of the cross-section of the tube in \ac{RTMPC}, centered around the current state of the robot. The corresponding actions are obtained using the ancillary controller. This baseline is useful at studying the importance of using the entire tube as a support of the sampling distribution. 
\item \textbf{\ac{DR}}: domain randomization. 
\end{enumerate}
During demonstration-collection in the source environment $\mathcal{S}$, we do not apply disturbances, setting $\mathbb{W}_\mathcal{S} = \{ \emptyset \}$, with the exception for \ac{DR}, where we sample disturbances from $\mathbb{W}_\text{DR} = \mathbb{W}_\mathcal{T}$. 
In all the methods that use \text{DAgger} we set the probability of using actions of the expert $\beta$ (a hyperparameter of DAgger \cite{ross2011reduction}) to be $1$ at the first demonstration and $0$ otherwise (as this was found to be the best-performing setup).
\b{The number of samples generated for the baseline DA methods is $16$ per timesteps, matching the number used for \ac{SA}-sparse, while \ac{SA}-dense corresponds to $256$ samples per timestep.}
Demonstrations are collected at control rate ($0.1$ s). After every collected demonstration, the policy is trained for up to $50$ epochs using all the data available so far with the ADAM \cite{kingma2014adam} optimizer and a learning rate of $0.001$, and we use early stopping, terminating the training if the validation loss (from $30\%$ of the collected data) does now decreases within $7$ epochs.} The policy is then evaluated on the task for $10$ times (episodes), starting from slightly different initial states centered around the origin, in both $\mathcal{S}$ and $\mathcal{T}$. 

\noindent 
\textbf{Evaluation Metrics:}
We monitor:
\begin{enumerate}[i)]
    \item \textit{Robustness (Success Rate)}, as the percentage of episodes where the robot never violates any state constraint;
    \item \textit{Performance}, via either
        \begin{enumerate}[a)]
            \item $C_{\vbs{\xi}}({\pi_\theta}) := \sum_{t=0}^{T}\|\vbf{x}_t - \vbf{x}_t^\text{des}\|^2_{\vbf{Q}_x}+ \|\vbf{u}_t\|_{\vbf{R}_u}^2$ tracking error along the trajectory (\textit{MPC Stage Cost}); or
            \item $\|C_{\vbs{\xi}}({\pi_{{\theta}^*}}) - C_{\vbs{\xi}}(\pi_{\hat{\theta}^*})\|/\|C_{\vbs{\xi}}(\pi_{{\theta}^*})\|$ relative error between expert and policy tracking errors (\textit{Expert Gap}); 
        \end{enumerate} 
    \item \textit{Efficiency}
    \begin{enumerate}
        \item number of expert demonstrations (\textit{Num. Demonstrations Used for Training}), and
        \item wall-clock time to generate the policy (\textit{Training Time}~\footnote{Training time is the time to collect demonstrations and the time to train the policy, as measured by a wall-clock. In our evaluations, the simulated environment steps at its highest possible rate (in contrary to running at the same rate of the simulated physical system), providing an advantage to those methods that require a large number of environment interactions, such as the considered baselines.}).
    \end{enumerate}
\end{enumerate}

\subsection{Numerical Evaluation of Efficiency, Robustness, and Performance when Learning to Track a Single Trajectory}

\begin{figure}
\captionsetup[sub]{font=footnotesize}
\centering
\begin{subfigure}{\columnwidth}
    \centering
    
    \centering
    \includegraphics[width=\columnwidth, width=\columnwidth, trim={3cm 0.5cm 3cm 0.5cm},clip]{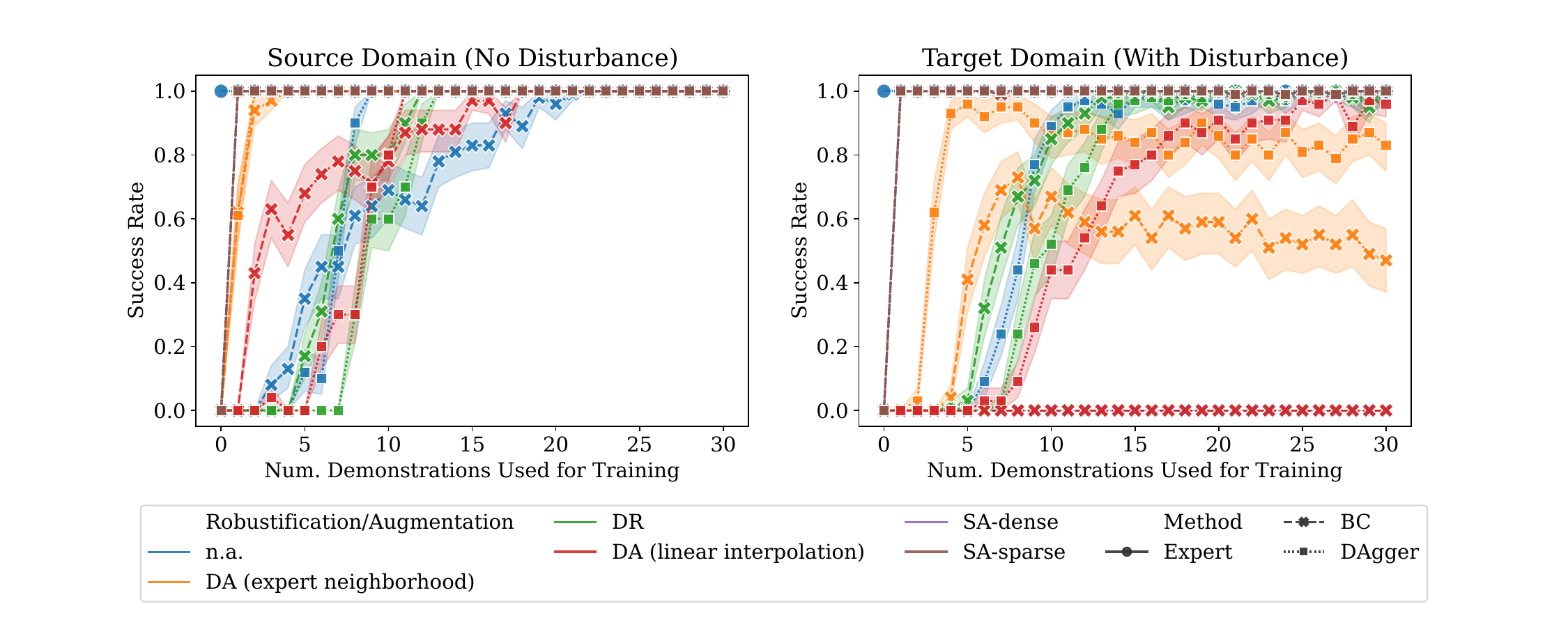}

\end{subfigure}%
    \caption{Robustness (\textit{Success Rate}) in the task of flying along a figure-8 trajectory ($7$ s long), with wind-like disturbances (right, target domain $\mathcal{T}_1$) and without (left, source domain $\mathcal{S}$), starting from different initial states. Evaluation across $10$ random seeds, $10$ times per demonstration per seed. Shaded lines are the $95\%$ confidence interval. The lines for the SA-based methods overlap.}
    \label{fig:single_trajectory_eval}
\end{figure}

\begin{table}[tb]
    \caption{
    Comparison of \ac{IL} methods to learn a policy from RTMPC. The proposed SA-methods simultaneously achieve high robustness, demonstration efficiency and performance close to the one of the expert, unlike the considered baselines.
    Note: Robustness and performance are evaluated at convergence (demonstration $20$-$30$ for non-SA methods, and $1$-$11$ for SA-methods). 
    \textit{Demonstration-Efficiency}: number of demonstrations to achieve for the first time an average $100\%$ success rate. \textit{Easy}: no  disturbances applied during data collection. \textit{Safe}: no state constrain violations recorded during data collection.
    *Safe in our numerical evaluation, but not guaranteed as it requires executing actions of a policy that may be partially trained. Color-coding: better: green/white; worse: red.}
    \label{tab:comparison}
    
    \newcommand*{\opacity}{40}%

\definecolor{hight}{HTML}{ec462e} 
\definecolor{lowt}{HTML}{76f013}  %
\newcommand*{\minvalt}{1.0}%
\newcommand*{\maxvalt}{18.0}%
\newcommand{\grdt}[1]{
    \pgfmathparse{int(round(100*(min(max(#1, \minvalt), \maxvalt)/(\maxvalt-\minvalt))-(\minvalt*(100/(\maxvalt-\minvalt)))))}
    \xdef\tempa{\pgfmathresult}
    \cellcolor{hight!\tempa!lowt!\opacity} #1
}

\definecolor{highp}{HTML}{ec462e}
\definecolor{lowp}{HTML}{76f013}
\newcommand*{\minvalp}{6.0}%
\newcommand*{\maxvalp}{15.0}%
\newcommand{\grdp}[1]{
    \pgfmathparse{int(round(100*(min(max(#1, \minvalp), \maxvalp)/(\maxvalp-\minvalp))-(\minvalp*(100/(\maxvalp-\minvalp)))))}
    \xdef\tempa{\pgfmathresult}
    \cellcolor{highp!\tempa!lowp!\opacity} #1
}

\definecolor{highr}{HTML}{ffffff}
\definecolor{lowr}{HTML}{ec462e}
\newcommand*{\minvalr}{90}%
\newcommand*{\maxvalr}{100}%
\newcommand{\grdr}[1]{
    \pgfmathparse{int(round(100*(min(max(#1, \minvalr), \maxvalr)/(\maxvalr-\minvalr))-(\minvalr*(100/(\maxvalr-\minvalr)))))}
    \xdef\tempa{\pgfmathresult}
    \cellcolor{highr!\tempa!lowr!\opacity} #1
}

\newcommand{\gradientcell}[6]{
    \ifdimcomp{#1pt}{>}{#3 pt}{#1}{
        \ifdimcomp{#1pt}{<}{#2 pt}{#1}{
            \pgfmathparse{int(round(100*(#1/(#3-#2))-(\minval*(100/(#3-#2)))))}
            \xdef\tempa{\pgfmathresult}
            \cellcolor{#5!\tempa!#4!#6} #1
    }}
}
\newcolumntype{P}[1]{>{\centering\arraybackslash}p{#1}}
\tiny
\renewcommand{\tabcolsep}{1pt}
\centering
\begin{tabular}{|p{1.4cm}p{0.9cm}||P{0.7cm}|P{0.7cm}|P{0.7cm}|P{0.7cm}|P{0.7cm}|P{0.7cm}|P{0.7cm}|P{0.7cm}|}
\hline
\multicolumn{2}{|c||}{Method} &  
\multicolumn{2}{c|}{Data Collection} & 
\multicolumn{2}{c|}{\makecell{Robustness\\succ. rate (\%)}} &
\multicolumn{2}{c|}{\makecell{Performance\\expert gap (\%)}} &
\multicolumn{2}{c|}{\makecell{Demonstration\\Efficiency}} \\

\makecell{Robustification/ \\ Augmentation}   & Imitation          & Easy          & Safe             & $\mathcal{T}_1$       & $\mathcal{T}_2$ & $\mathcal{T}_1$  & $\mathcal{T}_2$       & $\mathcal{T}_1$        & $\mathcal{T}_2$ \\
\hline
\hline
\multirow{2}{*}{n.a.} 
                      & BC                 & Yes           & Yes              & \grdr{0.0}        & \grdr{100.0}          & \grdp{22.8}    & \grdp{37.2}  & -\cellcolor{hight!\opacity} & \grdt{18} \\
                      & DAgger             & Yes           & No \cellcolor{hight!\opacity}                        & \grdr{97.6}       & \grdr{100.0}          & \grdp{13.6}    & \grdp{2.9}   & -\cellcolor{hight!\opacity} & \grdt{9} \\ 
\hline
\multirow{2}{*}{DA \makecell{(linear\\ interpolation)}}
                      & BC                 & Yes           & Yes              & \grdr{0.0}        & \grdr{99.9}           & \grdp{23.3}    & \grdp{36.7}  & -\cellcolor{hight!\opacity}  & \grdt{16} \\
                      & DAgger             & Yes           & No \cellcolor{hight!\opacity}                        & \grdr{92.9}       & \grdr{100.0}          & \grdp{26.9}    & \grdp{3.7}   & -\cellcolor{hight!\opacity}  & \grdt{12} \\
\hline
\multirow{2}{*}{DA \makecell{(expert\\ neighborhood)}} 
                      & BC                 & Yes           & Yes              & \grdr{53.5}       & \grdr{100.0}          & \grdp{27.5}    & \grdp{2.7}   & -\cellcolor{hight!\opacity}  & \grdt{3} \\
                      & DAgger             & Yes           & No \cellcolor{hight!\opacity}                        & \grdr{83.3}       & \grdr{100.0}          & \grdp{22.3}    & \grdp{2.7}   & -\cellcolor{hight!\opacity}  & \grdt{2} \\
\hline
\multirow{2}{*}{DR}   
                      & BC                 & No \cellcolor{hight!\opacity}                     & Yes              & \grdr{98.7}       & \grdr{100.0}          & \grdp{6.8}     & \grdp{8.5}   & \grdt{15}      & \grdt{14} \\
                      & DAgger             & No \cellcolor{hight!\opacity}                     & No \cellcolor{hight!\opacity}                        & \grdr{99.1}       & \grdr{100.0}          & \grdp{6.7}     & \grdp{2.8}   & \grdt{20}      & \grdt{9} \\
\hline
\multirow{2}{*}{\textbf{SA-Dense}} 
                      & BC                 & Yes           & Yes              & \grdr{100.0}      & \grdr{100.0}         & \grdp{6.3}      & \grdp{2.8}   & \grdt{1}              & \grdt{1} \\
                      & DAgger             & Yes           & Yes$^*$          & \grdr{100.0}      & \grdr{100.0}         & \grdp{6.3}      & \grdp{2.8}   & \grdt{1}              & \grdt{1} \\
\hline
\multirow{2}{*}{\textbf{SA-Sparse}} 
                      & BC                 & Yes           & Yes              & \grdr{99.9}        & \grdr{100.0}         & \grdp{6.2}     & \grdp{2.8}   & \grdt{1}              & \grdt{1} \\
                      & DAgger             & Yes           & Yes$^*$          & \grdr{100.0}       & \grdr{100.0}         & \grdp{6.3}     & \grdp{2.8}   & \grdt{1}              & \grdt{1} \\
\hline
\end{tabular}%

\end{table}

\begin{figure*}[htbp!]
\captionsetup[sub]{font=footnotesize}
\centering
\begin{subfigure}{0.21\paperwidth}
    \centering
    \includegraphics[height=3.5cm, trim={0.5cm 5.5cm 0.25cm 5.5cm},clip]{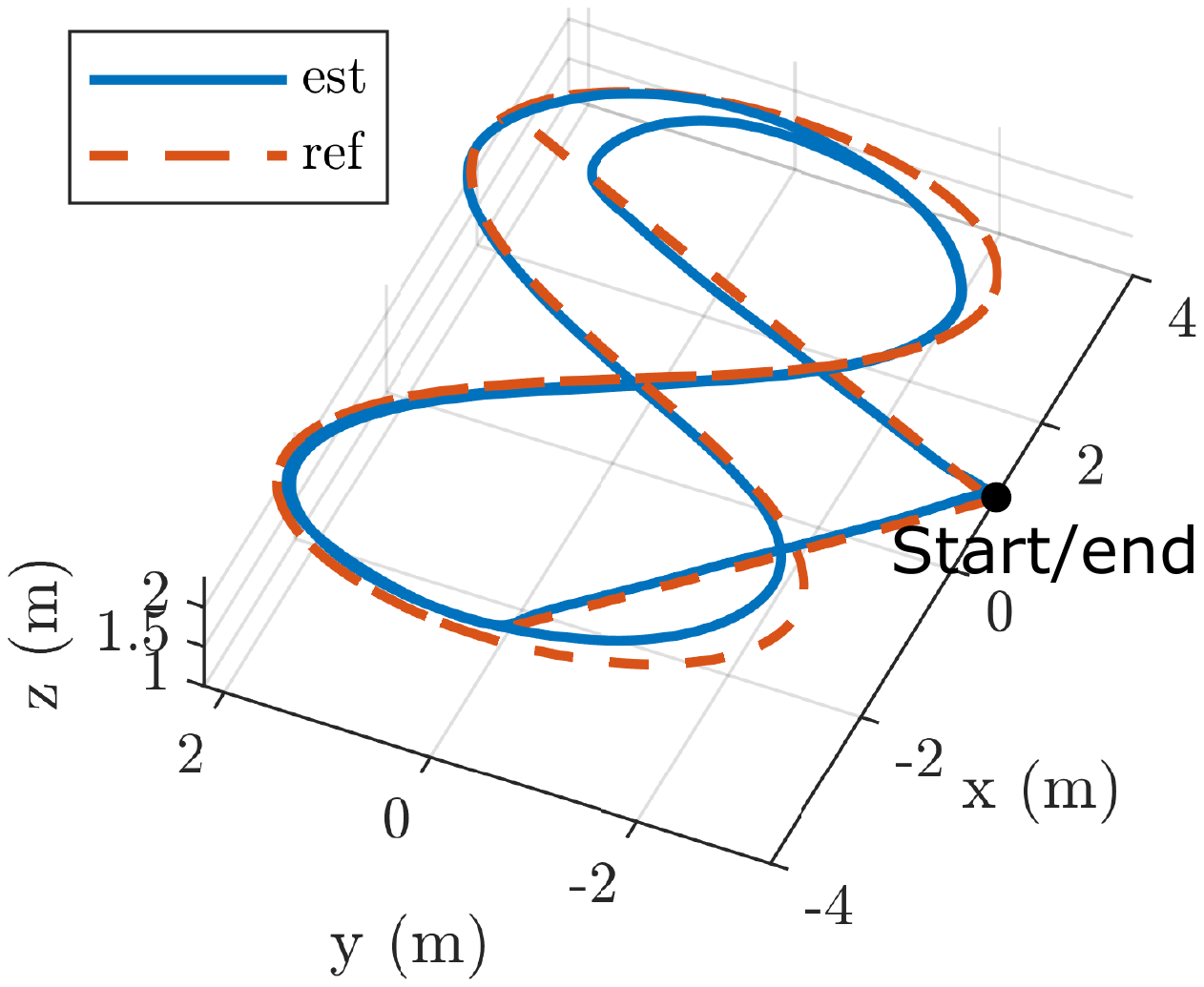}
    \caption{Reference and actual trajectory}
    \label{fig:reference_and_actual_trajectory}
\end{subfigure}%
\hspace{0.11cm}
\begin{subfigure}{0.38\paperwidth}
    \centering
    \includegraphics[height=3.5cm, trim={0.5cm 2cm 1cm 2cm}, clip]{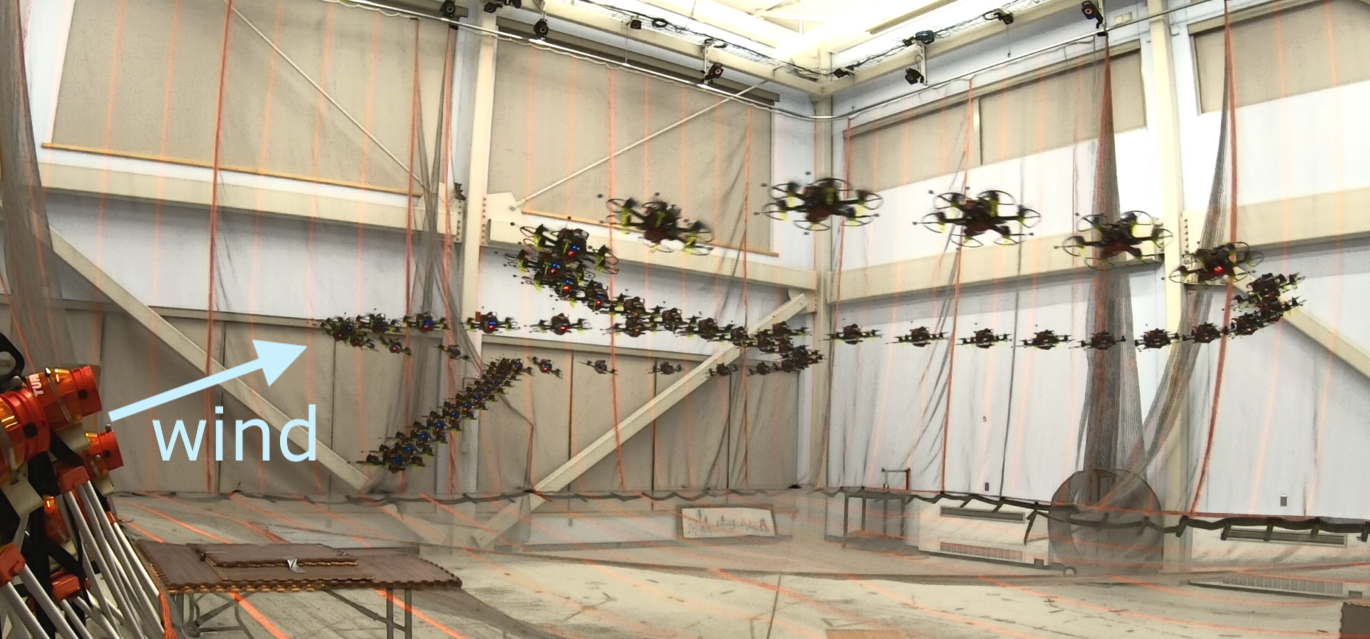}
    \caption{Time-lapse of the trajectory, and wind ($6$ m/s) from leaf blowers}
    \label{fig:experiment}
\end{subfigure}%
\hspace{0.11cm}
\begin{subfigure}{0.21\paperwidth}
    \centering
    \includegraphics[height=3.3cm, trim={0cm, 0.4cm, 0.4cm, 0.8cm}]{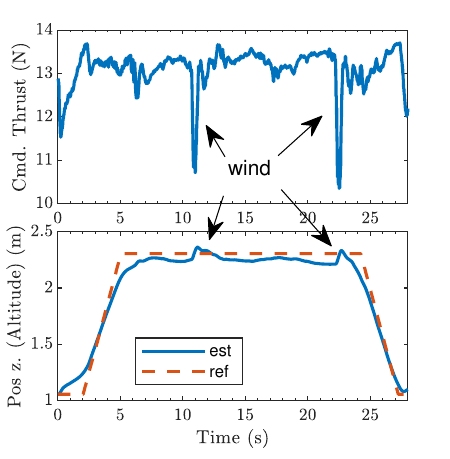}
    \caption{Effects of wind}
    \label{fig:thrust_and_altitude}
\end{subfigure}
    \caption{Experimental evaluation of a trajectory tracking policy learned from a \textit{single} linear \ac{RTMPC} demonstration collected in simulation, achieving \textit{zero-shot} transfer. The multirotor is able to withstand previously unseen disturbances, such as the wind produced by an array of leaf-blowers, and whose effects are clearly visible in the altitude errors (and change in commanded thrust) in \cref{fig:thrust_and_altitude}.  
    This demonstration-efficiency and robustness is enabled by Sampling-Augmentation (SA), our proposed tube-guided data augmentation strategy.}
    \label{fig:experiment_lemniscate}
\end{figure*}

\noindent
\textbf{Tasks Description.}
Our objective is to generate a policy from linear \ac{RTMPC} capable of tracking a $7$s long ($70$ steps), figure eight-shaped trajectory. We evaluate the considered \ac{IL} approaches in two different target domains, with wind-like disturbances ($\mathcal{T}_1$) or with model errors ($\mathcal{T}_2$). Disturbances in $\mathcal{T}_1$ are external force perturbations $\vbs{f}_\text{ext}$ sampled from $\mathbb{W}_{\mathcal{T}_1} \approx \{f_\text{ext}|0.25mg \leq f_\text{ext} \leq 0.3mg\}$. Model errors in $\mathcal{T}_2$ are applied via mismatches in the drag coefficients used between training and testing, representing uncertainties not explicitly considered during the design of the linear \ac{RTMPC}.

\noindent
\b{\textbf{Comparison with IL baselines.}} We start by evaluating the robustness in $\mathcal{T}_1$ as a function of the number of demonstrations collected in the source domain.
The results are shown in \cref{fig:single_trajectory_eval}, highlighting that:
\begin{inparaenum}[i)]
\item while all the approaches achieve robustness (full success rate) in the source domain, \ac{SA} achieves full success rate after only a single demonstration, \b{being $3$ times more sample efficient than the most demonstration-efficient baseline, DA (expert neighborhood), which however does not achieve full robustness in the target domain};
\item \ac{SA}, instead, is also able to achieve full robustness in the target domain, while baseline methods do not fully succeed or converge at a much lower rate. 
\end{inparaenum}
These results emphasize the presence of a distribution shift between the source and target, which is not fully compensated for by baseline methods such as \ac{BC} due to a lack of exploration and robustness.

The performance evaluation and additional results are summarized in \cref{tab:comparison}. We highlight that in the target domain $\mathcal{T}_1$, \ac{SA} achieves the performance that is closest to the expert.
\cref{tab:comparison} additionally presents the results for the target domain $\mathcal{T}_2$. Although this task is less challenging (i.e., all the approaches achieve full robustness), the proposed method (\ac{SA}-sparse) achieves the highest demonstration-efficiency and \b{among the} lowest expert gap, with similar trends as in $\mathcal{T}_1$.

\noindent
\textbf{Training Time.} %
\begin{table*}[]
\centering
    \renewcommand{\arraystretch}{1.2} %
    \setlength{\tabcolsep}{3pt}

    \scriptsize
    \resizebox{1\textwidth}{!}{

    \begin{tabular}{|lll||r|cc|cc||r|cc|cc|cc|cc|||c|}
    \hline
     &  &  & \multicolumn{5}{c||}{\textbf{Figure-8} ($27$ s, $3.0$ m/s, w/o tight pos constr.)} & \multicolumn{9}{c|||}{\textbf{Circle} ($27$ s, $3.5$ m/s, w/ tight position constraints)} & \multirow[c]{3}{*}{\makecell{\textbf{Constraints} \\ \textbf{satisfied}}} \\
     &  &  & \multirow[l]{2}{*}{\makecell{\# of\\Dem.}}  & \multicolumn{2}{c}{No Disturbance} & \multicolumn{2}{c||}{Slung Load} & \multirow[l]{2}{*}{\makecell{\# of\\Dem.}}& \multicolumn{2}{c}{No Disturbance} & \multicolumn{2}{c}{Slung Load} & \multicolumn{2}{r}{Slung Load+Wind} & \multicolumn{2}{r|||}{Drag+Wind} & \\ 
    \textbf{Method} & \textbf{Agent} & \textbf{MAE type (m, \( \downarrow \))} & & x,y & z & x,y & z &  & x,y & z & x,y & z & x,y & z & x,y & z &   \\
    \hline
    
    \hline
    \multirow[c]{2}{*}{Expert} & MPC & Tracking & n.a. & \textbf{0.143} & 0.145 & \textbf{0.158} & \textbf{0.418} & n.a.  &\textbf{0.151} & 0.183 & \textbf{0.207} & 0.563 & \textbf{0.193} & 0.506 & \textbf{0.219} & \textbf{0.284} & No\\
     & RTMPC & Tracking & n.a. & 0.144 & \textbf{0.114} & \textbf{0.158} & 0.423 & n.a. & 0.270 & \textbf{0.161} & 0.309 & \textbf{0.561} & 0.291 & \textbf{0.433} & 0.338 & 0.287 & \textbf{Yes} \\
     \hline 
     
     \multirow[c]{2}{*}{Student} & DAgger+DR & Gap from RTMPC & 10 & 0.062 & 0.124 & 0.091 & \textbf{0.060} & 20 & \textbf{0.035} & \textbf{0.034} & \textbf{0.059} & 0.103 & 0.043 & \textbf{0.033} & \textbf{0.052} & \textbf{0.025} & \textbf{Yes} \\
     & SA-Sparse & Gap from RTMPC & 1 & \textbf{0.057} & \textbf{0.121} & \textbf{0.081} & 0.100 & 1 &  0.037 & \textbf{0.034} & 0.067 & \textbf{0.090} & \textbf{0.036} & 0.040 & 0.060 & 0.033 & \textbf{Yes} \\
     \hline
    \end{tabular}
    } %

    \caption{Mean Absolute Error (MAE) in tracking a trajectory in experiments. \textit{Tracking MAE} is the distance of the agent's trajectory from the reference. \textit{Gap from RTMPC} is the distance of the agent's trajectory from the trajectory obtained using RTMPC (under the same type of disturbance). Results averaged across $3$ Circles and $2$ Figure-8 per agent.}
    \label{tab:lmpc_experimental_comparision}
\end{table*}

\cref{fig:single_trajectory_eval} highlights that the best-performing baseline, DAgger+DR, requires about $10$ demonstrations to learn to robustly track a $7$s long trajectory, which corresponds to a total training time of $10.8$s. Among the proposed approaches, DAgger+SA-sparse instead only requires $1$ demonstration, corresponding to a training time of $3.8$s, a $64.8\%$ reduction in wall-clock time required to learn the policy. DAgger+SA-dense, instead, while requiring a single demonstration to achieve full robustness, necessitates $114$s of training time due to the large number of samples generated. 
Due to its effectiveness and greater computational efficiency, we use \ac{SA}-sparse rather than \ac{SA}-dense for the rest of the evaluations.

\begin{figure}
    \centering
    \ifblue
    \begin{tikzpicture}
    \node[inner sep=0pt] (table) {
    \fi
    \includegraphics[width=\columnwidth, trim={1cm, 0.9cm, 0, 0.9cm}, clip]{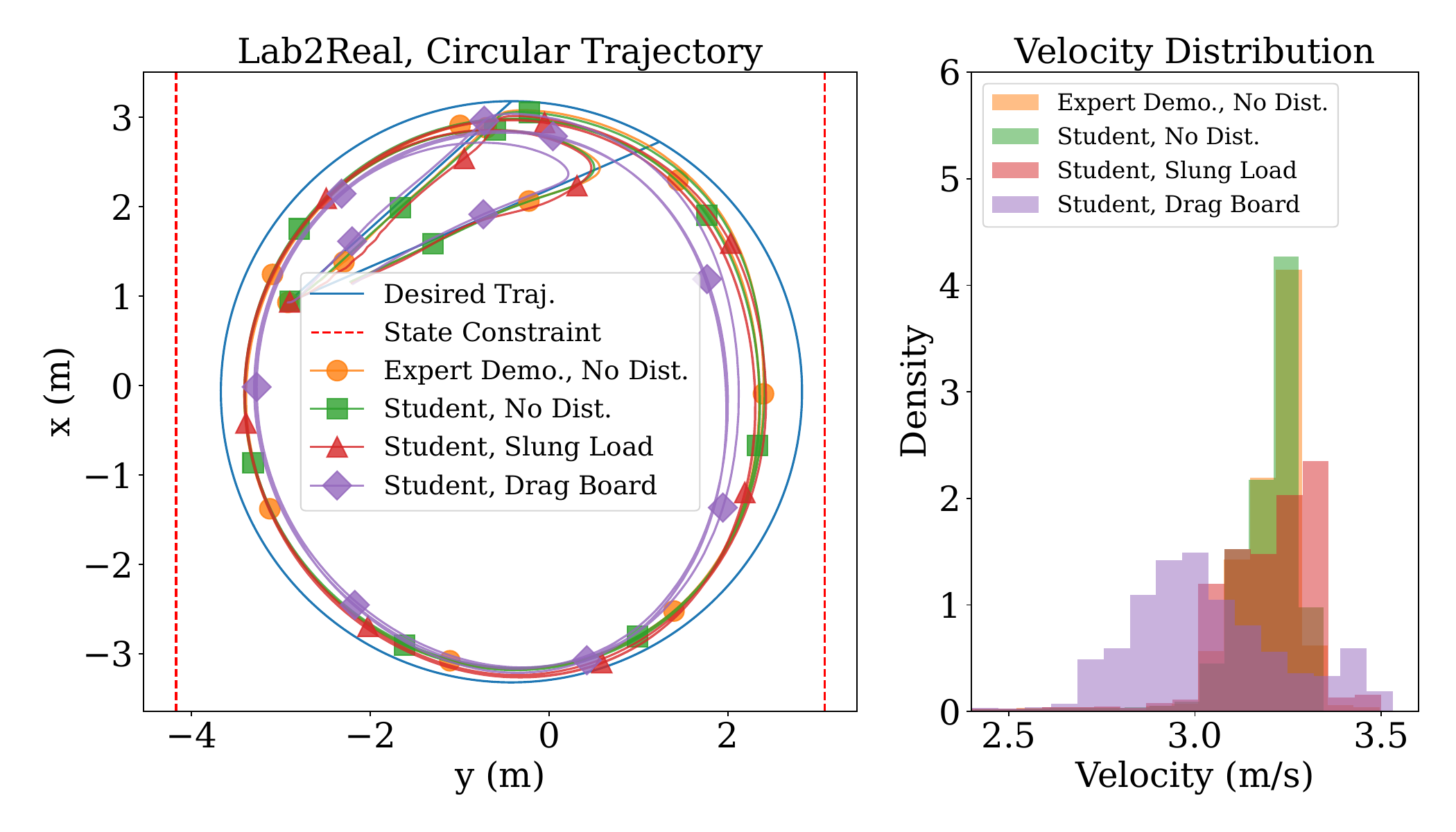}
    \ifblue
    };
    \draw[blue, thick] (table.north west) rectangle (table.south east);
    \end{tikzpicture}
    \fi
     
    \caption{Example of \textit{lab2real} transfer, where one RTMPC demonstration (Expert Demo.) collected with the actual robot is used to train a policy (Student) that is robust to previously unseen disturbances. Policy runs onboard at $500$ Hz. }
    \label{fig:lab2real_transfer}
\end{figure}

\subsection{Hardware Evaluation for Tracking a Single Trajectory from a Single Demonstration}
\noindent \textbf{Sim2Real Transfers.}
We validate the demonstration-efficiency, robustness, and performance of the proposed approach by experimentally testing policies trained after a \textit{single} demonstration collected in simulation using DAgger/BC (which operate identically since we use DAgger with $\beta=1$ for the first demonstration), combined with \ac{SA}-sparse. We use the MIT/ACL open-source snap-stack\footnote{https://gitlab.com/mit-acl/fsw/snap-stack} for controlling the attitude of the MAV. The learned policy runs at $100$ Hz on the onboard \texttt{Nvidia Jetson TX2} (CPU), with the reference trajectory provided at $100$ Hz. State estimation is from a motion capture system or onboard \ac{VIO}.

The task is to track a figure eight-shaped trajectory, with velocities up to $3.4$ m/s. We evaluate the robustness of the learned policy by applying a wind-like disturbance produced by an array of $3$ leaf blowers (\cref{fig:experiment_lemniscate}). The given position reference and the corresponding trajectory are shown in \cref{fig:reference_and_actual_trajectory}. The effects of the wind disturbances are clearly visible in the altitude errors and changes in commanded thrust in \cref{fig:reference_and_actual_trajectory} (at $t=11$ s and $t=23$ s). 
These experiments show that the learned policy can robustly track the desired reference, withstanding challenging perturbations unseen during the training phase. 

\noindent \textbf{Lab2Real Transfer.} We evaluate the ability of SA to learn from a single demonstration collected on a real robot in a controlled environment (lab) and generalize to previously unseen disturbances (real). We do so by using a \ac{RTMPC} demonstration of a circular trajectory (velocity up to $3.5$ m/s, with tight position constraints) with the multirotor, augmenting the collected demonstration with \ac{SA}-sparse, and deploying the learned policy while we apply previously unseen disturbances (drag board, slung load). As shown in the sequence in \cref{fig:lab2real_transfer}, despite the large distribution shifts in velocity, the policy reproduces the expert demonstration and it is robust to previously unseen disturbances. Our video\footnote{\url{https://youtu.be/-uiarBY1STU}} shows more examples.

\noindent
\textbf{Experimental Comparison.} \cref{tab:lmpc_experimental_comparision} reports a real-world comparison of SA (one demonstration) with MPC, RTMPC, and DAgger+DR ($10$ or $20$ demonstrations from RTMPC) on the task of tracking different trajectories with a duration of $27$ s, while the robot is subject to (1) wind speed up to $10$ m/s, (2) a slung load of $250$ grams, and (3) a surface attached at the bottom of the robot that produces extra drag ($0.2$ m$^2$). All the policies are learned in simulation. The results confirm our numerical findings, highlighting that SA-sparse achieves better or comparable performance and robustness than DAgger+DR, but under reduced training effort (one demonstration instead of $10$-$20$). In addition, the evaluation highlights that RTMPC achieves larger tracking errors when the position constraints are tight (e.g., the reference trajectory is close to the position constraint), due to RTMPC's ability to maintain a safe distance from such constraints. However, this same property allows RTMPC to be safe (no constraint violation), unlike MPC which violates position constraints in the case of Slung Load + Wind. The learned policy runs at $500$ Hz, while the RTMPC/and MPC run at their maximum rates ($100$ Hz, occupying the entire CPU).

\begin{table}[t]
    \renewcommand{\arraystretch}{1.4}
    \caption{Time (ms) to compute an action for the linear RTMPC expert (L-RTMPC) and the \ac{DNN} policy (Policy). \textbf{The \ac{DNN} policy is} $\mathbf{280}$ \textbf{times faster than the optimization-based expert (onboard)}, and $25$ times faster (offboard). Offboard computer (numerical evaluation and training): \texttt{Intel i9-10920} with two {RTX 3090} GPUs. Onboard implementation (C++, optimized for speed): on \texttt{NVIDIA TX2 CPU}.}
    \label{tab:computational_cost_linear_rtmpc}    \centering
    
    \resizebox{1.0\columnwidth}{!}{
    \ifblue
    \begin{tikzpicture}
    \node[draw, blue, thick, inner sep=0pt] (table) {
    \fi

    \begin{tabular} {|C{1.9cm}| C{1.7cm} |C{2.7cm}|C{0.5cm}|C{0.5cm}|C{0.5cm}|C{0.5cm}| }
    \multicolumn{3}{c}{} & \multicolumn{4} {c}{Time (ms)} \\
    \hline
        \textbf{Computer} & \textbf{Method}  & \textbf{Setup} & \textbf{Mean} & \textbf{SD} & \textbf{Min} & \textbf{Max}\\
        \hline 
        \hline 
        \multirow{2}{*}{Offboard} & L-RTMPC & CVXPY/OSQP  & $4.28$ & $0.39$ & $4.21$ & $16.66$ \\ %
        \cline{2-7}
        & \textbf{Policy}        & PyTorch           & $\mathbf{0.17}$ & $\mathbf{0.00}$ & $\mathbf{0.17}$ & $\mathbf{0.22}$  \\
        \hline
        \hline
        \multirow{2}{*}{Onboard} & L-RTMPC  & C++/CVXGEN&  $8.4$ & $1.4$ & $4.5$ & $15.9$  \\ %
        \cline{2-7}
                                 & \textbf{Policy}   & C++/Eigen  & \textbf{0.03} & \textbf{0.01} & \textbf{0.02} & \textbf{0.24} \\  
        \hline
    \end{tabular}
    \ifblue
    };
    \end{tikzpicture}
    \fi
    } %
\vskip-4ex
\end{table}
\noindent
\textbf{Computation.} \cref{tab:computational_cost_linear_rtmpc} shows that the \ac{DNN} policy is \b{$280$} times faster than the expert on the CPU of the onboard computer (\texttt{Nvidia Jetson TX2}). \b{Note that the computational cost of a traditional linear MPC is comparable to the one of its linear RTMPC variant\cite{mayne2005robust}, further highlighting the computational benefits of our approach when compared to traditional MPC}.

\begin{figure}
\captionsetup[sub]{font=footnotesize}
\centering
\begin{subfigure}{\columnwidth}
    \centering
    \includegraphics[width=\columnwidth, trim={2cm 2cm 2cm 0.5cm},clip]{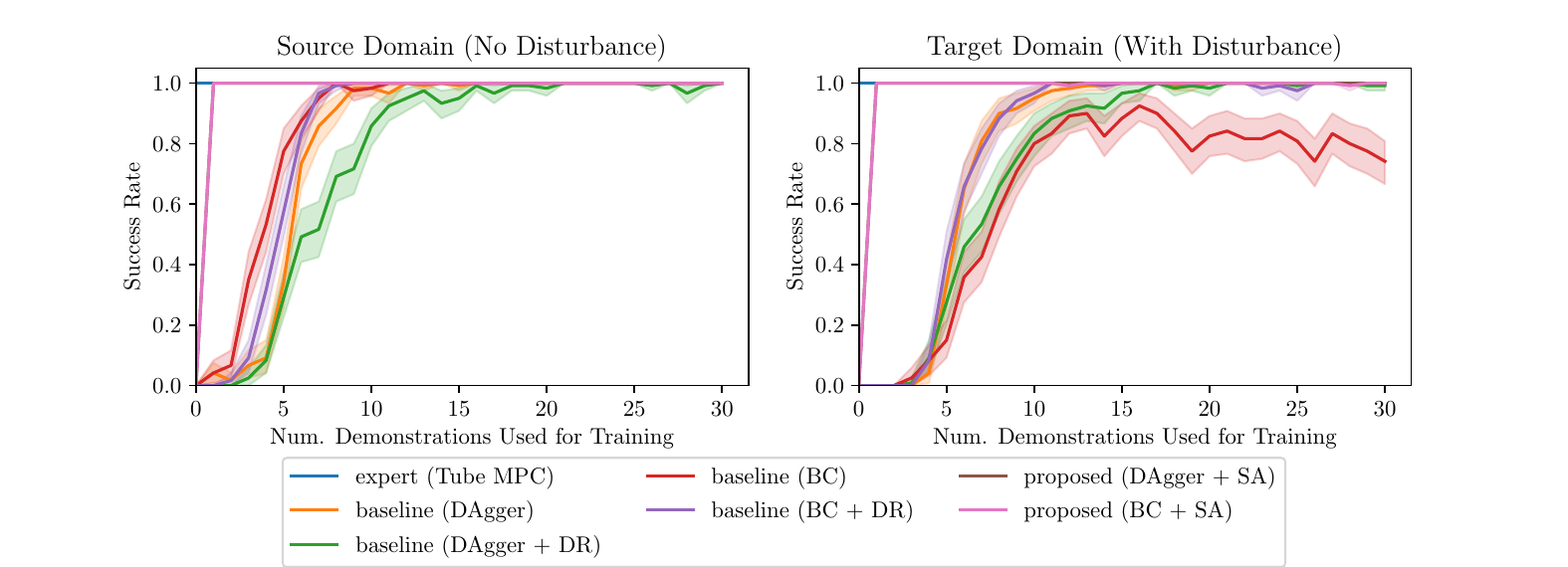}
\end{subfigure}%
\hspace{0.1cm}
\begin{subfigure}{\columnwidth}
    \centering
    \includegraphics[width=\columnwidth, trim={2cm 0 2cm 0},clip]{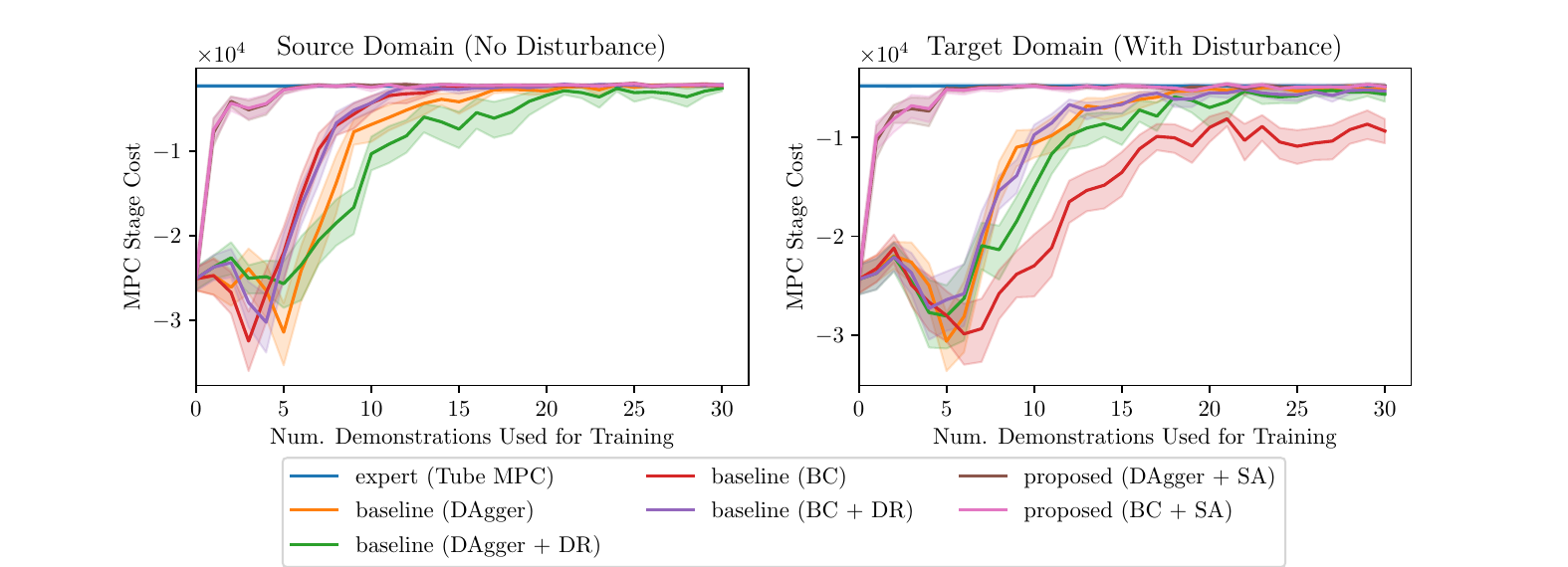}
\end{subfigure}
    \caption{Robustness (\textit{Success Rate}) and performance (\textit{MPC Stage Cost}) of SA (with $95\%$ confidence interval), compared to the number of demonstrations used for training. The task is tracking previously unseen trajectories, without and with wind-like disturbances. The proposed SA-sparse strategy learns and generalize to unseen trajectories with fewer demonstrations. The lines for SA-based methods overlap. Evaluation across $20$ randomly sampled trajectories per demonstration, for $6$ random seeds.} %
    \label{fig:learning_multiple_trajectories}
    \centering
    \ifblue
    \begin{tikzpicture}
    \node[inner sep=0pt] (figure) at (0,0) {
    \fi
    \includegraphics[width=\columnwidth, trim={1.2cm, 0.5cm, 1.2cm, 0.5cm}, clip]{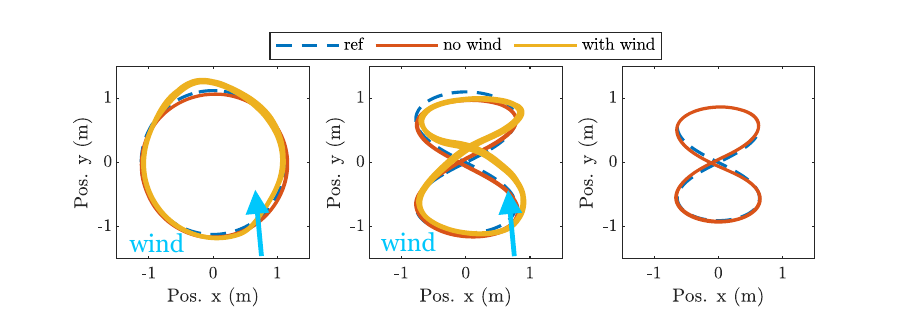}
    \ifblue
    };
    \draw[blue, very thick] (figure.south west) rectangle (figure.north east);
    \end{tikzpicture}
    \fi
    \caption{Examples of different trajectories arbitrary chosen from the training distribution, and tested in hardware experiments with and without strong wind-like disturbances produced by leaf blowers. The employed policy is trained with $10$ demonstrations (when other baseline methods have not fully converged yet, see \cref{fig:learning_multiple_trajectories}) using DAgger+SA (sparse). This highlights that SA can learn multiple trajectories in a more sample-efficient way than other IL methods, retaining \ac{RTMPC}'s robustness and performance.}
    \label{fig:learn_multiple_trajectories_experiment}
    \vspace{-3ex}
\end{figure}

\subsection{Numerical and Hardware Evaluation for Learning and Generalizing to Multiple Trajectories}
We evaluate the ability of the proposed approach to track multiple trajectories while generalizing to unseen ones. To do so, we define a training distribution of reference trajectories (circle, position step, figure-8) and a distribution for these trajectory parameters (radius, velocity, position). 
During training, we sample at random a desired, $7$ s long ($70$ steps) reference with randomly sampled parameters, collecting a demonstration and updating the proposed policy, while testing on a set of $20$, $7$ s long trajectories randomly sampled from the defined distributions. We monitor the robustness and performance of the different methods, with force disturbances (from $\mathbb{W}_{\mathcal{T}_1}$) applied in the target domain. The results of the numerical evaluation, shown in \cref{fig:learning_multiple_trajectories}, confirm that \ac{SA}-sparse
\begin{inparaenum}[i)]
    \item achieves robustness and performance comparable to the expert in a sample efficient way, requiring fewer than half the number of demonstrations needed for the baseline approaches; 
    \item simultaneously learns to generalize to multiple trajectories randomly sampled from the training distribution.
\end{inparaenum}
\b{Note that at convergence (from demonstration $20$ to $30$), DAgger+SA achieves the closest performance to the expert ($2.7\%$ \textit{expert gap}), followed by BC+SA ($3.0\%$  \textit{expert gap})}.
The hardware evaluation, performed with DAgger+SA-sparse ($10$ demonstrations), is shown in \cref{fig:learn_multiple_trajectories_experiment}. It confirms that the obtained policy is experimentally capable of tracking multiple trajectories under real-world disturbances/model errors.

\b{\subsection{Extra Comparisons and Hyperparameter Study}}
\begin{figure}
  \centering
  \ifblue
  \begin{tikzpicture}
    \node[inner sep=0pt] (figure) at (0,0) {
    \fi
  \includegraphics[width=\columnwidth, trim={0, 0.1cm, 0, 0}, clip]{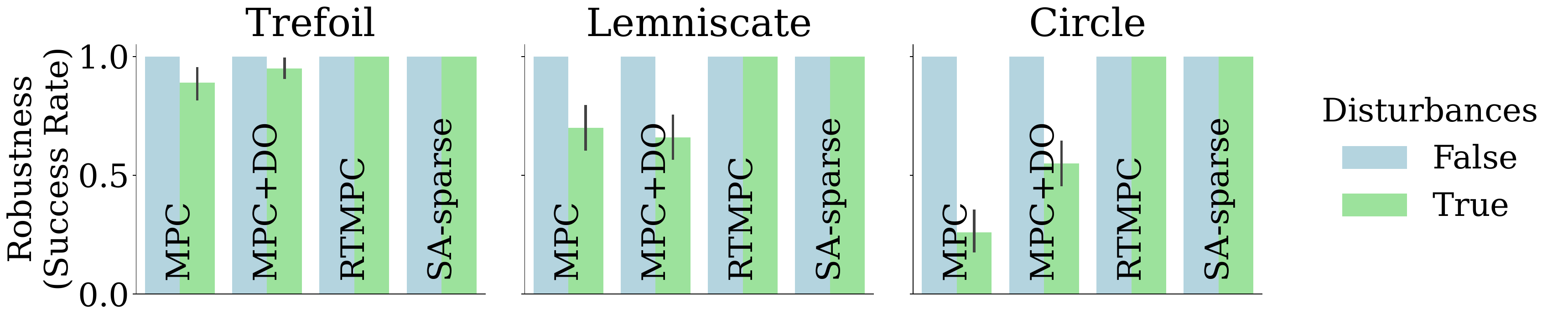}
  \ifblue
  };
    \draw[blue, very thick] (figure.south west) rectangle (figure.north east);
    \end{tikzpicture}
    \begin{tikzpicture}
    \node[inner sep=0pt] (figure) at (0,0) {
    \fi
    \includegraphics[width=\columnwidth, trim={0, 0.1cm, 0, 0}, clip]{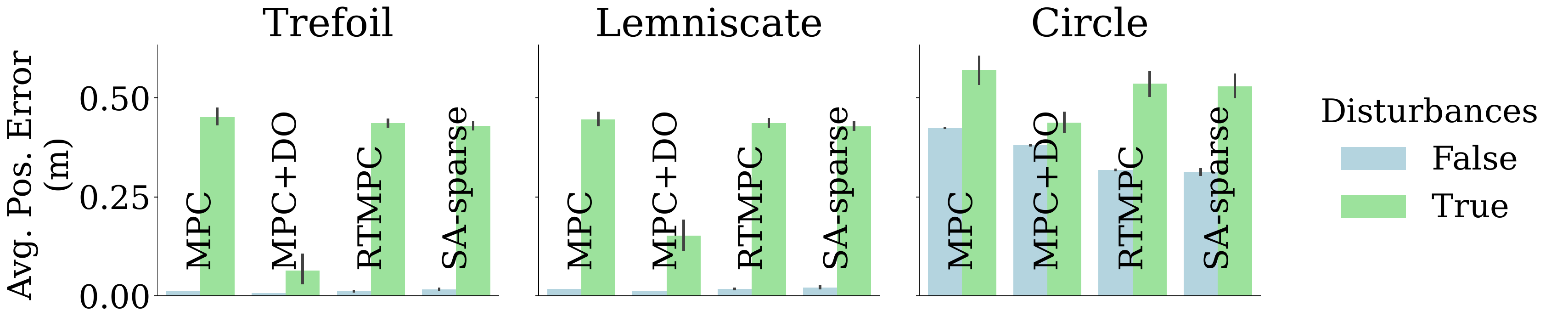}
    \ifblue
    };
    \draw[blue, very thick] (figure.south west) rectangle (figure.north east);
    \end{tikzpicture}
    \fi
  \caption{Comparison of performance and robustness of a traditional MPC, MPC combined with a disturbance observer (MPC+DO), robust tube MPC (RTMPC), and the policy learned from RTMPC (SA-sparse, one demonstration). The learned policy inherits the superior robustness properties of RTMPC, while achieving comparable or better performance than MPC. We consider two scenarios, with wind disturbances (True) or without (False).}
  \label{fig:expert_comparison}
\end{figure}

\noindent
\b{\textbf{Comparison with Other Optimal Control Approaches.}
SA-sparse (single demonstration) is compared in simulation with: 
\begin{inparaenum}
    \item a linear trajectory tracking MPC, based on \cite{kamel2017linear} (denoted \textbf{MPC}),
    \item the same MPC combined with a disturbance observer (Kalman filter) that estimates online additive force disturbances, updating the model used by the MPC \cite{kamel2017linear} (denoted \textbf{MPC+DO}), and 
    \item RTMPC (expert). 
\end{inparaenum}
The considered trajectories include position, velocity and actuation constraints, have velocities ranging in $2.0-3.5$ m/s, and have $30$ s duration each.
The results are presented in \cref{fig:expert_comparison} and highlight that, while the adaptive variant of MPC (MPC+DO) achieves the lowest average tracking errors, RTMPC is superior in terms of robustness (constraint satisfaction), while the learned policy successfully inherits the robustness properties of RTMPC, with minimal trade-offs in terms of position errors.}

\begin{figure}
    \centering
    \includegraphics[width=\columnwidth]{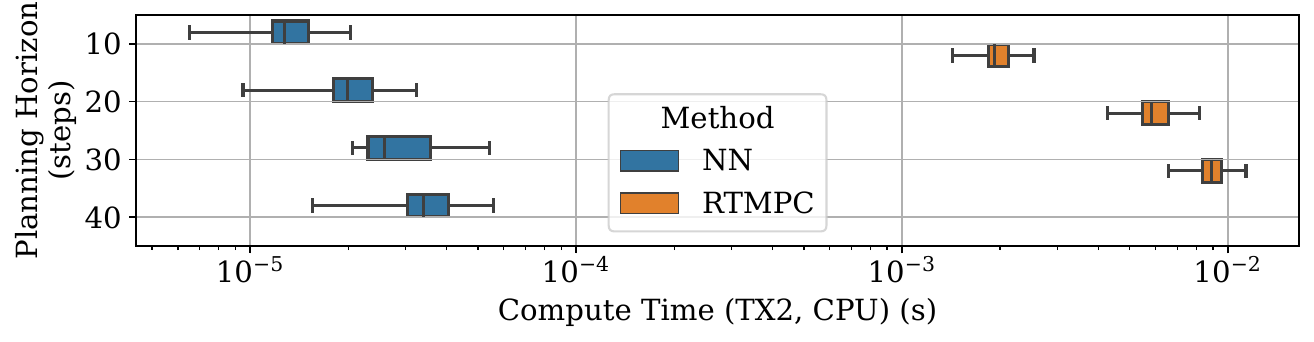}
    \caption{Computational cost of the DNN policy (two hidden layers, $32$ neurons/layer, C++) and the onboard \ac{RTMPC} expert (\texttt{CVXGEN}, C++) as a function of the planning horizon length. The DNN achieves over $2$ orders of magnitude improvement in computational efficiency.}
    \label{fig:nn_compute_vs_horizon}
\end{figure}

\begin{table}[]
    \caption{Comparison of DNN architectures for SA-sparse policies (one demonstration). The result shows low sensitivity to the choice of DNN architecture.}
    
    \renewcommand{\arraystretch}{1.2} %
    \setlength{\tabcolsep}{3pt}
        \resizebox{1\columnwidth}{!}{
    \centering
    \begin{tabular}{lcccccccc}
     \multirow{1}{*}{\makecell{NN\\(Neurons/Layer)}}
     & \multicolumn{2}{r}{\makecell{Robustness \\Succ. Rate (\%)}}
     & \multicolumn{2}{r}{\makecell{Performance\\Pos. Error (m)}} 
     & \multicolumn{2}{r}{\makecell{Performance\\Expert Gap (\%)}}
     & \multicolumn{2}{r}{\makecell{Computation\\(ms, TX2, CPU)}} \\
     & mean & std & mean & std & mean & std & mean & std \\
    \hline
    \hline
    [32, 32]        & 99.9  & 3.3 & 0.29 & 0.19 & 5.4 & 5.3 & 0.021 & 0.01 \\
    
    [64, 32]        & 100.0 & 0.0 & 0.30 & 0.20 & 4.9 & 4.9 & 0.030 & 0.01 \\
    
    [64, 64]        & 99.9  & 2.4 & 0.30 & 0.20 & 4.8 & 5.0 & 0.033 & 0.01 \\

    [64, 64, 32]    & 99.6  & 6.2 & 0.29 & 0.19 & 5.3 & 6.2 & 0.039 & 0.01 \\
    
    [128, 128]      & 99.9  & 3.3 & 0.30 & 0.20 & 4.8 & 5.2 & 0.163 & 0.05 \\
    \end{tabular}
        } %

    \label{tab:nn_architecture}
\end{table}

\begin{figure}
    \centering
    \includegraphics[width=\columnwidth, trim={0.1cm, 0.2cm, 0.1cm, 0.8cm}, clip]{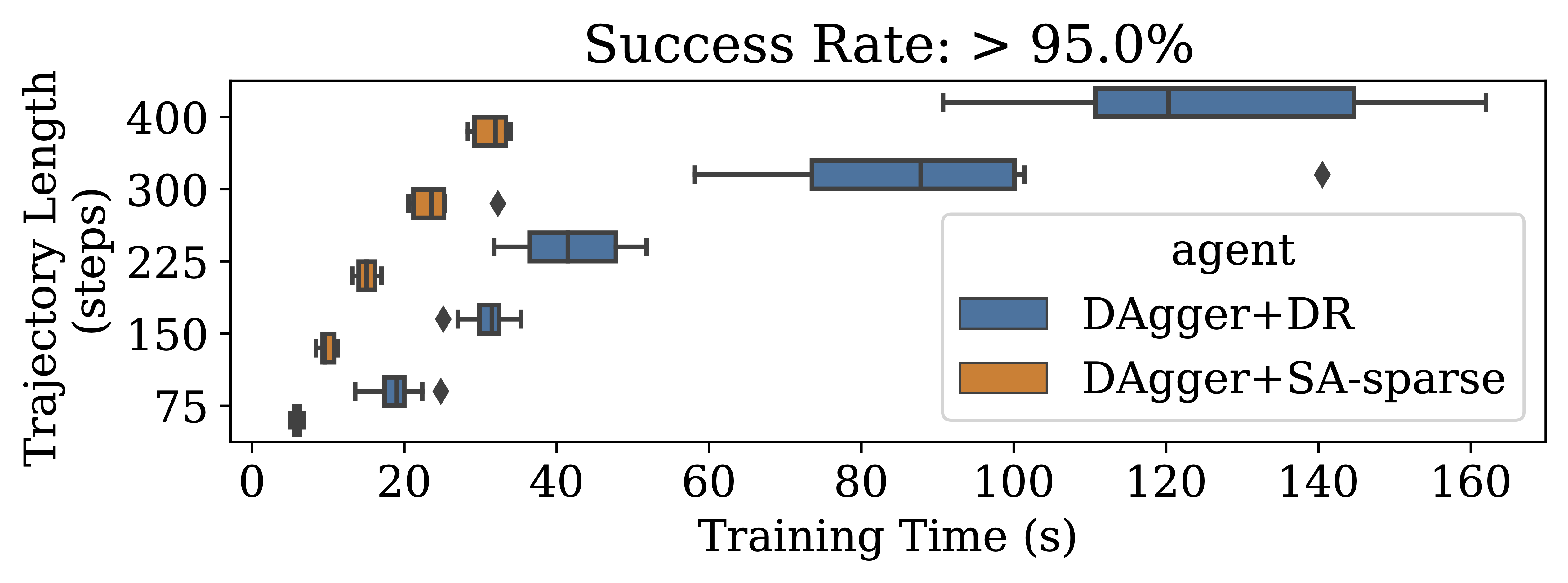}
    \caption{Time to generate a policy (data collection in simulation and training) compared to the complexity (number of time-steps in the trajectory) of the mission to be learned. The results highlight that SA-sparse enables learning of robust policies  in significantly lower time than DAgger+DR, and it scales better as the length of the task increases. Note: one step corresponds to $0.1$ s. Trajectory: eight-shaped (Lemniscate) followed by a vertical circular trajectory, with velocities up to $3.5$ m/s.}
    \label{fig:training_time_vs_traj_len}
\end{figure}

\b{
\noindent
\textbf{Hyperparameter Study}.
First, \cref{fig:nn_compute_vs_horizon} studies the effects on onboard computation when varying the planning horizon, highlighting (1) two-orders-of-magnitude improvements in the onboard computation of the \ac{DNN} policy compared to the RTMPC expert, and that (2) the computational benefits of the policy increase as the planning horizon increases.  
Second, \cref{tab:nn_architecture} studies robustness, performance, and onboard computation of the learned policy as a function of the size (layers, hidden neurons/layer) of \ac{DNN}, highlighting that robustness and performance are minimally affected by these parameters. While onboard computation grows with the size of the network, it remains significantly lower than the onboard RTMPC expert.
Last, \cref{fig:training_time_vs_traj_len} studies the time required to train a policy that achieves a success rate $> 95\%$ under wind, as a function of task complexity (length of the trajectory). This result highlights significant improvements in the scalability of our method when compared to the most robust baseline, DAgger+DR.}

\section{Evaluation - Learning From Nonlinear RTMPC} \label{sec:evaluation_nonlinear}
In this Section, we evaluate the ability of our method to efficiently learn acrobatic flight maneuvers using demonstrations collected from nonlinear \ac{RTMPC}. 

\subsection{Evaluation Approach}
\noindent \textbf{Task Description.}
The goal is to perform a flip, i.e., a $360^\circ$ rotation about the body-frame $x$-axis, in near-minimum time. This is a challenging maneuver, as it covers a large nonlinear envelope of the dynamics of the \ac{MAV}, and the near-minimum time objective function, combined with the need to account for uncertainties, pushes the actuators close to their physical limits. 

\noindent
\textbf{Simulation Environment.} The simulation environment for training/numerical evaluations is the same as in \cref{sec:evaluation_linear}, i.e., implements the nonlinear multirotor model in \cref{subsec:mav_model}. In the training domain (source, $\mathcal{S}$), $\mathbb{W}_\mathcal{S} = \{ \emptyset \}$, while in the deployment domain (target, $\mathcal{T}$) $\mathbb{W}_\mathcal{T} = \{f_\text{ext} | 0.001 m g \leq f_\text{ext} \leq 0.3 m g \}$, sampled according to \cref{eq:disturbance_sampling}.

\noindent \textbf{Nonlinear \ac{RTMPC}.}
We generate a safe nominal flip trajectory using \texttt{MECO-Rockit} \cite{gillis2020effortless} and \texttt{IPOPT}. Because this nominal trajectory happens in the plane spanned by the orthogonal vectors defining the $y$ and $z$ axis of the inertial reference frame $\text{W}$, for simplicity, we project the dynamics onto the $y$ and $z$ axes, resulting in a two-dimensional model of the \ac{MAV} used to generate the nominal plan. The nominal flip trajectory can therefore be obtained by setting the initial rotation around the $z$ to be $0$, and the desired final attitude to be $2 \pi$, while the remaining initial/terminal states are all set to zero. 

The ancillary \ac{NMPC} is solved using the \ac{SQP} solver \texttt{ACADOS} \cite{Verschueren2021}, and runs in simulation at $50$ Hz. Sensitivities for \ac{DA} (\cref{eq:tangential_predictor_qp}) are computed using the built-in sensitivity computation in the chosen solver, \texttt{HPIPM} \cite{frison2020hpipm}.
We remark that the employed ancillary \ac{NMPC} uses the full 3D multirotor model in \cref{eq:mav_reduced_model_for_nmpc}, therefore performing 3D disturbance rejection -- a critical requirement for real-world deployments. %
For a more challenging and interesting comparison to the considered \ac{IL} baselines, the ancillary \ac{NMPC} uses the \texttt{SQP\_RTI} setting of \texttt{ACADOS}. This setting performs only a single \ac{SQP} iteration per timestep, enabling significant speed-ups in the solver, and it is often employed in real-time, embedded implementations of \ac{NMPC}. This setting creates an advantage, in terms of training time, to \ac{IL} methods that require querying the expert multiple times (the baselines of our comparison), as it speeds-up the computation time of the expert. 
The other \texttt{ACADOS} parameters given in \cref{tab:acados_parameters} were chosen as they enabled higher overall performance/accuracy in the selected acrobatic maneuver. Last, we introduce a discount factor $\gamma = 0.95$ in the stage cost of \cref{eq:ancillary_nmpc_eq} to aid the convergence of the solver.

\noindent \textbf{Student Policy.}
The student is a $2$-hidden layers, fully connected \ac{DNN} with $\{64, 32\}$ neurons/layer, and \texttt{ReLU} activation functions. The input has dimension $14$, as it contains the current state ($n_x = 10$), time $t$, and a desired final position $\vbs{p}^\text{des}$ (fixed to the origin). To simplify the learning and \ac{DA} procedure, we enforce continuity to the quaternion input of the policy via \cite[Eq. 3]{kusaka2022stateful}, avoiding the need to increase the training data/demonstrations at every timestep to account for the fact that $\vbs{q}$ and $-\vbs{q}$ encode the same orientation.
\begin{table}
\caption{Parameters for the ancillary \ac{NMPC}, solved via \texttt{ACADOS} \cite{Verschueren2021}.}
\vspace*{-0.1in}
\begin{center}
\begin{tabular}{ r c }
Parameter(s) & Value(s) \\
\hline 
\hline 
Hessian Approximation  & Gauss-Newton \\
QP solver & Partial Condensing \texttt{HPIPM} \cite{frison2020hpipm} \\
NLP/QP Tolerance &  $10^{\ensuremath{-}8}$/$10^{\ensuremath{-}8}$ \\
Levenberg-Marquardt & $10^{\ensuremath{-}4}$\\
Integrator Type & Implicit Runge-Kutta  \\
Max. \# Iterations QP Solver & $100$ \\
Horizon ($N$, steps)/(time, seconds) & $50$/$1.0$ \\
\end{tabular}
\label{tab:acados_parameters}
\end{center}
\vskip-5ex
\end{table}

\noindent
\textbf{Baselines and Evaluation Metrics} The baselines match those in \cref{sec:evaluation_linear} (\ac{DAgger}, \ac{BC} and their combination with \ac{DR}, \b{DA-$N_s$ (linear interpolation) and DA-$N_s$ (expert neighborhood), generating $N_s$ samples per timestep. The monitored metrics (\textit{Robustness}, \textit{Performance} and \textit{Training Time}) match those in \cref{sec:evaluation_linear}}, with the difference that performance is based on the stage cost of the ancillary \ac{NMPC} \cref{eq:ancillary_nmpc_eq}. 

\noindent \textbf{Training Details.}
As in \cref{sec:evaluation_linear}, training is performed by collecting demonstrations with the multirotor starting from slightly different initial states inside the tube centered around the origin. The nominal flip maneuver is pre-generated, as the goal state $\vbf{x}^e_{\tzr}$ (with $\tzr = 0)$ does not change, and only the ancillary \ac{NMPC} is solved at every timestep. The resulting flip maneuver takes about $T_f = 2.5$ s, and demonstrations are collected over an episode of length $3.0$ s, at $50$ Hz ($T = 150$ environment steps per demonstration).
\textit{Data collection.} \b{For \ac{SA}-methods, we collect demonstrations one-by-one, and we implement the fine-tuning procedure described in \cref{subsec:rob_perf_under_approx_samples} by performing \ac{DA} with the first collected demonstration, while we do not perform \ac{DA} for the following demonstrations. Because of its computational efficiency, we always use the sensitivity-based \ac{DA} (i.e., \cref{eq:approximate_ancillary_controller}, assuming no changes in active set of constraints). In addition, to study the effects of varying the number of samples used for \ac{DA}, we introduce \ac{SA}-$N_S$, a variant of \ac{SA} where we sample uniformly inside the tube $N_S = \{25, 50, 100\}$ samples for every timestep.
For the baselines, in order to speed-up the demonstration collection phase, and thereby avoid excessive re-training of the policy, we collect demonstrations in batches of $10$, for $20$ batches. An exception is made for \ac{DA}-$N_s$ (expert neighborhood) where, similar to \ac{SA}, we collect demonstrations one-by-one to better study its sample efficiency, and the corresponding actions are obtained using the same sensitivity-based approximation of the ancillary \ac{NMPC} employed by \ac{SA}. Therefore, \ac{DA}-$N_s$ (expert neighborhood) constitutes an ablation of \ac{SA}-$N_s$, where sampling is restricted to a smaller volume than the entire tube.} 
\textit{Evaluation} Each time the policy is updated, we evaluate it $20$ times in source $\mathcal{S}$ and target $\mathcal{T}$ environments. The evaluations are repeated across $10$ random seeds. To further speed-up training of all the methods, we update the previously trained policy using only the newly collected batch of demonstrations (or single demonstration, for \ac{SA}). All the policies are trained using the ADAM optimizer for up to $400$ epochs, but we terminate training if the validation loss (from $30\%$ of the data) does not decrease within $30$ epochs. %

\subsection{Numerical Evaluation: Robustness, Performance, Efficiency} 

\begin{figure}
    \centering
    \vspace{1ex}
    \includegraphics[width=\columnwidth, trim={3cm 0.5cm 3cm 1cm}, clip]{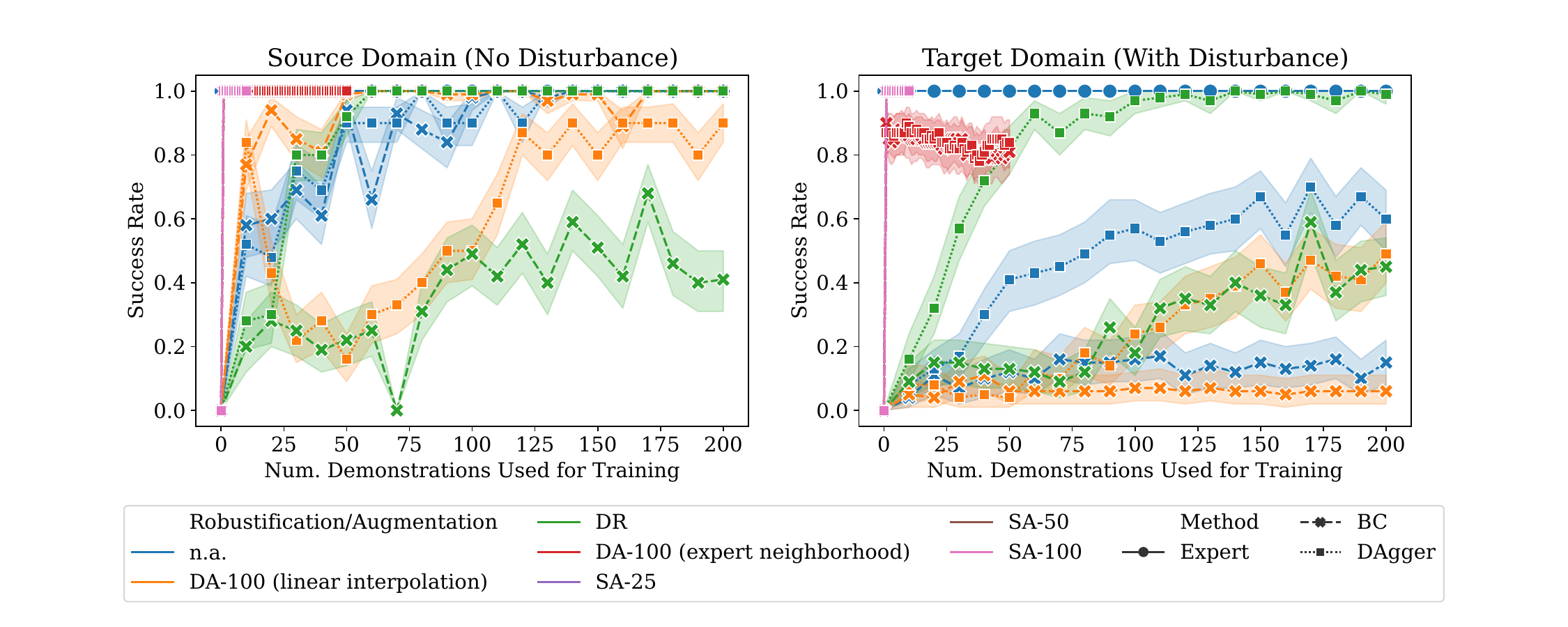}
    
    \caption{Robustness as a function of the number of training demonstrations. The proposed SA-methods overlap on the top-left part of the diagram, achieving full success rate in both the environment without and with wind-like disturbances.}
    \label{fig:nmpc_robustness}

    \centering
    \vspace{1ex}
    \includegraphics[width=\columnwidth, trim={3cm 0.5cm 3cm 1cm}, clip]{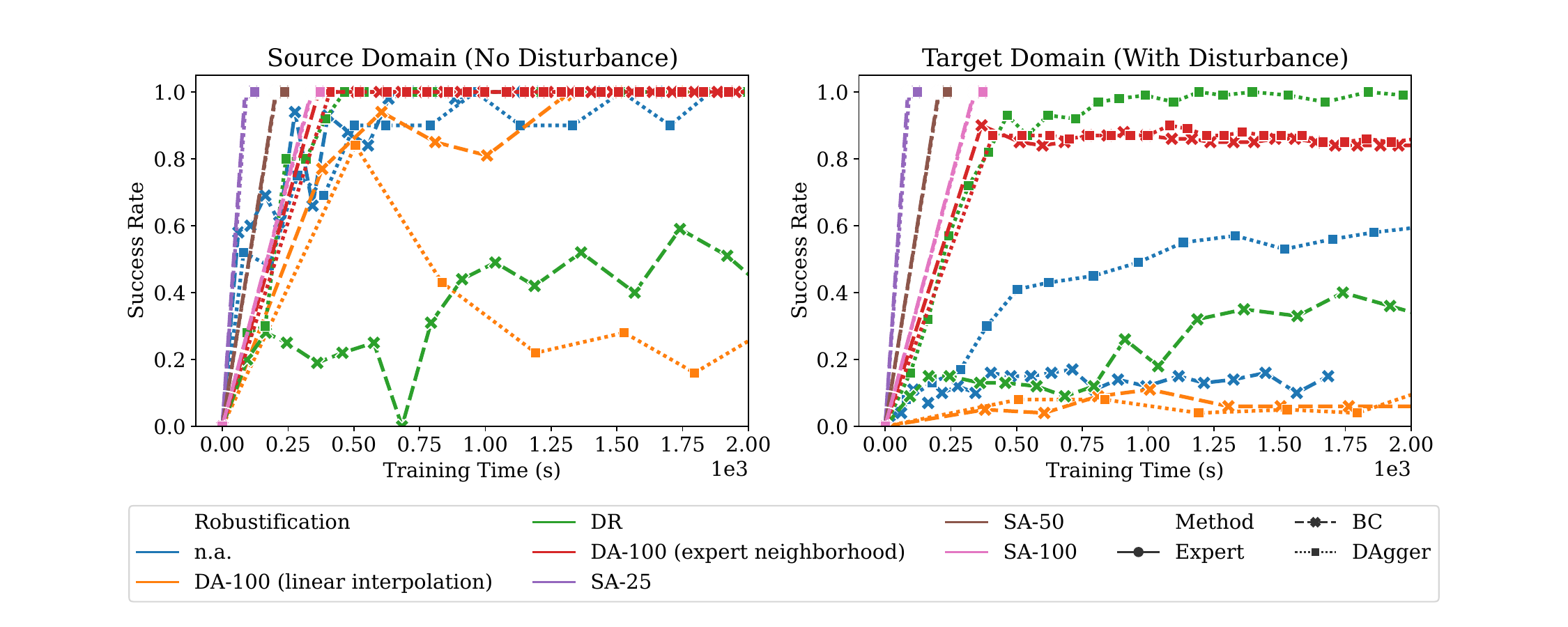}
    \caption{Robustness as a function of the training time. SA-methods achieve full robustness under uncertainties in a fraction of the training time required by the best performing robust baseline, DAgger + DR.}
    \label{fig:nmpc_efficiency}

    \centering
    \vspace{1ex}
    \includegraphics[width=\columnwidth, trim={3cm 0.5cm 3cm 1cm}, clip]{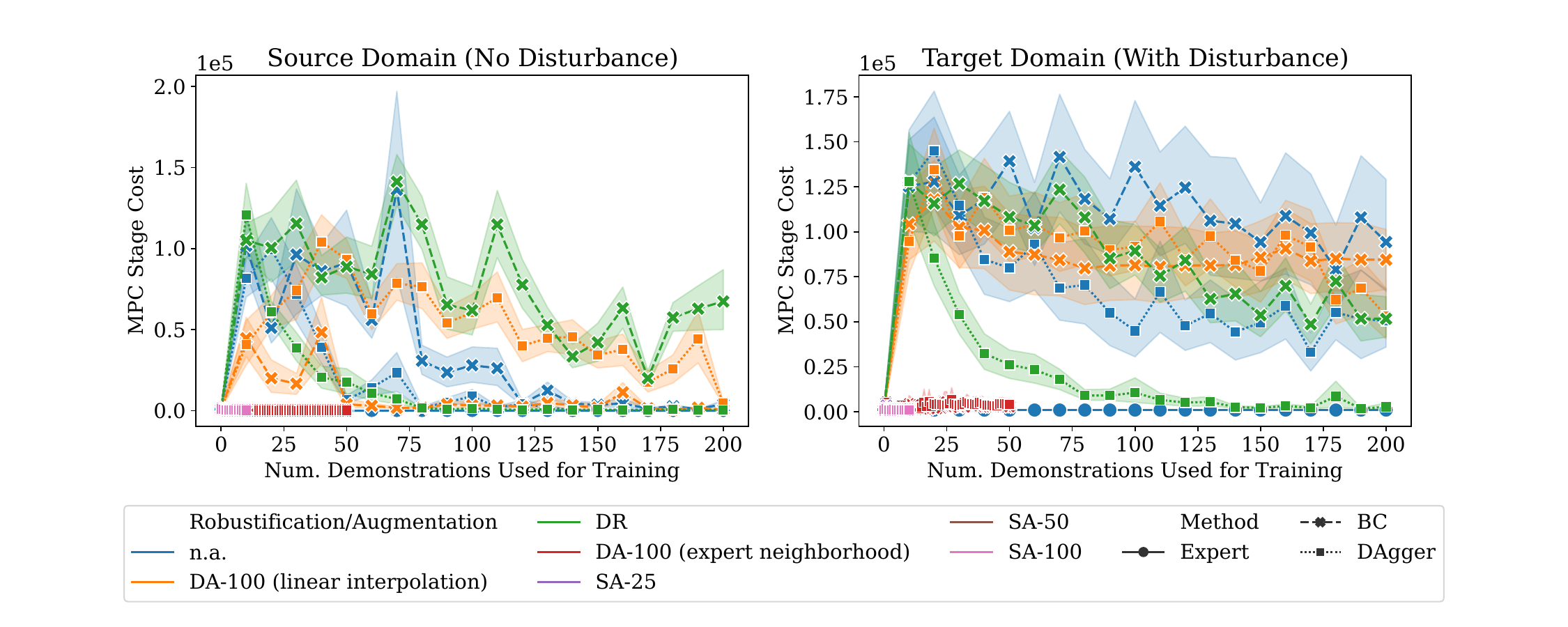}

    \caption{Performance as a function of the number of training demonstrations. SA-methods achieve performance close to the expert in less than $10$ demonstrations. The best-performing baseline, DA-100 (expert neighborhood), achieves comparable performance, but is not robust when subject to wind disturbances, as shown in Fig. \ref{fig:nmpc_robustness}.}
    \label{fig:nmpc_performance}
    \vspace{-3ex}
\end{figure}
\noindent

\noindent \textbf{Comparison with Baselines.}
We start by evaluating the robustness and performance of the proposed approach as a function of the number of demonstrations collected in simulation, and as a function of the training time. 

\cref{fig:nmpc_robustness} shows the robustness of the considered method as a function of the number of expert demonstrations. It reveals that \ac{SA}-based approaches can achieve full success rate in the environment with disturbances (target, $\mathcal{T}$) and without disturbances (source, $\mathcal{S}$) after a single demonstration, while the best-performing baseline, \ac{DAgger}+\ac{DR}, requires about $60$ demonstrations to achieve full robustness in $\mathcal{S}$, and more than $100$ in $\mathcal{T}$. \ac{SA}-based methods, therefore, enable more than one order of magnitude reduction in the number of demonstrations (interactions with the environment) compared to \ac{DAgger}+\ac{DR}.
As previously observed in \cref{sec:evaluation_linear}, \ac{DAgger} alone is not robust. 
Additionally, \ac{BC} methods fail to converge, potentially due to the lack of sufficiently meaningful exploration and the forgetting caused by the iterative training strategy employed. 
In addition, DA (expert neighborhood) confirms the importance of using the tube as a support of the sampling distribution, as the method achieves robustness and demonstration efficiency in the source domain, but fails to achieve robustness in the target domain, unlike SA. DA (linear interpolation) offers an initial boost in demonstration efficiency but struggles to achieve high robustness even within the source domain. This may be due to the introduction of far-from optimal actions.

\cref{fig:nmpc_efficiency} additionally shows the robustness as a function of the training time (recall, this includes demonstration collection and policy train). The results show that the demonstration-efficiency of \ac{SA}-based methods translates into significant improvements in training time, as \ac{DAgger}+\ac{DR} requires more than $3$ times the training time than \ac{SA}-based approaches. These improvements are larger for the variants of \ac{SA} that generate fewer extra samples (e.g., \ac{SA}-$25$).

Last, \cref{fig:nmpc_performance} reports the \textit{performance} as a function of the number of demonstrations. The results indicate that \ac{SA}-based methods can achieve low tracking errors even after a single demonstration. Furthermore, employing a fine-tuning phase (after the initial demonstration) proves highly advantageous in further reducing this error, thereby reducing the performance gap between policies obtained via \ac{SA} and the expert.

\begin{table}
    \tiny
    \renewcommand{\tabcolsep}{1pt}
    \centering
    \tiny

\def\thinline{\noalign{\hrule height.05pt}} 

\newcommand*{\opacity}{40}%

\definecolor{hight}{HTML}{ec462e} 
\definecolor{lowt}{HTML}{76f013}  %
\newcommand*{\minvalt}{300.0}%
\newcommand*{\maxvalt}{600.0}%
\newcommand{\grdt}[1]{
    \pgfmathparse{int(round(100*(min(max(#1, \minvalt), \maxvalt)/(\maxvalt-\minvalt))-(\minvalt*(100/(\maxvalt-\minvalt)))))}
    \xdef\tempa{\pgfmathresult}
    \cellcolor{hight!\tempa!lowt!\opacity} #1
}

\definecolor{highp}{HTML}{ec462e}
\definecolor{lowp}{HTML}{76f013}
\newcommand*{\minvalp}{30}%
\newcommand*{\maxvalp}{800}%
\newcommand{\grdp}[1]{
    \pgfmathparse{int(round(100*(min(max(#1, \minvalp), \maxvalp)/(\maxvalp-\minvalp))-(\minvalp*(100/(\maxvalp-\minvalp)))))}
    \xdef\tempa{\pgfmathresult}
    \cellcolor{highp!\tempa!lowp!\opacity} #1
}

\definecolor{highr}{HTML}{ffffff}
\definecolor{lowr}{HTML}{ec462e}
\newcommand*{\minvalr}{90}%
\newcommand*{\maxvalr}{100}%
\newcommand{\grdr}[1]{
    \pgfmathparse{int(round(100*(min(max(#1, \minvalr), \maxvalr)/(\maxvalr-\minvalr))-(\minvalr*(100/(\maxvalr-\minvalr)))))}
    \xdef\tempa{\pgfmathresult}
    \cellcolor{highr!\tempa!lowr!\opacity} #1
}

\newcommand{\gradientcell}[6]{
    \ifdimcomp{#1pt}{>}{#3 pt}{#1}{
        \ifdimcomp{#1pt}{<}{#2 pt}{#1}{
            \pgfmathparse{int(round(100*(#1/(#3-#2))-(\minval*(100/(#3-#2)))))}
            \xdef\tempa{\pgfmathresult}
            \cellcolor{#5!\tempa!#4!#6} #1
    }}
}

\newcolumntype{P}[1]{>{\centering\arraybackslash}p{#1}}

\begin{tabular}{|P{0.6cm}P{1.1cm}P{0.75cm}||P{0.5cm}P{0.5cm}|P{0.5cm}P{0.5cm}|P{0.5cm}P{0.5cm}|P{0.5cm}P{0.5cm}|P{0.5cm}P{0.5cm}|}%
\hline
{}     & 
{}     & 
{}     & 
\multicolumn{4}{c|}{\makecell{Robustness\\success rate\\($\%$, $\uparrow$)}} & 
\multicolumn{4}{c|}{\makecell{Performance\\expert gap\\($\%$, $\downarrow$)}} & 
\multicolumn{2}{c|}{\makecell{Efficiency\\training time\\($s$, $\downarrow$)}} 
\\
Method & 
\multirow{2}{*}{\makecell{Robustification/\\\newline Augmentation}} & 
\# of Demonstr. & 
\multicolumn{2}{c|}{$\mathcal{S}$} & 
\multicolumn{2}{c|}{$\mathcal{T}$} & 
\multicolumn{2}{c|}{$\mathcal{S}$} & 
\multicolumn{2}{c|}{$\mathcal{T}$} &
\multicolumn{2}{c|}{$-$} 
\\
&       & {} &    mean & std & mean & std &  mean &   std &  mean &   std &        mean & std \\
\hline
\hline
BC & \multirow{2}{*}{\makecell{DA-100 (expert\\neighborhood)}} 
              & 1   &     100 &   0 &   \grdr{90} &  30 &          \grdp{9} &     6 &   \grdp{562} &  1576 &         \grdt{367} &  87 \\
       &      & 2   &     100 &   0 &   \grdr{85} &  36 &          \grdp{6} &     4 &   \grdp{312} &   671 &         \grdt{512} &  90 \\
       &      & 10  &     100 &   0 &   \grdr{86} &  35 &          \grdp{5} &     4 &   \grdp{462} &  2932 &        \grdt{1166} & 135 \\
       &      & 50  &     100 &   0 &   \grdr{81} &  39 &          \grdp{5} &     2 &   \grdp{323} &   838 &        \grdt{4629} & 185 \\
\hline
DAgger & \multirow{2}{*}{\makecell{DA-100 (expert\\neighborhood)}} 
              & 1   &     100 &   0 &   \grdr{87} &  34 &         \grdp{12} &     8 &   \grdp{562} &  2375 &         \grdt{409} &  90 \\
       &      & 2   &     100 &   0 &   \grdr{87} &  34 &         \grdp{9} &    10 &   \grdp{425} &  1478 &         \grdt{520} &  97 \\
       &      & 10  &     100 &   0 &   \grdr{89} &  31 &         \grdp{4} &     3 &   \grdp{187} &   354 &        \grdt{1149} & 107 \\
       &      & 50  &     100 &   0 &   \grdr{84} &  37 &         \grdp{3} &     2 &   \grdp{440} &  1623 &        \grdt{4608} & 141 \\
\hline
DAgger & DR    & 50  &      \grdr{92} &  27 &   \grdr{82} &  39 &       \grdp{9094} & 20608 &  \grdp{3096} &  5497 & \grdt{392} &  34 \\
       &       & 100 &     100 &   0 &   \grdr{97} &  17 &        \grdp{634} &   711 &  \grdp{1277} &  4947 & \grdt{810} & 104 \\
       &       & 200 &     100 &   0 &   \grdr{99} &  10 &         \grdp{91} &    73 &   \grdp{274} &  1247 & \grdt{1970} & 154 \\
\hline
BC     & SA-sparse (18)&1& 100 &   0 &  100 &   0 &        \grdp{553} &   462 &   \grdp{211} &   228 &          \grdt{84} &   9 \\
       &       & 2   &     100 &   0 &   \grdr{97} &  17 &         \grdp{41} &    39 &   \grdp{371} &  1808 &          \grdt{88} &  10 \\
       &       & 10  &     100 &   0 &   \grdr{97} &  17 &         \grdp{33} &    42 &   \grdp{226} &   815 &         \grdt{115} &  10 \\
\thinline
       & SA-25 & 1   &     100 &   0 &  100 &   0 &        \grdp{956} &   402 &   \grdp{270} &   384 &          \grdt{87} &  15 \\
       &       & 2   &     100 &   0 &  100 &   0 &        \grdp{148} &   140 &   \grdp{107} &   150 &          \grdt{90} &  14 \\
       &       & 10  &     100 &   0 &  100 &   0 &        \grdp{107} &   175 &    \grdp{90} &    83 &         \grdt{117} &  14 \\
\thinline
       & SA-50 & 1   &     100 &   0 &  100 &   0 &        \grdp{421} &   193 &   \grdp{105} &   116 &         \grdt{204} &  32 \\
       &       & 2   &     100 &   0 &  100 &   0 &         \grdp{76} &    31 &   \grdp{66}&    56 &         \grdt{207} &  31 \\
       &       & 10  &     100 &   0 &  100 &   0 &         \grdp{55} &    28 &   \grdp{76}&   102 &         \grdt{235} &  31 \\
\thinline
       & SA-100& 1   &     100 &   0 &  100 &   0 &        \grdp{291} &   154 &    \grdp{89} &   105 &         \grdt{339} &  72 \\
       &       & 2   &     100 &   0 &  100 &   0 &         \grdp{57} &    20 &    \grdp{76} &    85 &         \grdt{342} &  72 \\
       &       & 10  &     100 &   0 &  100 &   0 &         \grdp{33} &    21 &    \grdp{97} &   118 &         \grdt{369} &  72 \\
\hline
DAgger & SA-sparse (18)&1& 100 &   0 &  100 &   0 &        \grdp{747} &   705 &   \grdp{319} &   879 &          \grdt{85} &   6 \\
       &       & 2   &     100 &   0 &  100 &   0 &        \grdp{222} &   122 &   \grdp{142} &   219 &          \grdt{89} &   6 \\
       &       & 10  &     100 &   0 &  100 &   0 &        \grdp{29}  &    24 &   \grdp{114} &   150 &         \grdt{117} &   6 \\
\thinline
       & SA-25 & 1   &     100 &   0 &  100 &   0 &        \grdp{579} &   224 &   \grdp{160} &   168 &          \grdt{92} &  12 \\
       &       & 2   &     100 &   0 &  100 &   0 &        \grdp{366} &   279 &   \grdp{122} &   182 &          \grdt{96} &  12 \\
       &       & 10  &     100 &   0 &  100 &   0 &        \grdp{110} &   115 &   \grdp{100} &   120 &         \grdt{124} &  12 \\
\thinline
       & SA-50 & 1   &     100 &   0 &  100 &   0 &        \grdp{361} &   161 &    \grdp{78} &    91 &         \grdt{206} &  28 \\
       &       & 2   &    100 &   0 &  100 &   0 &         \grdp{169} &   117 &    \grdp{77} &    80 &         \grdt{210} &  28 \\
       &       & 10  &     100 &   0 &  100 &   0 &         \grdp{56} &    77 &    \grdp{82} &   107 &         \grdt{237} &  28 \\
\thinline
       & SA-100 & 1   &     100 &   0 &  100 &  0 &        \grdp{309} &   133 &    \grdp{92} &   105 &         \grdt{342} &  29 \\
       &       & 2   &     100 &   0 &  100 &   0 &        \grdp{100} &    61 &    \grdp{77} &    96 &         \grdt{346} &  29 \\
       &       & 10  &     100 &   0 &  100 &   0 &         \grdp{30} &    28 &    \grdp{90} &   109 &         \grdt{373} &  29 \\
\hline
\end{tabular}

    \caption{Performance, robustness and training time for SA-based methods after $1$, $2$, and $10$ demonstrations, compared with the best performing baselines, DAgger+DR \b{and sampling in the expert neighborhood (DA-100 (expert neighborhood))}, in the environment without wind disturbances ($\mathcal{S}$, source), and with ($\mathcal{T}$, target). 
    Robustness is color-coded from white ($100\%$) to red ($90\%$ or below). Performance and training time are color-coded from green (fast training time, small expert gap) to red (long training time, large expert gap). The results highlight that \ac{SA}-methods achieve high robustness and close to expert performance compared to DAgger+DR, even after a single demonstration, and their performance can be further improved via additional fine-tuning demonstrations. \b{Methods based on DA-100 (expert neighborhood) are not robust and struggle to achieve high performance (low expert gap) in the target domain}. We note that \ac{DAgger} and \ac{BC}-based approaches differ at one demonstration due to non-determinism in the training procedure.}
    \label{tab:sa_analysis}
    \vspace{-2ex}
\end{table}

\noindent \textbf{Comparison of Sampling Strategies.}
\cref{tab:sa_analysis} provides a detailed comparison of performance, robustness, and training time of the different variants of \ac{SA} methods, as a function of the number of demonstrations ($1$, $2$ and $10$), and compares those with the best-performing baselines, \ac{DAgger}+\ac{DR}, \b{and methods based on sampling in the expert neighborhood}. As expected, \ac{SA} methods that require fewer samples obtain significant improvements in training time compared to \ac{DAgger}+\ac{DR}, while increasing the number of samples is beneficial in reducing the mean and the variance of the expert gap, both with and without disturbances\b{, while \ac{DA} (expert neighborhood) struggles to achieve high robustness and low expert gap in the target domain, despite the large number of samples used per timestep ($100$)}. \cref{tab:sa_analysis} additionally highlights the benefits of fine-tuning, as even methods that use few samples (e.g., SA-sparse, SA-$25$) can obtain a significant performance improvement after a single fine-tuning demonstration ($2$ demonstrations in total), while there are diminishing returns for additional fine-tuning demonstrations (e.g., $10$ demonstrations). \b{In addition, in the data-sparse regime (e.g., SA-$18$), using DAgger for fine-tuning   appears more beneficial than BC.}

\begin{figure}
    \centering
        \includegraphics[width=\columnwidth, trim={0, 0.25cm, 0, 0.2cm}, clip]{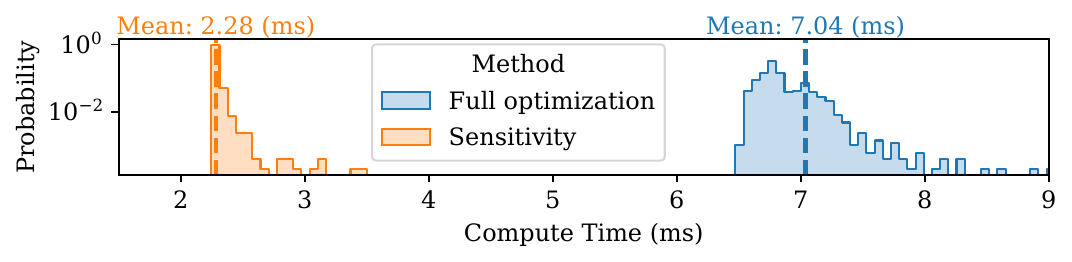}

    \caption{\b{Distribution of the time (ms) to solve the full optimization problem for the ancillary NMPC (\cref{eq:ancillary_nmpc_eq}), and to additionally compute the sensitivity matrix (\cref{eq:tangential_predictor_qp}). The sensitivity matrix require only $32\%$ of the time to solve the full optimization problem. Analysis performed on a \texttt{Intel i9-10920} using \texttt{ACADOS} with settings in \cref{tab:acados_parameters}. Note the log scale of the $y$ axis.}}
    \label{fig:compute_cost_of_sensitivity}
    \vskip-2ex
\end{figure}

\begin{table}[t]
    \renewcommand{\arraystretch}{1.4}
    
    \setlength{\tabcolsep}{0.9\tabcolsep} %
    \caption{\b{Time (ms) to compute a new action for the ancillary NMPC, the safe planner of the nonlinear RTMPC expert (N-RTMPC) and the proposed \ac{DNN} policy (Policy). \textbf{The policy is $\mathbf{180}$ times faster than the NMPC in \cite{sun2022comparative} (on the same onboard computer)}. The offboard CPU is an \texttt{Intel i9-10920}, onboard is an \texttt{NVIDIA Jetson TX2} (CPU). Note that the faster inference time than the linear case is caused by the input dimension being smaller ($14$ vs $188$).}}
    \label{tab:computational_cost_rtnmpc}    
    \centering
    \resizebox{1.0\columnwidth}{!}{
    \ifblue
    \begin{tikzpicture}
    \node[draw, blue, thick, inner sep=0pt] (table) {
    \fi
    \begin{tabular} {|C{1.0cm}| C{3.2cm} |C{1.6cm}|C{0.6cm}|C{0.5cm}|C{0.5cm}|C{0.5cm}| }
    \multicolumn{3}{c}{} & \multicolumn{4} {c}{Time (ms)} \\
    \hline
        \textbf{CPU} & \textbf{Method}  & \textbf{Setup} & \textbf{Mean} & \textbf{SD} & \textbf{Min} & \textbf{Max}\\
        \hline 
        \hline 
        \multirow{3}{*}{Offboard}& N-RTMPC, ancillary NMPC& \texttt{ACADOS} \cite{Verschueren2021} &  $7.28$ & $0.15$ & $7.05$ & $8.00$ \\
                                 & N-RTMPC, safe plan     & \texttt{IPOPT} &  $5812$ & $226$ & $4828$ & $6010$ \\
        \cline{2-7}
                                 & \textbf{Policy}               & PyTorch         & $\mathbf{0.11}$ & $\mathbf{0.01}$ & $\mathbf{0.11}$ & $\mathbf{0.27}$ \\
        \hline
        \hline
        \multirow{2}{*}{Onboard} & NMPC (from \cite[Fig. 17]{sun2022comparative}) & \texttt{ACADO}~\cite{Houska2011a} & $2.7$ & n.a. & n.a. & n.a. \\
        \cline{2-7}
                                & \textbf{Policy}              & C++/Eigen & $\mathbf{0.015}$ & $0.005$ & $0.006$ & $0.101$\\  
        \hline
    \end{tabular}
    \ifblue
    };
    \end{tikzpicture}
    \fi
    
    } %

\vskip-1ex
\end{table}
\noindent
\textbf{Computation.} The computation time is reported in \cref{tab:computational_cost_rtnmpc}, highlighting that \b{the average computation time of the policy on the onboard \texttt{Nvidia Jetsion TX2} is $0.015$ ms, an $180$-fold improvement compared to the value reported in \cite{song2023reaching} for a state-of-the-art NMPC for quadrotors.}
\b{In addition, \cref{fig:compute_cost_of_sensitivity} reports the time to compute the sensitivity matrix, highlighting that it requires on average $2.28$ ms, about only $32\%$ of the time required to solve the full optimization problem. More importantly, this matrix is computed only once per timestep and can be used to draw many sampled at small additional cost (equivalent to solving a vector-matrix multiplication) instead of solving the full optimization problem for each sample.} The average time to step the training environment is $2.1$ ms.

\begin{figure}
    \centering
    \begin{subfigure}{\columnwidth}
      \centering
      \includegraphics[width=\columnwidth]{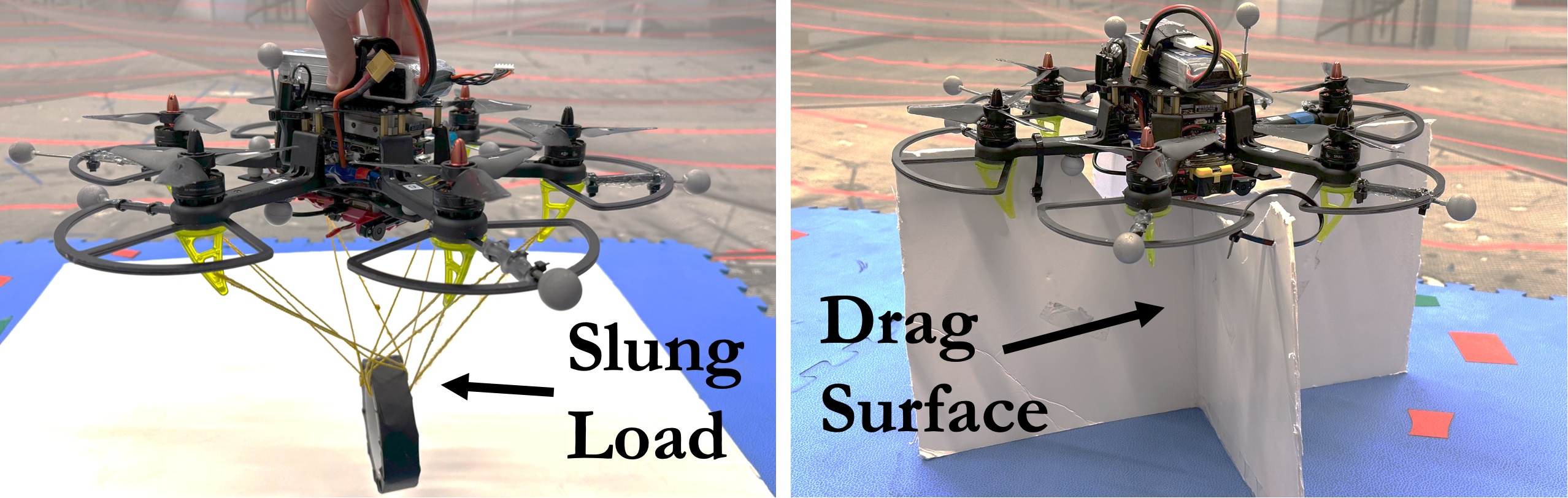}
    \end{subfigure}%
    
    \begin{subfigure}{.333\columnwidth}
      \centering
      \includegraphics[width=\columnwidth, trim = {24in, 0.5in,  22.5in, 0.5in}, clip]{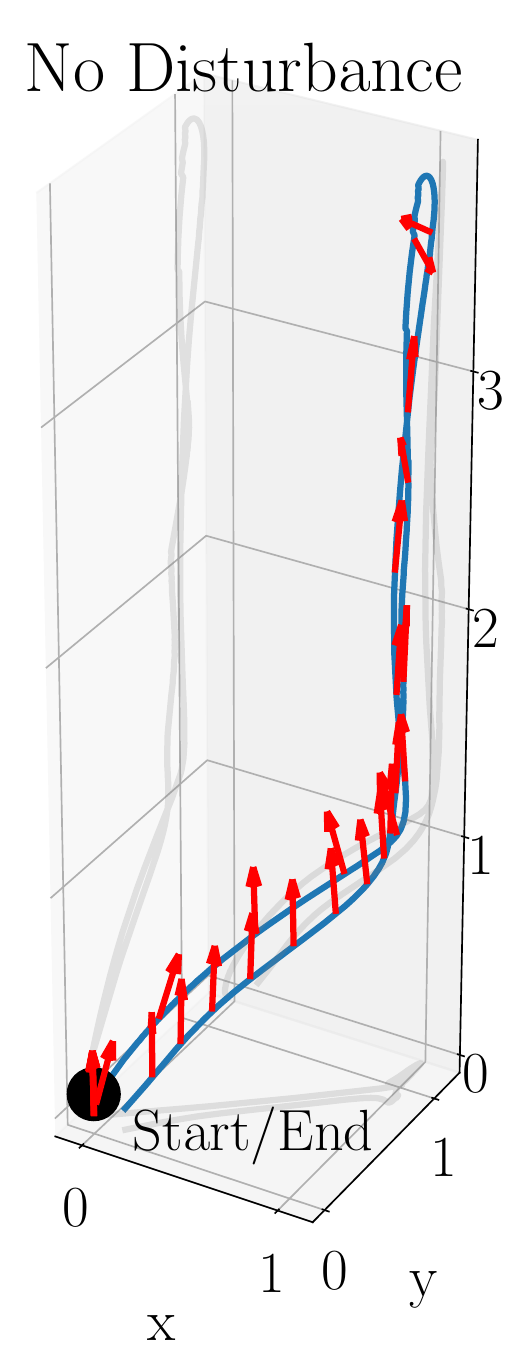}
    \end{subfigure}%
    \begin{subfigure}{.333\columnwidth}
      \centering
      \includegraphics[width=\columnwidth, trim = {24in, 0.5in,  22.5in, 0.5in}, clip]{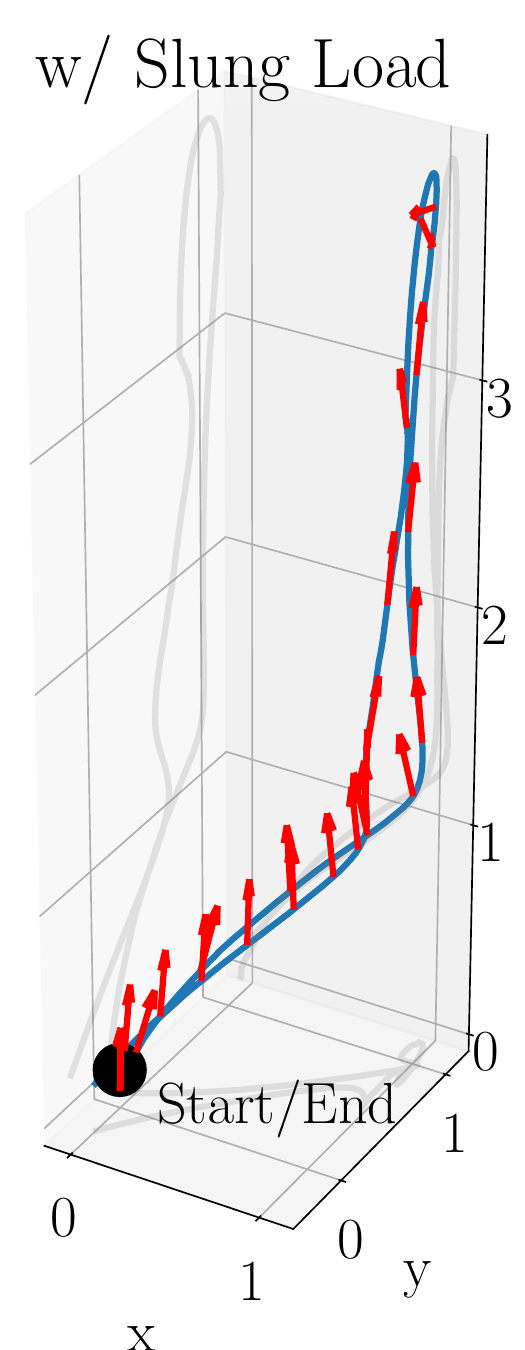}
    \end{subfigure}%
    \begin{subfigure}{.333\columnwidth}
      \centering
      \includegraphics[width=\columnwidth, trim = {24in, 0.5in,  22.5in, 0.5in}, clip]{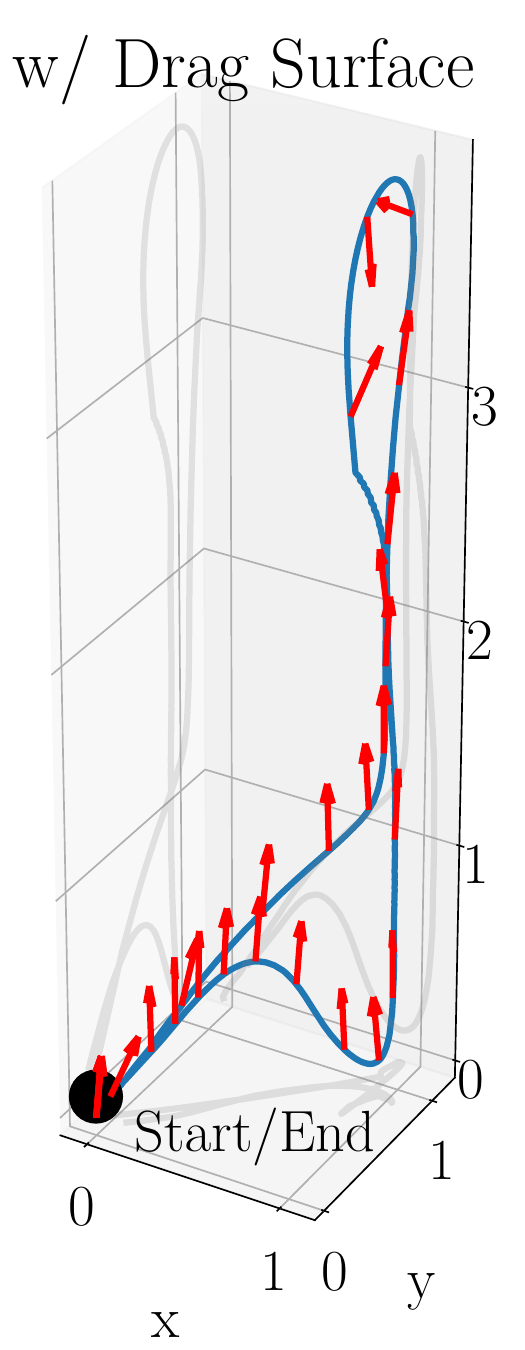}
    \end{subfigure}%
    \caption{Aerobatic (flip) flight in experiments, using a policy learned from a nonlinear Robust Tube MPC in about $100$ s of data collection (in sim., on a single CPU) and training time. 
    The policy runs onboard (TX2, CPU, at up to $500$ Hz, average inference time $0.015$ ms) and is robust to  disturbances (slung load of $0.18$ Kg., drag surface of $0.2$ m$^2$ and $0.13$ Kg). Red arrows denote the direction of the thrust vector, showing that the flip occurs at the point of highest altitude. Units in (m).} 
    \label{fig:rtnmc_experiment_state_traj}
    \vspace{-3ex}
\end{figure}

\begin{figure}
    \centering
    \includegraphics[width=\columnwidth, trim={0.4cm 0.5cm 0.5cm 0.5cm}, clip]{figs/flip_cmds/flip_cmds_v6.png}
    \caption{
    Control inputs and relevant states during the real--world acrobatic flip maneuver. 
    Despite the large level of uncertainties (inaccurate thrust-to-battery voltage mappings, hard-to-model aerodynamic effects), that require the usage of the maximum thrust allowed, the maneuver is completed successfully, performing a flip with an angular velocity of about $11$ rad/s.
    Note that the actual thrust $t_\text{cmd}$ can be related to the normalized thrust $\bar{t}_\text{cmd}$ via $t_\text{cmd} = m g (1 + \bar{t}_\text{cmd})$,
    where $ m g $ is the weight force of the robot.}
    \label{fig:rtnmc_experiment_actuation}
    \vspace{-3ex}
\end{figure}

\subsection{Hardware Evaluation}
We experimentally evaluate the ability of the policy to perform a flip on a real multirotor, under real-world uncertainties such as model errors (e.g., inaccurate thrust to battery voltage mappings, aerodynamic coefficients, moments of inertia) and external disturbances (e.g., ground effect). The tested policy is obtained using DAgger+SA-$25$ trained after $2$ demonstrations (the first with \ac{DA}, the second for fine-tuning), as the method represents a good trade-off between performance, robustness and training time. As in \cref{sec:evaluation_linear}, we deploy the learned policy on an onboard \texttt{Nvidia Jetson TX2}, where it runs at $100$ Hz. The maneuver includes a take-off/landing phase consisting of a $1$ m ramp on $x$-$y$-$z$ in $\text{W}$ and overall has a total duration of $6$ s. The maneuver is repeated $5$ times in a row, to demonstrate repeatability, recording successful execution of the maneuver and successful landing at the designated location in all the cases. \cref{fig:flip_timelapse} shows a time-lapse of the different phases of the maneuver (excluding the ramp from and to the landing location). The 3D position of the robot, as well as the direction of its thrust vector, are shown for two runs in \cref{fig:rtnmc_experiment_state_traj}, highlighting the large distance and altitude traveled in a short time. \cref{fig:rtnmc_experiment_actuation} additionally shows some critical parameters of the maneuvers, such as the attitude and the angular velocity, as well as thrust and the vertical velocities. It highlights that the robot rotates at up to $11$ rad/s, and the overall $360^\circ$ rotation takes about $0.5$ s. \b{In addition, the maneuver is repeated under even more challenging uncertainties, obtained by attaching either (a) a slung-load or (b) a drag surface to the robot, as shown in \cref{fig:rtnmc_experiment_state_traj}, deploy the policy onboard at $500$ Hz. The experiment is repeated $3$ consecutive times per disturbances, achieving $100\%$ success rate, and a resulting trajectory for each disturbance is shown in \cref{fig:rtnmc_experiment_state_traj}\footnote{Note that the gains of the cascaded attitude controller were increased compared to the scenario without extra disturbances. This was done to account for the fact that the attitude controller is not explicitly robust/adaptive to these new uncertainties, unlike the learned policy}}. Overall, these results validate our numerical analysis and highlight the robustness and performance of a policy efficiently trained from $2$ demonstrations and about $100$ s of training time. Our video submission \cite{video_submission} includes an additional experiment demonstrating near-minimum time navigation from one position to another, starting and ending with velocity close to zero, using a policy trained with two demonstrations (DAgger+SA-$25$).

\section{Discussion, Limitations and Future Work} \label{sec:conclusion}

This work has demonstrated that it is possible to generate policies from \ac{MPC} that are fast and robust in the real world, while requiring
\begin{inparaenum}[(a)]
    \item few queries to the expensive controller, 
    \item few environment interactions,
    \item and short training times.
\end{inparaenum}
Evaluations have validated the performance and data/computation efficiency, additionally showing that increasing the number of samples in \ac{DA} or introducing a fine-tuning procedure can further improve performance. 
These findings have broad applicability beyond the \ac{MPC} and \ac{IL} communities. For example, our method can serve as an efficient policy pre-training procedure, using model and uncertainty priors, for subsequent fine-tuning via model-free \ac{RL}, reducing inefficient random exploration in \ac{RL} or simplifying reward design.

We acknowledge some limitations, which open many exciting opportunities for future work. 
First, while our methodology has demonstrated real-world robustness, in the future we would like to leverage \ac{DNN} reachability tools \cite{everett2021reachability, rober2023backward, sidrane2022overt} to provide robustness certificates, enabling the deployment on safety-critical systems.  
Second, while easy-to-compute fixed-size approximations of the tube have been sufficient to guide our \ac{DA} strategy, 
future work will focus on leveraging tubes with varying cross-sections, enabling even more aggressive expert demonstrations. 
Third, while our approach showed robustness to small uncertainties in the rotational dynamics, we aim to combine our method with adaptive variants of the cascaded attitude controllers to avoid tuning the attitude controller under large uncertainties in the rotational dynamics.
Additionally, we would like to exploit the training efficiency of our approach to design adaptation strategies for trajectory tracking, for example by quickly generating new policies once new estimates of the model/environment become available. Last, we would like to leverage the efficiency and robustness of the obtained policies on aerial platforms with extreme payload/compute constraints \cite{giernacki2017crazyflie, chen2021collision}.

\section{CONCLUSION}
This work has presented an \ac{IL} strategy to efficiently train a robust \ac{DNN} policy from \ac{MPC}. Key ideas were to 
\begin{inparaenum}[(a)]
\item leverage a Robust Tube variant of \ac{MPC}, called \ac{RTMPC}, to collect demonstrations using existing \ac{IL} methods (\ac{DAgger}, \ac{BC}), and
\item augment the collected demonstrations with \textit{efficiently-generated} extra state-and-actions samples \textit{from the tube of the controller}, an approximation of the support of the state distribution that the learned policy will encounter when subject to uncertainties.
\end{inparaenum}
While the linear ancillary controller in linear \ac{RTMPC} provides extra data in a computationally efficient way, as shown in our conference paper \cite{tagliabue2022efficient}, the same efficiency can be challenging to achieve when leveraging nonlinear variants of \ac{RTMPC} \cite{mayne2011tube}. 
Therefore, in this journal extension of \cite{tagliabue2022efficient} we have presented a strategy to efficiently perform tube-guided \ac{DA} leveraging a sensitivity-based approximation of the ancillary controller in NMPC and using a fine-tuning phase to reduce the errors caused by these approximations.
Experimental evaluations on a multirotor have validated our numerical findings of efficiency and robustness, showing that a policy trained in only $100$ s can perform a flip under uncertainties, while requiring only $15 $ $\mu s$ to compute commands onboard.

\bibliographystyle{IEEEtran} %

\bibliography{root}

\begin{IEEEbiography}[{\includegraphics[width=1in,height=1.25in,trim={0, 18cm, 0, 6cm},clip,keepaspectratio]{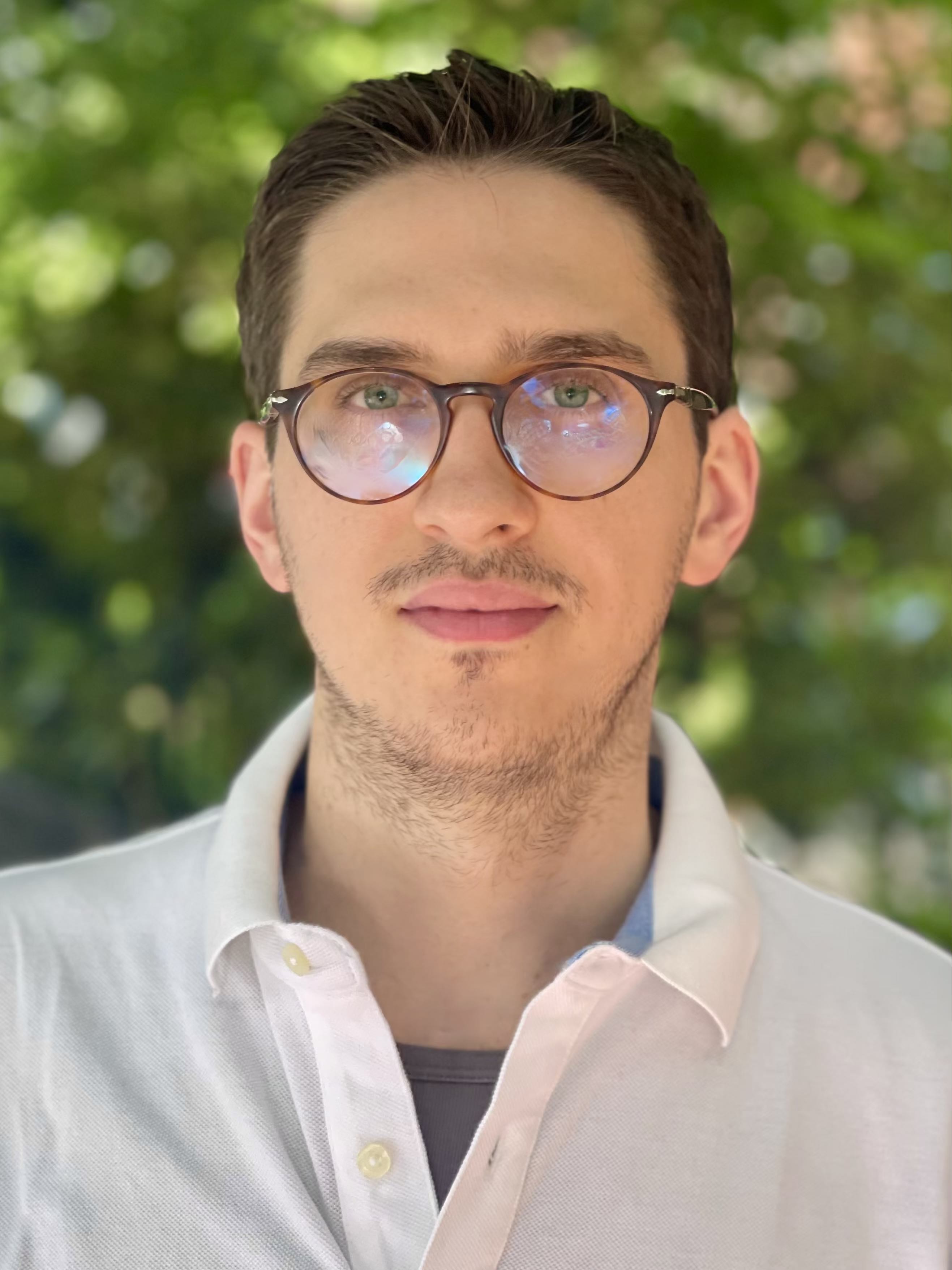}}]{Andrea Tagliabue}
received the B.Sc. degree in automation engineering from Politecnico di Milano, Milano, Italy (2015), the M.Sc. degree in robotics, systems, and control from ETH Zurich, Zurich, Switzerland (2018), and a Ph.D. degree in aeronautics and astronautics from the Massachusetts Institute of Technology (MIT), Cambridge, MA, USA (2024). Dr. Tagliabue was a visiting researcher at U.C. Berkeley, Berkeley, CA, USA (2017-2018), and an Engineer Affiliate with NASA's Jet Propulsion Laboratory, Pasadena, CA, USA (2018-2019). Dr. Tagliabue's interests include learning, perception and control for agile systems. His work was awarded finalist for the Best Paper in Dynamics and Control at ICRA 2023.
\end{IEEEbiography}
\begin{IEEEbiography}[{\includegraphics[width=1in,height=1.25in,clip,keepaspectratio]{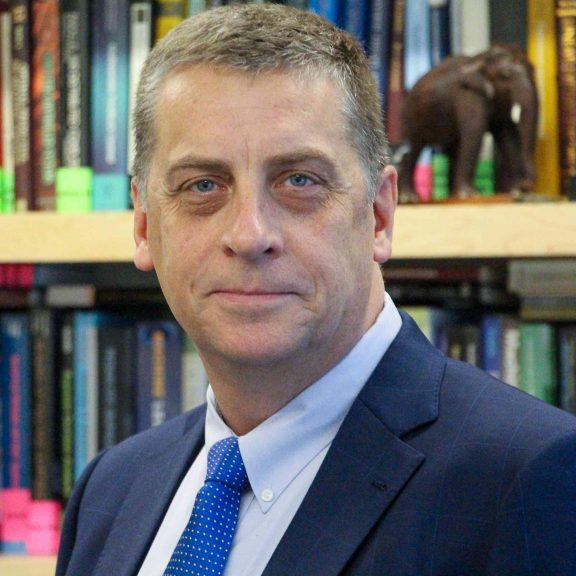}}]{Jonathan P. How (Fellow, IEEE)}
received the B.A.Sc. degree in aerospace from the University of Toronto, Toronto, Canada, in 1987, and the S.M. and Ph.D. degrees in aeronautics and astronautics from the Massachusetts Institute of Technology (MIT), Cambridge, MA, USA, in 1990 and 1993, respectively. Prior to joining MIT in 2000, he was an Assistant Professor with Stanford University, Stanford, CA, USA. He is currently the Richard C. Maclaurin Professor of aeronautics and astronautics at MIT. Dr. How’s awards include the IEEE CSS Distinguished Member Award (2020), AIAA Intelligent Systems Award (2020), IROS Best Paper Award on Cognitive Robotics (2019), and the AIAA Best Paper in Conference Awards (2011, 2012, 2013). He was the Editor-in-Chief of IEEE Control Systems Magazine (2015–2019), is a Fellow of AIAA, and was elected to the National Academy of Engineering in 2021.
\end{IEEEbiography}
\vfill
\end{document}